\documentclass{article}
\usepackage{jfrExamplee}
\usepackage{graphicx}
\usepackage{apalike}
\usepackage{setspace}

\usepackage{amsmath}
\newcommand{\argmin}{\operatornamewithlimits{argmin}}
\usepackage{tabu}
\usepackage{multirow}
\usepackage{multicol}
\usepackage{array}
\usepackage{adjustbox}

\usepackage{color}
\usepackage{url}
\usepackage{subcaption}
\usepackage{enumitem}
\usepackage{soul}
\usepackage{makecell}
\usepackage{siunitx}

\title{Semantic Mapping for Orchard Environments by Merging Two-Sides Reconstructions of Tree Rows} % Tree Morphology for Phenotyping from\\ Semantic 3D Vision in Orchard Environments
\author{
Wenbo Dong\\
\texttt{dongx358@umn.edu}\\
\And
Pravakar Roy\\
\texttt{royxx268@umn.edu}\\
\And
Volkan Isler\thanks{W.~Dong, P.~Roy, and V.~Isler are with the Department of Computer Science and Engineering, University of Minnesota, Twin Cities, MN 55455.}\\
\texttt{isler@umn.edu}
}

\def\presuper#1#2%
  {\mathop{}%
   \mathopen{\vphantom{#2}}^{#1}%
   \kern-\scriptspace%
   #2}
\begin{document}

\maketitle

\begin{abstract}
Measuring semantic traits for phenotyping is an essential but labor-intensive activity in horticulture. Researchers often rely on manual measurements which may not be accurate for tasks such as measuring tree volume. To improve the accuracy of such measurements and to automate the process, we consider the problem of building coherent three dimensional (3D) reconstructions of orchard rows. Even though 3D reconstructions of side views can be obtained using standard mapping techniques, merging the two side-views is difficult due to the lack of overlap between the two partial reconstructions. Our first main contribution in this paper is a novel method that utilizes global features and semantic information to obtain an initial solution aligning the two sides. Our mapping approach then refines the 3D model of the entire tree row by integrating semantic information common to both sides, and extracted using our novel robust detection and fitting algorithms. Next, we present a vision system to measure semantic traits from the optimized 3D model that is built from the RGB or RGB-D data captured by only a camera. Specifically, we show how canopy volume, trunk diameter, tree height and fruit count can be automatically obtained in real orchard environments. The experiment results from multiple datasets quantitatively demonstrate the high accuracy and robustness of our method.
\end{abstract}
\section{Introduction} \label{sec:introduction}
\begin{figure}[t]
	\centering
	\begin{subfigure}{0.23\textwidth}
		\centering
		\includegraphics[width=1\linewidth]{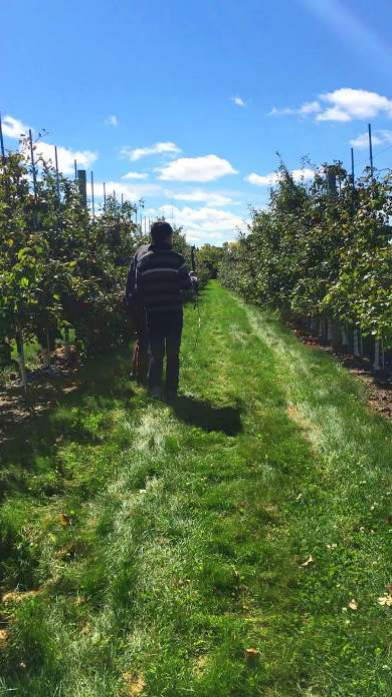}
		\caption{Data collection}
	\end{subfigure}
	\qquad
	\begin{subfigure}{0.23\textwidth}
		\centering
		\includegraphics[width=1\linewidth]{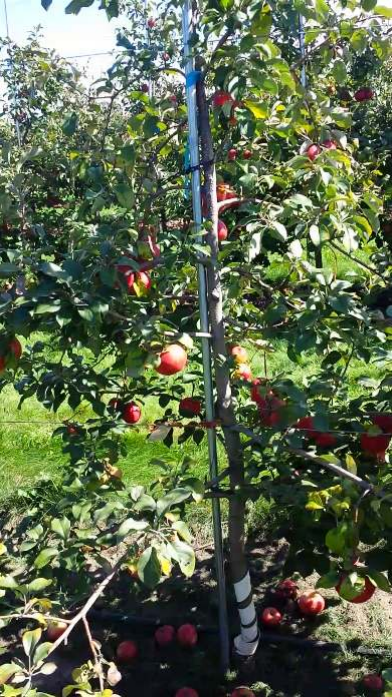}
		\caption{Side-view imaging}
	\end{subfigure}
	\qquad
	\begin{subfigure}{0.23\textwidth}
		\centering
		\includegraphics[width=1\linewidth]{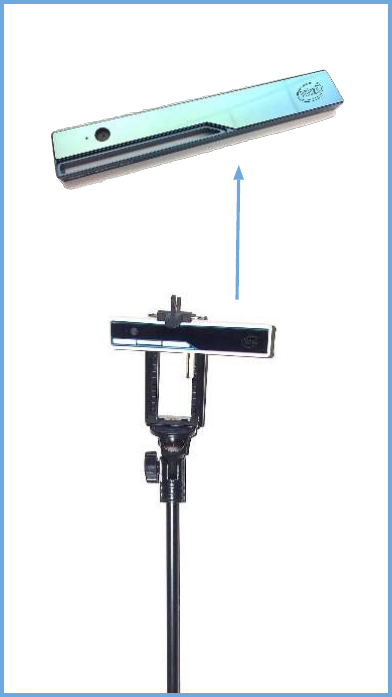}
		\caption{Imaging device}
	\end{subfigure}
	\caption{Overview of data capturing scenario. (a): A single-side data collection along a tree row. (b): The front view of an apple tree. Many apples in this tree are visible from both sides. If we add counts from individual sides, these apples will be counted twice. (c): The imaging device (Intel RealSense R200) capturing RGB or RGB-D data. The camera is mounted on a stick to capture data from either horizontal view or titled top-down view.}
	\label{fig:overview}
\end{figure}

The problem of building accurate 3D reconstructions of orchard rows arises in a number of agricultural automation tasks.
%3D models can be used for automated pruning of trees.
The estimation of morphological parameters of fruit trees (such as tree height, canopy volume and trunk diameter) is important in horticultural science, and has become an important topic in precision agriculture~\cite{rosell2012review,tabbrobotic}.
Accurate morphology estimation can help horticulturists study to what extent these parameters impact crop yield, health and development. For example, growers try different rootstocks to figure out which one produces a better yield per volume for a specific geographical area. They also measure parameters such as tree height or trunk diameter to model fruit production. This measurement process is labor-intensive and not necessarily accurate. However, these geometric traits used for phenotyping can be accurately extracted from reconstructed 3D models.
3D models can be further used for automated pruning of trees, and are also important for yield mapping: although image-based methods can be used to count fruits in individual images, 3D models can be used for tracking them across images and to avoid double counting fruits visible from both sides of the row. For example, in the case of the tree shown in Fig.~\ref{fig:overview}b, most of the apples are visible from both sides and can be counted twice in independent single-side scans.

There are many techniques such as Structure from Motion (SfM) or RGB-D Simultaneous Localization and Mapping (SLAM)~\cite{sturm2012benchmark,roy2016surveying} which can generate reconstructions of individual sides of the rows. However, existing methods can not merge these two reconstructions: even with the manual selection of correspondences, Iterative Closest Point (ICP) techniques fail (see Fig.~\ref{fig:simulationCompare}). Large-scale SfM techniques can produce consistent reconstructions with the presence of overlapping side views or with loop closure. Obtaining such views in orchard settings is hard because the rows can be extremely long (sometimes spanning a thousand meters or more). The use of very precise Real-Time Kinematic (RTK) GPS can be used to solve the registration problem, but it is costly and not always available.

%\begin{figure}[t]
%	\centering
%	\includegraphics[width=0.85\columnwidth]{figs/overview.pdf}
%	\caption{Overview of data capturing scenario. (a): The imaging device (Intel RealSense R200) capturing RGB or RGB-D data. (b): The camera is mounted on a stick to capture data from either horizontal view or titled top-down view.}
%	\label{fig:overview}
%    %\vspace*{-4mm}
%\end{figure}

The goal of our work is to use RGB or RGB-D videos to reconstruct a complete 3D model of tree rows from images of both sides and to perform semantic mapping (measuring tree morphology and estimating yield).
In this paper, we present a novel method to merge reconstructions from both sides of a row without the need for overlapping views or GPS coordinates (see Fig.~\ref{fig:problemGoal}).
We utilize a key observation: ``The orthographic projection of the occlusion boundary of the trees in a row in the fronto-parallel plane from opposite sides of the row are symmetric.''
Coupling this fact with the assumption of the existence of a common ground plane, we solve the problem of merging the reconstructions from both sides first by solving the problem of finding a rigid body transformation between the occlusion boundaries of a fronto-parallel view.
Next, we address the problem of estimating a single overlapping depth distance. We solve this problem using existing 2D shape matching methods and semantic constraints (e.g., the tree trunks are well approximated by cylinders, their projections in the side view and the front view have the same width).
%The method is evaluated using multiple simulated and real datasets.
Our method relies on establishing semantic relationships between each of the two-sides and integrating tree morphology into the reconstruction system, which in turn outputs optimized morphological parameters.

Fig.~\ref{fig:overview} illustrates an overview of our data collection. To the best of our knowledge, it is the first vision system for accurate estimation of tree morphology and fruit yield in orchards by using only an RGB or RGB-D camera (depth information is used only for measuring tree traits in the absolute scale). In summary, our work has the following key contributions:
\begin{itemize}
	\item We present a novel mapping approach on RGB or RGB-D videos that can separately reconstruct 3D models of fruit trees from both sides and accurately merge them based on semantics, i.e., tree trunks and local ground patches.
	\item We introduce robust detection and fitting algorithms to estimate the initial trunk size and local planar ground for each tree, and integrate tree-trunk diameters into semantic SfM to further localize trunks and local ground patches.
%	\item We integrate tree-trunk diameters into semantic SfM to further localize trunks and local ground patches.
	\item We measure tree height, canopy volume and trunk diameter through automated segmentation for each tree, and count fruits based on optimized information of 3D merged tree rows.
\end{itemize}

This paper is structured as follows. In the next section, we discuss the relevant literature. After discussing technical challenges, we introduce our proposed semantic mapping, followed by experimental results and a conclusion.

\section{Related Work} \label{sec:relatedWork}
2D or 3D LIDAR scanning has proven to be a viable option for generating 3D models of trees~\cite{underwood2015lidar,bargoti2015pipeline}. Usually, LIDAR sensors are mounted on a vehicle moving along the alleys of the fruit orchard to vertically scan the side of the tree rows~\cite{mendez2014deciduous,underwood2016mapping}. To obtain the 3D point cloud by adding subsequent 2D transects of laser scanning, the vehicle has to move with a steady velocity and along a linear track parallel to the tree row. However, these systems do not merge two scanned sides of trees. Morphological parameters are thus inaccurately computed by only scanning one side and multiplying by two or by adding the volumes of the two sides without merging them.
%based on the assumption that the tree canopy is symmetric from both sides. 
Generated two-sides point clouds can also be manually matched through CAD software~\cite{rosell2009obtaining}. However, tree models are partially misaligned from two sides due to accumulated errors of sensor poses during the movement. Even if position accuracy has been improved by combining Global Navigation Satellite System (GNSS) with LIDAR~\cite{del2015georeferenced}, the issue of accumulated orientation error still exists, especially for large scale scanning. Furthermore, the combination of these two sensors (e.g., GR3 RTK GNSS and LMS500) is expensive and may not be affordable.

Cameras are low-cost, lightweight compared to LIDAR sensors. Vision-based 3D dense reconstruction, with the ability to provide quantitative information of every geometric detail of an object, is a promising alternative for accurate morphology measurement. Although time-of-flight~\cite{van2012spicy}, stereo-vision systems~\cite{bac2014stem} and depth sensors~\cite{wang2014size} have been used to estimate parameters of low-height plants, these approaches have been limited to indoor environments with controlled conditions, such as constant background and artificial illumination. We focus on the outdoor case in natural orchard environments.
%As RGB-D cameras have been developed for both indoor and for outdoor environments, efficient 3D dense reconstruction becomes possible for low-height plants.

There has been a lot of recent work on yield estimation for specialty crops \cite{wang,das2015devices,roy2016surveying,roy2016counting,bargoti2017image,roy2017active,hani2018counting,hani2018comparative}. Most of the existing systems rely on external sensors to register the fruits from a single side or both sides of the row. Wang et al.~\cite{wang} use stereo cameras coupled with GPS and odometry sensors to avoid double counting. They align the apples globally in 3D space and remove the ones which are within $0.05$ meter of a previously registered apple. Hung et al.~\cite{hung2015feature} and Bargoti et al.~\cite{bargoti2017image} use sampling at certain intervals to remove overlap between images. Das et al.~\cite{das2015devices} use optical flow and navigational sensors to avoid duplicate apples. In our previous work~\cite{roy2016surveying}, we present a method for registering apples from the single side of a row based on affine tracking and incremental SfM. None of these previous methods build a consistent 3D model of the entire row. In contrast, we focus on aligning SfM reconstructions from both sides of a row to create a single consistent model.

Recent studies present methods for merging visually disconnected SfM models~\cite{cohen2016indoor} for urban environments. These methods estimate the scale and relative height for all sub-models based on the Manhattan world assumption and find possible connection points between the reconstructions utilizing semantic information (facade edges, windows, doors, etc). Afterward, they generate all possible fully connected models and find the most likely model according to loop closure and symmetry alignment. Modern map merging approaches~\cite{yu2015semantic,zhou2016fast,bonanni20173} register multiple partially overlapping 3D maps by non-rigid alignment. In contrast, there is no overlap between partial reconstructions in our scenario. For aligning SfM reconstructions from both sides of a tree row, we utilize the symmetry of their occlusion boundaries from a fronto-parallel view. Afterward, we adjust the overlap in depth direction using semantic information.

For a modern high-density orchard setting, it is not possible to perform mapping around each tree individually. Instead, two sides of tree rows are captured separately by a moving camera or in a loop trajectory. Obtaining accurate 3D models of fruit trees requires accurate camera poses, but estimating them reliably for long range RGB videos is a difficult problem. Especially in orchard environments, good features cannot be stably tracked through long subsequent frames because of motion due to the wind in the scene~\cite{dong2017linear}. Accumulated errors in camera poses will cause misalignment of tree models from both sides. As we show in Sec.~\ref{sec:technicalBackground}, state-of-the-art methods for volumetric fusion~\cite{newcombe2011kinectfusion}, SfM~\cite{wu2013towards} and SLAM~\cite{mur2017orb} are not reliable enough for tree volume and trunk diameter estimation. Since there is nearly no overlap of canopy surface between two sides of tree rows, misalignment of tree models cannot be addressed by ICP-based methods~\cite{medeiros2017modeling} or semantic tracking in loop closure~\cite{bowman2017probabilistic}.% (see Sec.~\ref{sec:technicalBackground} for details).

%----------------------------------------------------------------------
% SECTION II: Technical Background
%----------------------------------------------------------------------
\section{Technical Background} \label{sec:technicalBackground}
This section provides the problem formulation of semantic mapping for orchards with an overview of our system, and two main challenges of 3D reconstruction in orchard environments.
%This section provides the problem formulation of tree morphology estimation with an overview of our system, and two main challenges of 3D reconstruction in orchard environments.

\subsection{Problem Formulation} \label{subsec:problemFormulation}%of Two-sides Semantic Mapping
\begin{figure}[t]
	\centering
	\includegraphics[width=0.9\columnwidth]{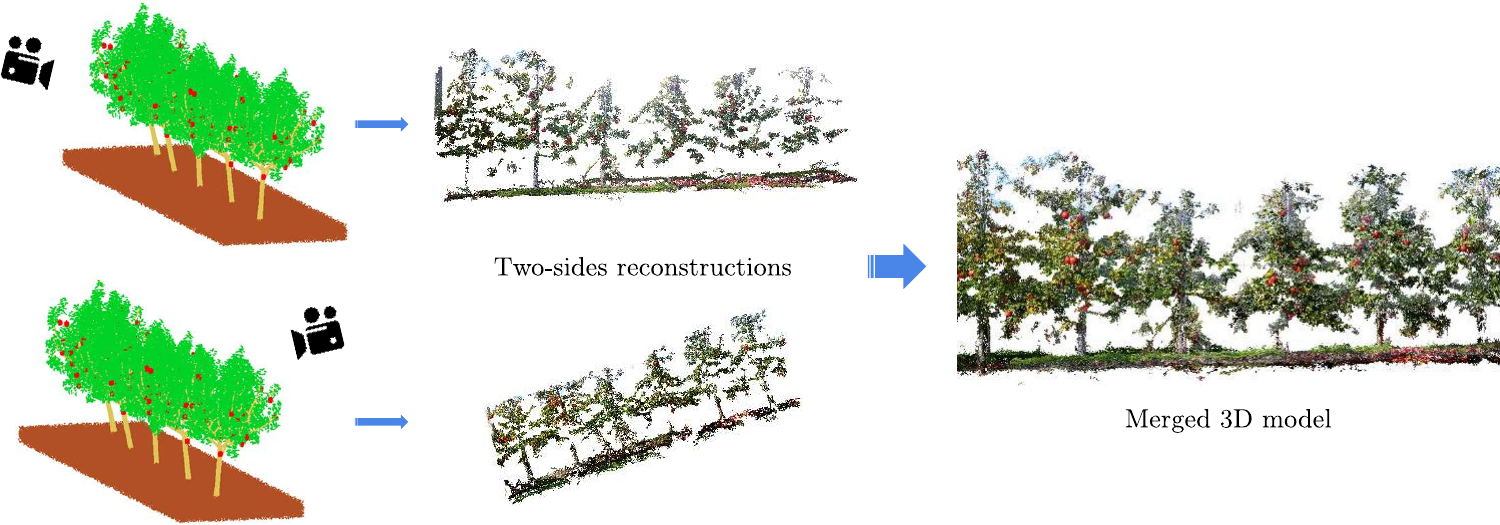}
	\caption{Given two up-to-scale reconstructions of two sides of a row, our goal is to merge the two reconstructions into a single coherent model.}
	\label{fig:problemGoal}
	%\vspace*{-5mm}
\end{figure}

%\subsection{Problem Formulation of Tree Morphology Estimation}
Consider a row of trees in an orchard, where an imaging device moves along one side of the row (arbitrarily called the ``front'' side) and captures images of static landmarks (3D points and 3D objects, such as trunks and local grounds), then it moves to the ``back'' side and captures the second set of images. The images can be standard RGB images or they may also include depth information (RGB-D data). The images in each set are used to obtain two independent reconstructions represented as point clouds. The main problem we address is to merge and optimize these two reconstructions (see Fig.~\ref{fig:problemGoal}), which is formalized as follows.
%Consider the problem of tree morphology estimation, in which a mobile camera separately moving along both sides of a tree row collects the RGB-D data of static landmarks (3D points and 3D objects, such as trunks and local grounds).

Given a set of imaging measurements $\bar{\mathcal{X}}_k$ extracted from two sets of images that capture the front and back sides of a row $\{{\mathcal{F}}, {\mathcal{B}}\}$, and object types $\mathcal{I}_j$, 
the task is to find a similarity transformation ${^{\mathcal{F}}_{\mathcal{B}}\mathcal{T}}$ that merges the back-side reconstruction with the front, and further estimate the object poses $\mathcal{S}^{\mathcal{I}}_j$ with their sizes $\mathcal{D}^{\mathcal{I}}_j$, along with the 3D point positions $\mathcal{X}_i$ and camera poses $\mathcal{C}_k$:
%The true models of the two sides are related by a rigid transformation $\mathcal{T}$.
%Given a set of RGB-D measurements $\bar{\mathcal{X}}_k$ and object types $\mathcal{I}_j$, the task is to estimate the object poses $\mathcal{S}^{\mathcal{I}}_j$ with their sizes $\mathcal{D}^{\mathcal{I}}_j$, the transformation $\mathcal{T}$, along with the 3D point positions $\mathcal{X}_i$ and camera poses $\mathcal{C}_k$:
\begin{equation} \label{probForm}
\argmin\limits_{\mathcal{S}^{\mathcal{I}}_j, \mathcal{D}^{\mathcal{I}}_j, \mathcal{X}_i, \mathcal{C}_k} \sum\limits_{j} \sum\limits_{k} \sum\limits_{i \in \mathcal{V}(j,k)} E_{\mathcal{S}} (\bar{\mathcal{X}}_k, {^{\mathcal{F}}_{\mathcal{B}}\mathcal{T}}, \mathcal{S}^{\mathcal{I}}_j, \mathcal{D}^{\mathcal{I}}_j, \mathcal{X}_i, \mathcal{C}_k) + \sum\limits_{k} \sum\limits_{i \in \mathcal{V}(k)} E_{\mathcal{X}} (\bar{\mathcal{X}}_k, {^{\mathcal{F}}_{\mathcal{B}}\mathcal{T}}, \mathcal{X}_i, \mathcal{C}_k) ,
\end{equation}
where ${^{\mathcal{F}}_{\mathcal{B}}\mathcal{T}}$ as one of the inputs is first calculated by minimizing the distance cost $E_{\mathcal{T}}$ for registering two-sides point clouds:
\begin{equation}
\argmin\limits_{{^{\mathcal{F}}_{\mathcal{B}}\mathcal{T}}} E_{\mathcal{T}} \left({\bigcup\limits_k \bar{\mathcal{X}}_k}, {\bigcup\limits_{i \in \mathcal{F}}\mathcal{X}_i}, {\bigcup\limits_{i \in \mathcal{B}}\mathcal{X}_i}, {^{\mathcal{F}}_{\mathcal{B}}\mathcal{T}}\right) ,
\end{equation}
$E_{\mathcal{S}}$ is the cost between a measured point and the object it belongs to, and $E_{\mathcal{X}}$ is the cost between a 3D point visible from a camera frame and its measurement.

\begin{figure}[t]
	\centering
	\includegraphics[width=0.9\columnwidth]{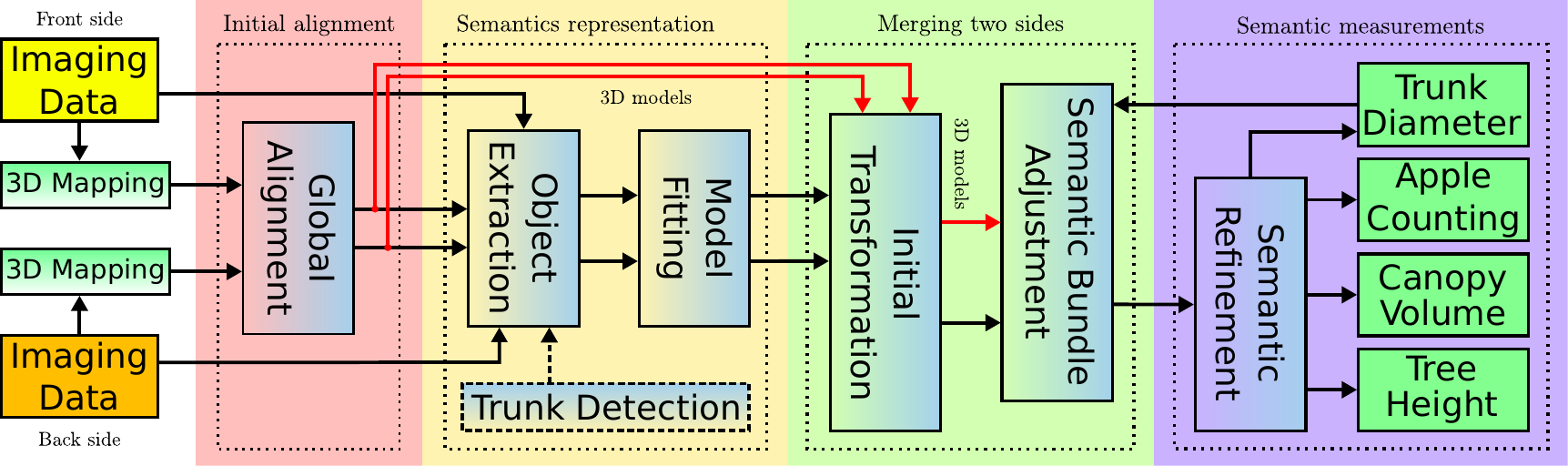}
	\caption{Overview of the proposed system for semantic mapping in orchards. The trunk detection (dashed line) in object extraction can be replace by object detection~\cite{salas2013slam++} if there is no need for trunk diameter estimation. The four colored blocks represent the four steps of our method.}
	\label{fig:system}
	%\vspace*{-5mm}
\end{figure}

The proposed vision system for semantic mapping is illustrated in Fig.~\ref{fig:system}. The estimation procedure is divided into four steps explained in Sec.~\ref{sec:methodology}. We note that even though our approach  starts with two independent reconstructions of the two sides, it refines them based on semantic information.

\subsection{Technical Challenges} \label{subsec:technicalChallenges}
In modern orchards, fruit trees are highly packed in each row and connected by supporting wires (see Fig.~\ref{fig:overview}). Without enough separation space, it is not possible to individually perform surrounding imaging data collection around each tree. Instead, we collect side-view data of tree rows by moving the camera along the path between tree rows.
%From the purpose of phenotyping study, different types of fruit trees are always planted in each row (around 100 trees per row) which is about 100 meters long. 
The rows can be hundreds of meters long. But the specific region of interest for a particular study can be only a subset of the row.
If we measure only this region from the two sides, the images across the sides may have no overlap. Alternatively, the entire row can be covered by following a loop around the row. In this section, we detail the technical challenges associated with these two approaches.

%\begin{figure}[t]
%	\centering
%	\includegraphics[width=0.9\columnwidth]{figs/background_loop.pdf}
%	\caption{The score matrix between all image frames generated by using a BoW model. High similarities are marked by colored boxes. The correct loop detection is marked by the red box. (a): Feature matching between a pair of frames detected by loop closure. (b): Camera trajectory before loop closure. (c): Camera trajectory after loop closure. (d): Simple alignment of two-sides 3D models is not feasible: camera trajectories from both sides are diverged and marked by the red box.}
%	\label{fig:loopClosure}
%	%\vspace*{-2mm}
%\end{figure}

\begin{figure}[t]
	\centering
	\includegraphics[width=0.8\columnwidth]{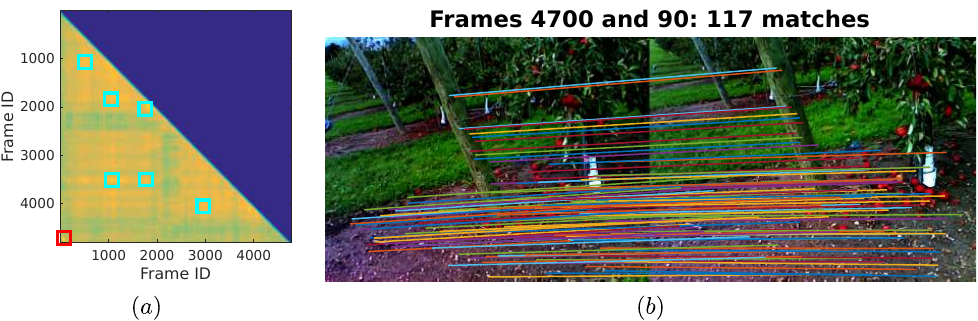}
	\caption{The score matrix between all image frames generated by using a BoW model. (a): High similarities are marked by colored boxes. The correct loop detection is marked by the red box. (b): Feature matching between a pair of frames detected by loop closure.}
	\label{fig:loopClosure}
	%\vspace*{-2mm}
\end{figure}

\begin{figure}[t]
	\centering
	\includegraphics[width=0.8\columnwidth]{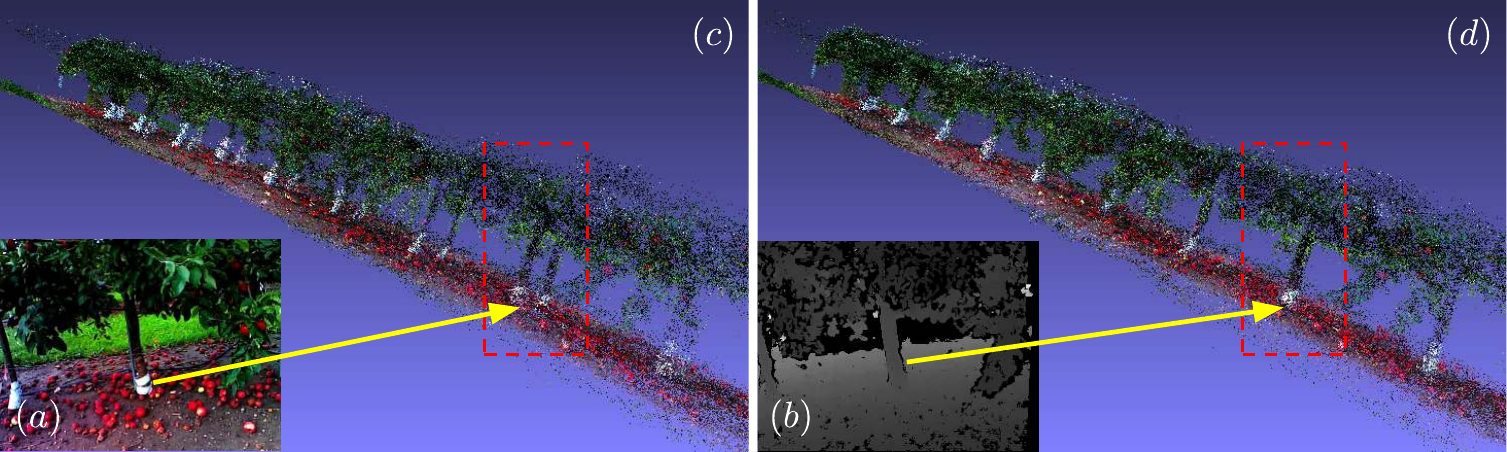}
	\caption{Even with loop closure, the 3D reconstruction of tree rows is not satisfactory: 3D models of tree trunks from two sides are misaligned. (a): The RGB image. (b): The depth image. (c): Misaligned trunks from both sides. (d): The 3D reconstruction is improved by integrating trunk information.}
	\label{fig:duplicated}
	%\vspace*{-4mm}
\end{figure}

First, ORB-SLAM2~\cite{mur2017orb} is tested on our RGB-D data captured in a loop around a tree row to create the 3D model. Unlike indoor cases, image features in orchard environments are unstable due to wind effect and thus hard to track across multiple frames, which causes the SLAM algorithm to frequently get lost. On the other hand, loop detection is not reliable because of the high similarity between fruit trees of the same type (see Fig.~\ref{fig:loopClosure}). With correct loop closure, the 3D dense reconstruction of the tree row from both sides is generated by converting depth maps into point clouds based on the optimized camera trajectory from the SLAM output. From Fig.~\ref{fig:duplicated}, we observe that although the loop is correctly closed the 3D model of the tree row is not satisfactory. The 3D dense reconstruction has separate trunks since there is no data overlap between two sides of the tree row. Measuring tree morphology based on inaccurate models is problematic, especially canopy volume and trunk diameter estimation.

\begin{figure}[t]
	\centering
	\includegraphics[width=0.95\columnwidth]{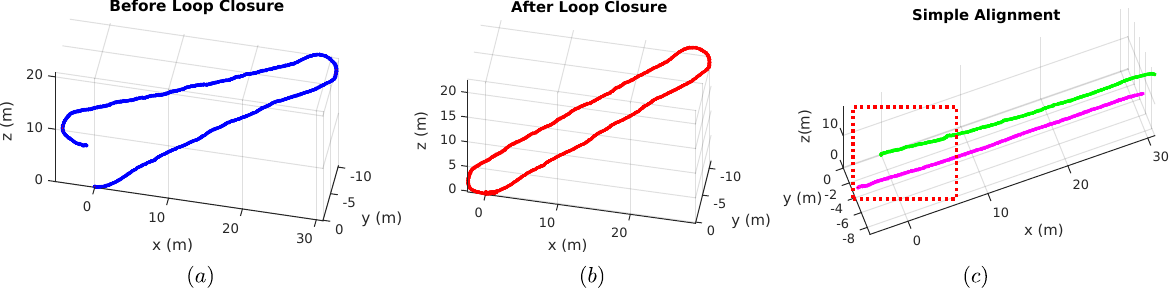}
	\caption{Estimated camera trajectories by SLAM algorithm. (a): Camera trajectory before loop closure. (b): Camera trajectory after loop closure. (c): Simple alignment of two-sides 3D models is not feasible: camera trajectories from both sides diverge as marked by the red box.}
	\label{fig:loopTrajectory}
	%\vspace*{-2mm}
\end{figure}

\begin{figure}[t]
	\centering
	\includegraphics[width=0.98\columnwidth]{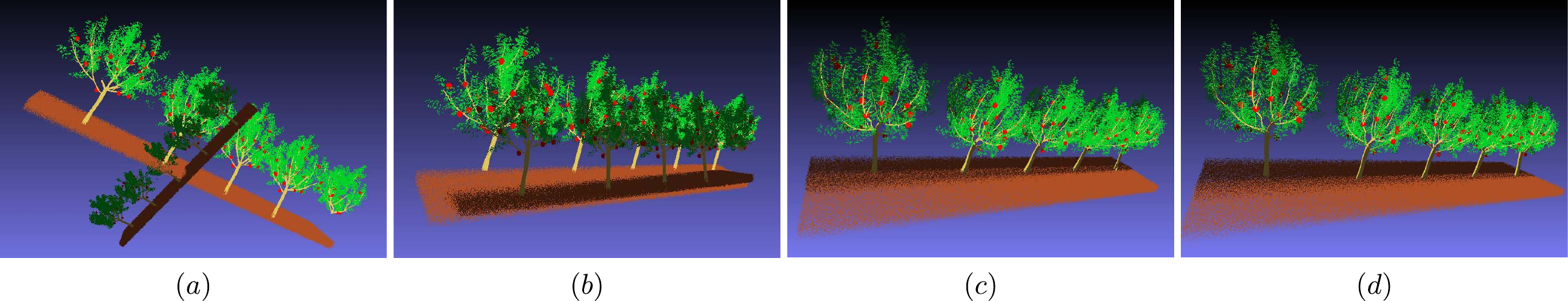}
	\caption{Comparison results using synthetic data. (a): Input reconstructions of two sides. (b): Applying the ICP algorithm fails to merge tree trunks. (c): Merging from the proposed initial alignment. (d): Further refinement using semantic information.}
	\label{fig:simulationCompare}
	%\vspace*{-2mm}
\end{figure}

For the data separately captured from both sides, simple alignment of two-sides 3D models can be performed by estimating the rigid transformation based on the trunks information. However, due to accumulated errors of camera poses, some trees are well-aligned from both sides (with parallel camera trajectories) while the rest are misaligned (see Fig.~\ref{fig:loopTrajectory}c). Fig.~\ref{fig:duplicated}d implies that two-sides 3D reconstruction should be further optimized based on semantic information to correct camera trajectories.
It is notable that, even with perfect reconstructions and manually provided correspondences, ICP-based methods fail to merge these two reconstructions (see Fig.~\ref{fig:simulationCompare}).
Standard SfM algorithm~\cite{wu2013towards} often fails to close loops when dealing with view-invariant feature matching, and may converge to a local minimum. Hence, we adjust the single-side 3D reconstruction by integrating essential elements from SLAM and SfM algorithms.

\subsection{Single-Side Reconstruction} \label{subsec:singleSideBA}
In this section, we present the proposed approach for initially reconstructing each side independently using established techniques.
For each pair of consecutive frames (in RGB or RGB-D data), the relative rigid transformation is calculated by applying a RANSAC-based three-point-algorithm~\cite{forsyth2011computer} on the SIFT matches~\cite{lowe2004distinctive} with valid depth values if available.
%For each pair of consecutive frames, the relative rigid transformation is calculated by applying a RANSAC-based three-point-algorithm~\cite{forsyth2011computer} on the SIFT matches~\cite{lowe2004distinctive} with valid depth values.
Pairwise Bundle Adjustment (BA) is performed to optimize the relative transformation and 3D locations of matches by minimizing 2D reprojection errors.
For loop detection, we build a Bag of Words (BoW) model~\cite{sivic2009efficient} to characterize each frame with a feature vector, which is calculated based on different frequencies of visual words. The score matrix is obtained by computing the dot products between all pairs of feature vectors (see Fig.~\ref{fig:loopClosure}).
Possible loop pairs are first selected by a high score threshold and then tested by RANSAC-based pose estimation which determines whether a reasonable number of good matches are obtained (e.g., 100 feature matches). Loop pairs are thus accurately detected and linked with pairs of consecutive frames by a covisibility graph. Loop detection allows us to capture every single tree back and forth from different views on a single side.

For each frame in consecutive pairs, we first perform local BA to optimize its local frames which have common features. To effectively close the loop, pose graph optimization~\cite{strasdat2010scale} is then performed followed by global BA to finally optimize all camera poses and 3D points.
Given the fact that depth maps in outdoor cases are generated by infrared stereo cameras, we integrate 3D error information into the objective function of bundle adjustment as follows:
\begin{equation} \label{objectBA}
\begin{gathered}
\argmin_{s_c, \mathbf{R}_c, \mathbf{t}_c, \mathbf{X}_p} J = \sum_{c} \sum_{p \in \mathcal{V}(c)} \rho\left( E_o(c,p) \right) + \eta \cdot \rho\left( E_i(c,p) \right) \\
E_o(c,p) = \| ^c\bar{\mathbf{x}}_p - \mathbf{K}_o s_c [\mathbf{R}_c | \mathbf{t}_c ] \mathbf{X}_p \|^2 \\
E_i(c,p) = \| \mathbf{K}_i [\mathbf{R}_i | \mathbf{t}_i ] ^c\bar{\mathbf{X}}_p - \mathbf{K}_i [\mathbf{R}_i | \mathbf{t}_i ] [\mathbf{R}_c | \mathbf{t}_c ] \mathbf{X}_p \|^2
\end{gathered} ,
\end{equation}
where $\eta = 0$ for RGB data, while for RGB-D data $\eta = 1$ and $s_c = 1$. 
$\rho$ is the robust Huber cost function~\cite{huber1992robust}, $\mathbf{K}_o$ and $\mathbf{K}_i$ are the intrinsics matrices of the RGB camera and the left infared camera, $[\mathbf{R}_i | \mathbf{t}_i ]$ is the relative transformation between these two cameras, $s_c[\mathbf{R}_c | \mathbf{t}_c ]$ is the RGB camera pose, $\mathbf{X}_p$ is the 3D location of a point visible from the camera frame $c$, and $^c\bar{\mathbf{x}}_p$ and $^c\bar{\mathbf{X}}_p$ are the observed 2D feature and 3D location in the RGB camera frame, respectively.

% to balance between the speed and the accuracy.

%\subsection{Technical Challenges} \label{subsec:technicalChallenges}
%3D mapping techniques for RGB-D cameras can be classified into feature-based methods and voxel-based methods. Feature-based methods perform 3D mapping by optimizing 3D feature points. ORB-SLAM2~\cite{mur2017orb} is one of the state-of-the-art for 3D mapping in real time. SfM algorithms are able to globally generate accurate 3D models using all unordered input-images but time-consuming. Voxel-based methods, such as KinectFusion~\cite{newcombe2011kinectfusion}, reconstruct 3D surface models by using a volumetric representation of the 3D space.
%
%As demonstrated in Sec.~\ref{subsec:technicalChallenges}, real-time SLAM algorithm cannot be directly applied in our scenario. For single-side reconstruction of tree rows, we take advantage of pipelines from both SLAM and SfM algorithms to balance between the speed and the accuracy. Depth maps are improved by using a Truncated Signed Distance Function (TSDF)~\cite{curless1996volumetric}.

%----------------------------------------------------------------------
% SECTION III: Methodology
%----------------------------------------------------------------------
\section{Methodology} \label{sec:methodology}
In this section, we present our main technical contribution: merging and refining the reconstructions of the two sides using semantic information.
% method of tree morphology estimation from merged 3D reconstruction of both sides. 
The proposed semantic mapping consists of four steps (see Fig.~\ref{fig:system}). The cost functions $E_{\mathcal{T}}$, $E_{\mathcal{S}}$ and $E_{\mathcal{X}}$ (described in Sec.~\ref{subsec:problemFormulation}) are specified and explained in Sec.~\ref{subsec:globalign}, and Sec.~\ref{subsec:merging}, respectively.

\subsection{Initial Alignment using Global Features} \label{subsec:globalign}
%\input{jfrGlobalAlignment}
%Consider a row of trees in an orchard. Suppose an imaging device moves along one side of the row  (which we arbitrarily call the ``front side") and captures images. Then it moves to the ``back" side and captures the second set of images.  The images can be standard RGB images or they may also include depth information. The images in each set are used to obtain two independent reconstructions represented as point clouds. 
%The main problem we address is to merge these two reconstructions by computing the scale, rotation and translation to align them. It is  formalized as follows:

%At this stage, given two sets of input images $\{{\presuper{\mathcal{F}}{\mathcal{I}}}_i,{\presuper{\mathcal{B}}{\mathcal{I}}}_j\}$ from the front and back sides of a row, two corresponding reconstructions $ \{{\mathcal{F}}, {\mathcal{B}}\}$ along with extrinsic camera poses $\{{\presuper{\mathcal{F}}{\mathbf{T}}}_i,{\presuper{\mathcal{B}}{\mathbf{T}}}_j\}$  where $ i = 1, \ldots, m$ and $j = 1,\ldots, n$ of each image, the goal is to merge these two reconstructions into a single coherent model $\presuper{}{\mathcal{P}}$ with combined set of camera poses $\{\mathbf{T}\}$ by finding a transformation  $^{\mathcal{F}}_{\mathcal{B}} \mathbf{T} = s [ ^{\mathcal{F}}_{\mathcal{B}} \mathbf{R} | ^{\mathcal{F}}_{\mathcal{B}} \mathbf{t} ]$ that merges the back side reconstruction with the front (see Fig.~\ref{fig:intro_goal}).

%Next, we introduce the objective function used for optimization.

\begin{figure}[t]
	\centering
	\includegraphics[width=0.75\columnwidth]{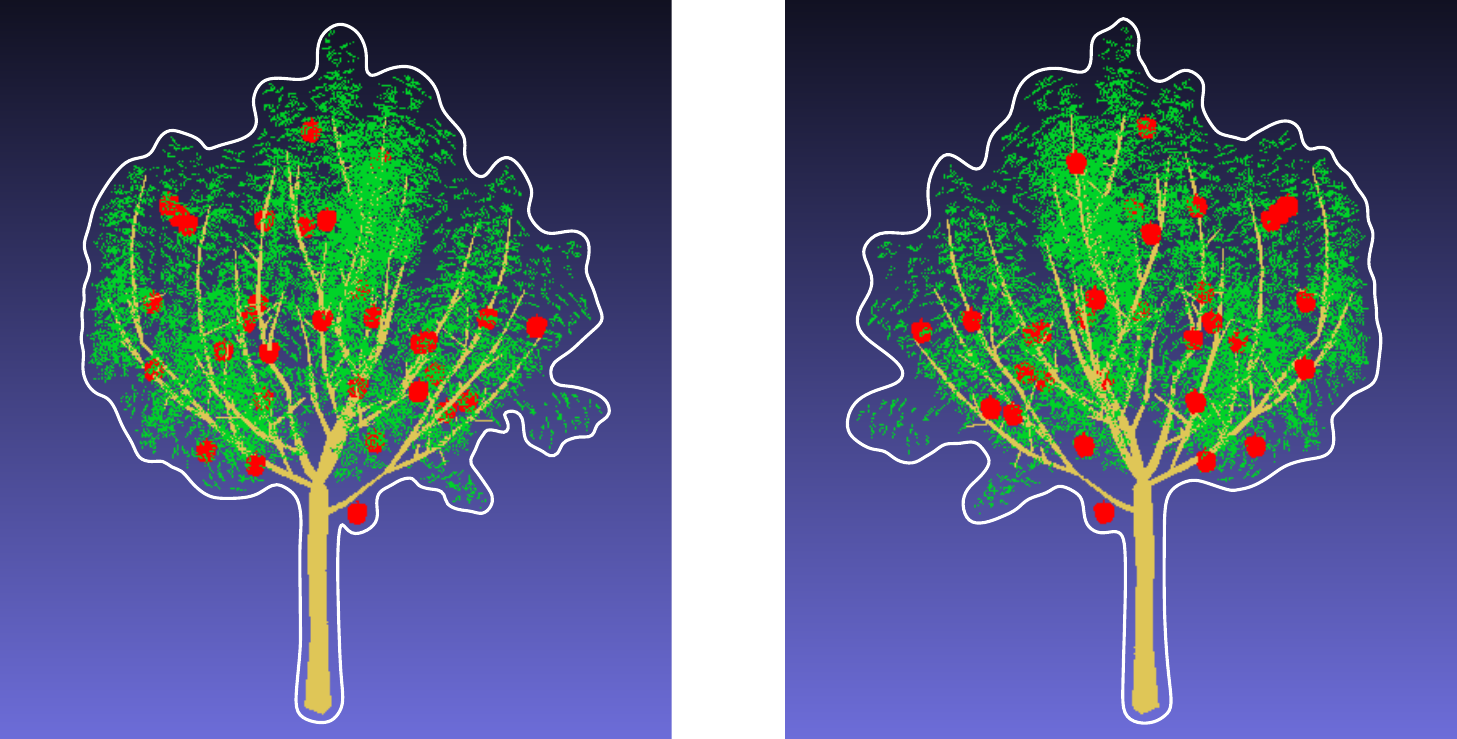}
	\caption{The silhouettes of an apple tree from two sides. As the silhouettes are from the orthographic front and back view of the tree, they should align.}
	\label{fig:Tree_Boundary}
\end{figure}

Given two reconstructions from the front and back sides of a row, we aim to align these two sets of 3D point clouds into a single coherent model by finding an initial transformation $^{\mathcal{F}}_{\mathcal{B}} \mathbf{T}_{\text{in}} = s_{\text{in}} [ ^{\mathcal{F}}_{\mathcal{B}} \mathbf{R}_{\text{in}} | ^{\mathcal{F}}_{\mathcal{B}} \mathbf{t}_{\text{in}} ]$ (i.e., scale, rotation and translation).
%\textbf{Objective Function:}
The front and back side reconstructions ${^{\mathcal{F}}\mathcal{P}}$,  ${^{\mathcal{B}}\mathcal{P}}$ generally do not share any local feature matches (point correspondences).
To constrain the system, we propose to use global features based on the following observations: (1)~the occlusion boundary of an object from the front and back orthographic views are the same (see Fig.~\ref{fig:Tree_Boundary}); (2)~tree trunk segments at the same height from two sides can be treated approximately as cylinders. When projected to the ground plane, they share the same center of the elliptical shape. If we align the median plane of the detected trunks, the maximum depth alignment error is bounded by the trunk widths.

We use the following geometric concepts to formulate the optimization problem.
An orthographic front view is the parallel projection of all the points in the $XY$-plane. The occlusion boundaries of this view can be approximated by the well-known concept of alpha shapes~\cite{edelsbrunner1994three}. An alpha hull is the generalized version of the convex hull. The boundaries of an alpha hull $B_{\alpha}$ are point pairs that can be touched by an empty disc of radius alpha. 
In most orchard settings, there are no leaves or branches attached to the trunks near the ground plane, and the number of 3D points in this region is very sparse. We can detect and cluster the 3D points belonging to tree trunks (see Sec.~\ref{subsubsec:trunkDetection} for trunk detection).
Let $^{\mathcal{F}}\mathcal{P}_{\text{t}}$, $^{\mathcal{B}}\mathcal{P}_{\text{t}}$ denote the detected trunk points close to the median trunk plane. ${\mathbf{P}_{XY}}$ and ${\mathbf{P}_{ZX}}$ denote the orthogonal projection matrix to the front and top plane. With this we can define our problem as the following minimization problem:
\begin{equation} \label{eq:base}
\argmin\limits_{{s_{\text{in}}}, {^{\mathcal{F}}_{\mathcal{B}} \mathbf{R}_{\text{in}}}, {^{\mathcal{F}}_{\mathcal{B}} \mathbf{t}_{\text{in}}}} D_{\text{s}}\left(B_{\alpha}\left({\mathbf{P}_{XY}} \cdot {^{\mathcal{F}}\mathcal{P}}\right), ~B_{\alpha}\left({\mathbf{P}_{XY}} \cdot {^{\mathcal{F}}_{\mathcal{B}} \mathbf{T}_{\text{in}}} \cdot {^{\mathcal{B}}\mathcal{P}}\right)\right)
+ D_{\text{s}}\left( {\mathbf{P}_{ZX}} \cdot {^{\mathcal{F}}\mathcal{P}_{\text{t}}}, ~{\mathbf{P}_{ZX}} \cdot {^{\mathcal{F}}_{\mathcal{B}} \mathbf{T}_{\text{in}}} \cdot {^{\mathcal{B}}\mathcal{P}_{\text{t}}} \right) ,
\end{equation}
where $B_{\alpha}$ computes the alpha shape boundary points. To find similarity between two point sets $\mathcal{P},\mathcal{Q}$ in ${\rm I\!R}^d$, we use the following metric:
\begin{equation}
D_{\text{s}}\left(\mathcal{P},\mathcal{Q}\right) = \sum_{\forall p \in \mathcal{P}} \min_{\forall q \in \mathcal{Q}} {\left(p-q\right)}^2 .
\label{eq:dist}
\end{equation}

%\textbf{Technical Approach:}
We can solve Eq.~\eqref{eq:base} using trusted region methods such as the Levenberg-Marquardt (LM) algorithm~\cite{levenberg1944method,marquardt1963algorithm} for which we need to find a good initial solution. Trivial initial values such as $\mathbf{I}_{3\times 3}$ and zero translation $\mathbf{0}_{3\times 1}$ do not work. 
In this paper, we develop a method to find a good initial solution. Essentially, we solve the two parts on the right-hand side of Eq.~\eqref{eq:base} sequentially along with some preliminary steps. We perform the following steps:
\begin{enumerate}
%\item For both ${\mathcal{F}}, {\mathcal{B}}$, we detect the ground plane.
\item We detect the ground plane from both ${\mathcal{F}}$, ${\mathcal{B}}$, and perform a Principal Component Analysis (PCA) on both reconstructions. Utilizing the ground plane normal and camera poses, we find the depth and up direction of the PCA components and align the point clouds roughly. To eliminate the rotational component left after PCA, we perform alignment of the ground plane normals. We fix the scale of the reconstructions using median scene height and fix the height of the ground plane using the median height of the ground plane inliers.
%\item To get rid of the rotational component left after PCA, we perform alignment of the ground plane normals.
%\item We fix the scale of the reconstructions using median scene height and fix the height of the ground plane using the median height of the ground plane inliers.
\item Now, we only have to compute the translations. We use alpha volume analysis to compute the occlusion boundary and 2D shape matching techniques~\cite{myronenko2010point} to compute the translation in the $XY$-plane.
\item Next, we align the trunk points  close to the median trunk plane.
\end{enumerate}

After these steps, the point clouds are roughly aligned and the trivial initial solution $s_{\text{in}} = 1$, ${^{\mathcal{F}}_{\mathcal{B}}\mathbf{R}_{\text{in}}} = \mathbf{I}_{3\times 3}$, ${^{\mathcal{F}}_{\mathcal{B}}\mathbf{t}_{\text{in}}} = \mathbf{0}_{3\times 1}$ leads to convergence. This method does not compute the trunk overlap distance precisely. However, it provides the correspondence between the trunks from both sides for trunk modeling (see Sec.~\ref{subsec:trunkGround}).
%Using the Random Sample Consensus (RANSAC) scheme~\cite{fischler1981random}, the trunk and the ground area around each tree are modeled as a cylinder and a plane, respectively. This semantic information, i.e., trunks and ground areas, can be exploited into the bundle adjustment to further eliminate misalignment of two-sides reconstructions by adjusting camera poses and 3D information of semantic objects and feature points.
Each of these steps is explained in detail as follows. 
%We start with the initial step of PCA and ground plane alignment.

\subsubsection{Ground Plane Estimation and Alignment using PCA} \label{sec:pca}
The main goal of this step is to eliminate most of the rotational difference required to align the two reconstructions. As is well known, this is normally solved by PCA. We assume that the length of the portion of the row covered by the input reconstruction is always longer than the height of the trees and the depth captured. Therefore, the first principal component always denotes the length of the row covered. The other two principal components vary from reconstruction to reconstruction. Therefore, while aligning the principal components, we need to be aware of which component denotes scene depth and which one denotes height. 

\begin{figure}[t]
	\centering
	\includegraphics[width=0.99\columnwidth]{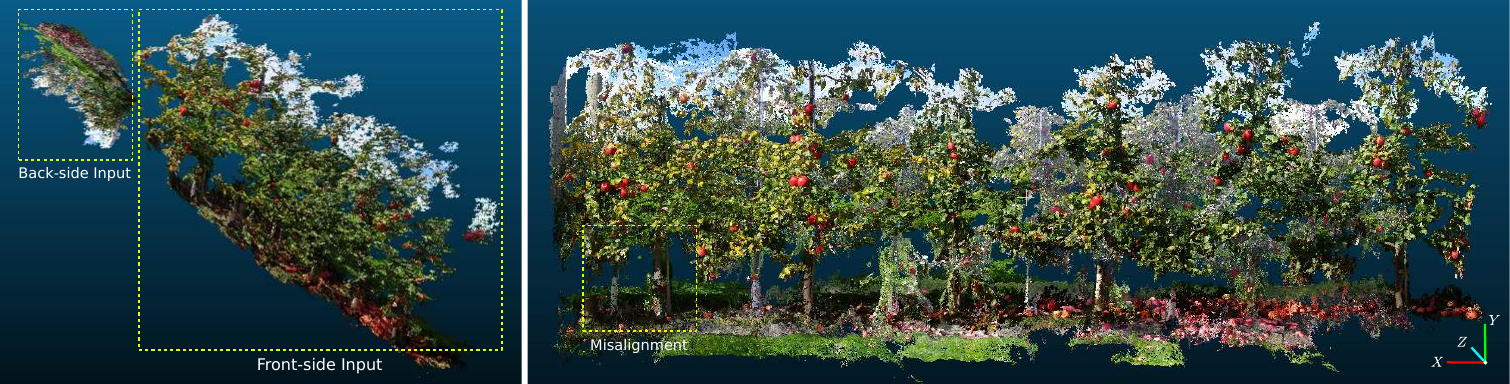}
	\caption{Input and PCA standardization. The left image shows the input to our algorithm. The right image shows the reconstructions after PCA-based alignment, ground plane alignment, and scale and height adjustment. In the input, we have only six trees but in the right image, we see more than six trees. This shows that, even though most of the rotational differences are gone, the alignment is wrong in terms of translation in the $X$-direction.}
	\label{fig:inputpca}
\end{figure}

To automatically figure out the scene ``up'' and ``depth'' directions, we estimate the ground plane.
We perform a simple three-point RANSAC method~\cite{fischler1981random} for plane estimation. Then, we align the corresponding principal components. If necessary, we flip the depth direction of one of the reconstructions to ensure that the frontal depth planes are opposing each other. Subsequently, we align the ground plane normals and rotate the point clouds to a canonical frame of reference $\{X,Y,Z\}$ where $X = [1, 0, 0]^{\top}$, $Y = [0, 1, 0]^{\top}$ (up direction) and $Z = [0, 0, 1]^{\top}$ (depth direction). Next, we fix the scale of the reconstructions using the median height of the trees (We assume that the two reconstructions correspond to roughly the same section of the row. Otherwise, the median can be off.), and fix the height of the ground plane using inliers. The reconstructions are now roughly aligned in terms of rotation and translation in the $Y$-direction. Fig.~\ref{fig:inputpca} shows the result after these steps in a sample input reconstruction.

\subsubsection{Alignment of Orthographic Projection Boundaries} \label{sec:projsymmetry}
In the last section, we roughly aligned the two reconstructions in terms of rotation, scale, and translation with respect to the ground plane. Now we have to estimate translation in the canonical directions $X = [1,0,0]^{\top}$ and $Z = [0, 0, 1]^{\top}$. We start with solving for the translation in the $X$-direction. In practice, reconstructions are not perfect, and the ground plane is not perfectly planar. Consequently, our estimation in the previous step contains some error in terms of rotation, scale, and translation in $X$- and $Y$-directions. We use a method that computes this residual rotation, translation, and scaling along with the translation in the $X$-direction.

\begin{figure}[t]
	\centering
	\includegraphics[width=0.99\columnwidth]{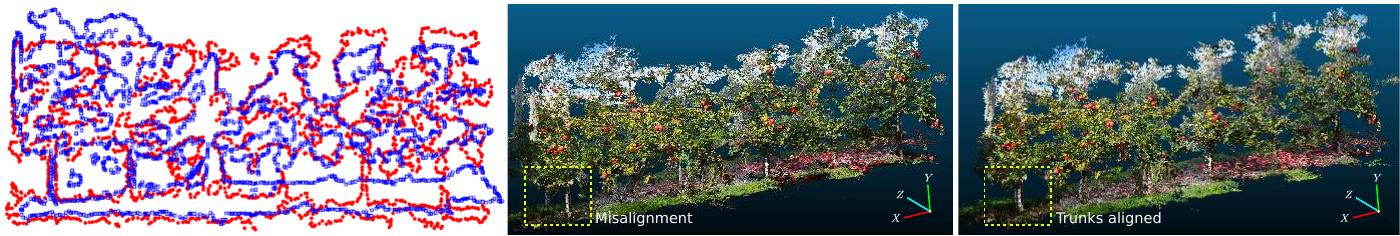}
	\caption{Alignment based on occlusion boundary and trunks. The left image shows how we align the occlusion boundaries of the reconstructions. The reconstructions after this step are shown in the middle image which demonstrates some alignment error in the depth direction (trunks and poles do not merge). The right image shows the final merged reconstruction using trunk information which does not have any duplicated trunks or poles.}
	\label{fig:occltrunk}
\end{figure}

As outlined in Sec.~\ref{subsec:globalign}, to solve this we utilize the occlusion boundary of the reconstructions from orthographic front views. We use alpha volume analysis to compute the occlusion boundaries. The alpha hull boundaries are basically a set of 2D points. Thus, essentially we are solving a 2D point set registration problem. This problem is very well studied and many solutions exist in the literature. As our alpha boundaries are noisy we use a well-known shape alignment method, Coherent Point Drift (CPD) algorithm~\cite{myronenko2010point}. Myronenko et al.~\cite{myronenko2010point} cast the point set registration problem as a probability density estimation problem. They represent one of the input point set as the centroids of a Gaussian Mixture Model (GMM) and the other input as data. For the rigid transformation case, they reparameterize the GMM centroids in terms of rotation and scale transformation. They estimate the parameters by minimizing the negative log likelihood using the Expectation-Maximization algorithm. Additionally, they add an extra component in the GMM to account for noise and outliers. At the optimum value of the parameters, two point sets are aligned. We apply the transformation computed by CPD to the entire point cloud to align them in the $XY$-directions. The left figure in Fig.~\ref{fig:occltrunk} shows the alignment of occlusion boundaries.

\subsubsection{Alignment in Depth Direction using Trunk Information} \label{subsubsec:trunkAlignment}
The principal ambiguity left is the relative depth distance between the two reconstructions. In an orchard row, trees are generally planted in straight lines and tree trunks are perpendicular to the ground. Therefore, we can imagine the existence of a central trunk plane bisecting the trunks. For each individual reconstruction, this bisector plane can be approximated by the median depth-plane of the detected trunks (see Sec.~\ref{subsubsec:trunkDetection}) and we can align the reconstructions roughly by aligning the points close to this median plane.
%
%\textbf{Trunk Detection:} In most orchard settings, there are no leaves/branches attached to the trunks near the ground plane. Consequently, the number of 3D points in the trunk region close to the ground is very small. We capture this region by binning the 3D points in terms of their distance from the ground plane (see Fig.~\ref{fig:trunkDetection}). The flat region in this curve shown in Fig.~\ref{fig:trunkDetection} is easily captured by utilizing the derivative. We use the region with the minimum number of points and two knee points around it to find the trunks close to the ground.
%
%\begin{figure}[thbp]
%        \centering
%            \includegraphics[width =0.80\columnwidth]{figs/Wenbo/trunk_detection.pdf}           
%   \caption{Trunk detection. Figure on the left shows the detected trunks on the simulated trees. Figure on the right shows a plot of the number of 3D points plotted against their distance from the ground plane. The flat region in this plot correspond to the 3D points belonging to the trunk regions.}
%   \label{fig:trunkDetection}
%\end{figure}
%
We perform a simple median filtering of the segmented trunk points $^{\mathcal{F}}\mathcal{P}_{\text{t}}$, $^{\mathcal{B}}\mathcal{P}_{\text{t}}$. Then, like in the previous section, we can align them using the CPD method.
%
%After this step, both reconstructions are in the same frame of reference and have a very small difference in terms of rotation, translation, and scale. The trivial initial solution of $s = 1,{^{\mathcal{F}}_{\mathcal{B}} \mathbf{R}} = \mathbf{I}_{3\times 3}, {^{\mathcal{F}}_{\mathcal{B}} \mathbf{t}} = \mathbf{0}_{3\times 1}$ leads to fast convergence. 
The right figure in Fig.~\ref{fig:occltrunk} shows the merged reconstruction.
%As a byproduct from this process, we have established trunk to trunk correspondences between the two reconstructions. We can use this to optimize our alignment further. We discuss the details of this procedure in the next section.
%
%

\subsection{Trunk Modeling and Local Ground Estimation} \label{subsec:trunkGround}
Accurate geometry estimation relies on good depth maps. Raw depth maps are usually noisy, especially in orchard environments. The large uncertainty of depth values around frequent occlusions between trees and leaves causes generated 3D points floating in the air~\cite{sotoodeh2006outlier}.
For RGB data, the depth map of each frame is generated from a single-side dense reconstruction using the commercial software Agisoft~\cite{agisoft2017agisoft}.
For RGB-D data, we first improve the depth map using the Truncated Signed Distance Function (TSDF)~\cite{curless1996volumetric} to accumulate depth values from nearby frames (e.g., 10 closest frames) with the camera poses obtained in Sec.~\ref{subsec:singleSideBA}. The pixel value of the raw depth is ignored if it is largely different from the corresponding value in the fused depth obtained by ray casting. A floating pixel removal filter~\cite{sotoodeh2006outlier} is further applied to eliminate any pixel of the raw depth that has no nearby 3D points within a certain distance threshold.

\subsubsection{Trunk Detection} \label{subsubsec:trunkDetection}
\begin{figure}[t]
	\centering
	\includegraphics[width=0.95\columnwidth]{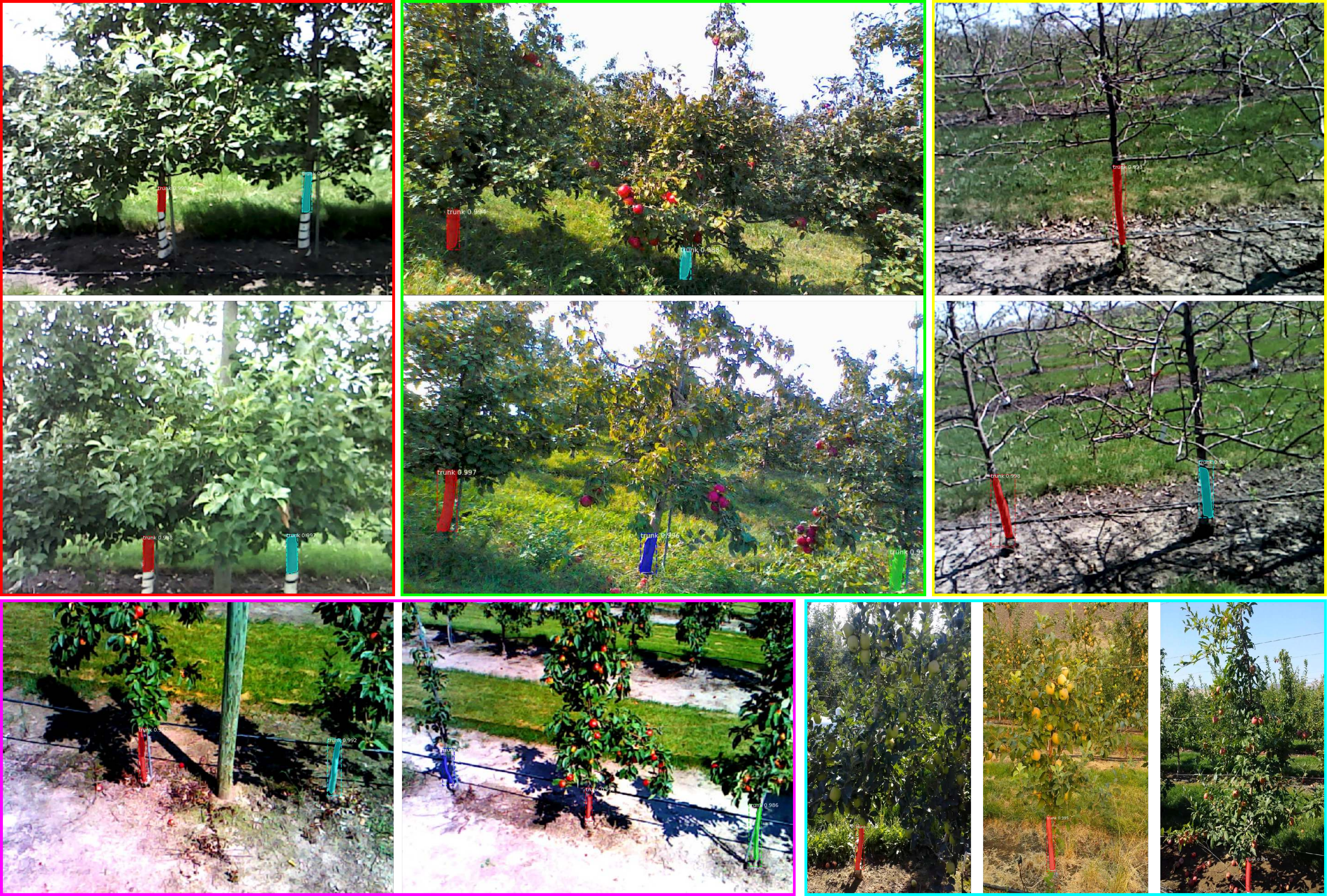}
	\caption{Trunk detection results on test orchards. Images are separated by five color boxes that correspond to five test orchards.}
	\label{fig:trunkDetection}
\end{figure}

In orchard settings, one of the challenges in establishing visual correspondences is identifying good features to track.
Traditional geometric features, such as corners and lines, are highly ambiguous and rarely visible from two sides of a tree row.
Instead, tree trunks are easily captured from both sides, and provide robust features to establish correspondences~\cite{roy2018registering}.
Detecting tree trunks is also beneficial because horticulture scientists use trunk diameter as a phenotypic trait~\cite{dong2018tree}. To detect and extract trunk information (see Fig.~\ref{fig:trunkDetection}), we perform segmentation using Mask R-CNN~\cite{he2017mask}.

Mask R-CNN is the state-of-the-art algorithm for instance segmentation. Its model decouples class prediction and mask generation, and has three outputs for each object instance: a class label, a bounding box and a mask extracting a fine spatial layout. The whole image as the input is taken by the Region Proposal Network (RPN) which outputs bounding box proposals. Each proposal is supposed to be an object. Based on the bounding box, the Fully Convolutional Network (FCN) performs an object segmentation. The Mask R-CNN model we used is based on an open-source implementation by Matterport\footnote{\url{https://github.com/matterport/Mask_RCNN}}, built on Feature Pyramid Network (FPN) and ResNet-101~\cite{he2016deep} as the backbone.

\textbf{Datasets:} 1000 images (around 2000 instances annotated using the VGG Image Annotator\footnote{\url{http://www.robots.ox.ac.uk/~vgg/software/via}}) are selected from five orchards that have different environment settings. We split all images into five different datasets (including training and evaluation) based on the test orchard ID. For example, to test the trunk detector on orchard 1, we build the training data using 90\% of images from orchards 2 to 5 and 50\% of images from orchard 1. Therefore, in the whole training data, the percentage of the data from orchard 1 is only around 10\%$\sim$20\%. We leave the rest as the data for evaluation. Each dataset simulates the case that when the orchard environment is changed, we need to tune the trunk detector by adding only a small portion of new images to the training pool.

\begin{figure}[t]
	\centering
	\begin{subfigure}{0.45\textwidth}
		\centering
		\includegraphics[width=1\linewidth]{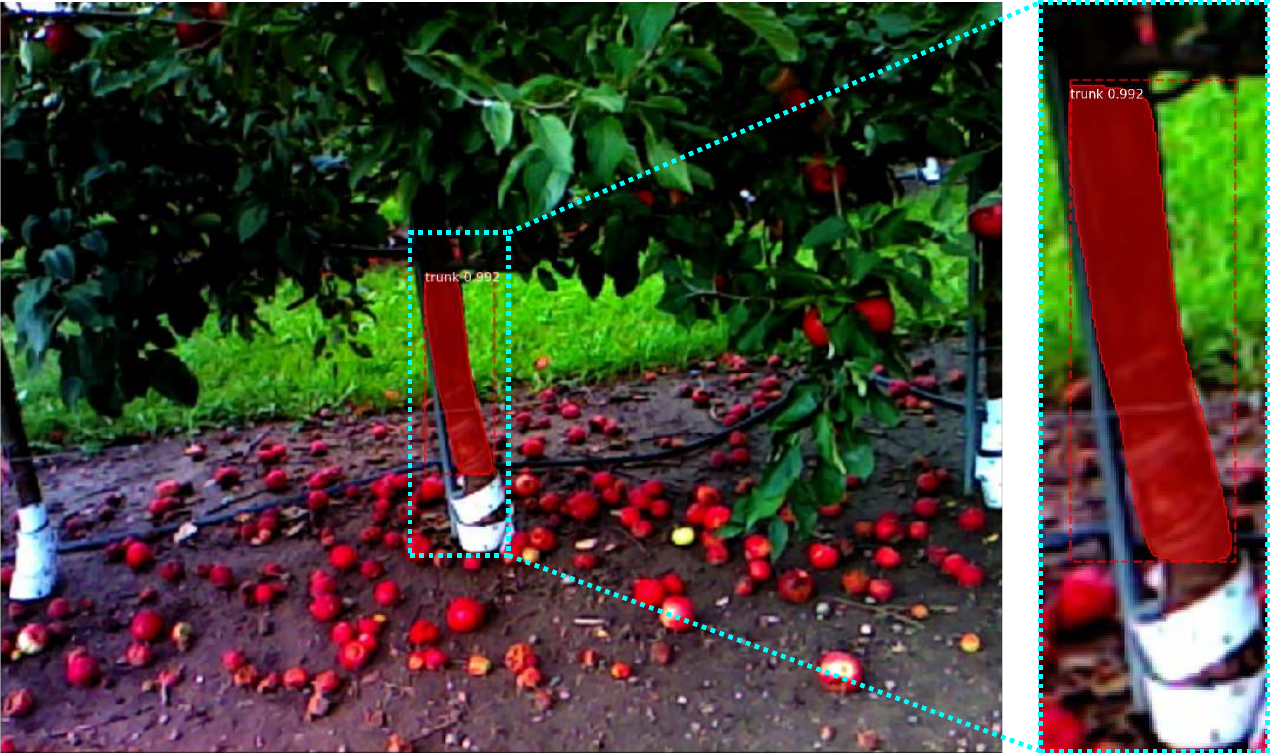}
		\caption{Front side}
	\end{subfigure}
	~
	\begin{subfigure}{0.45\textwidth}
		\centering
		\includegraphics[width=1\linewidth]{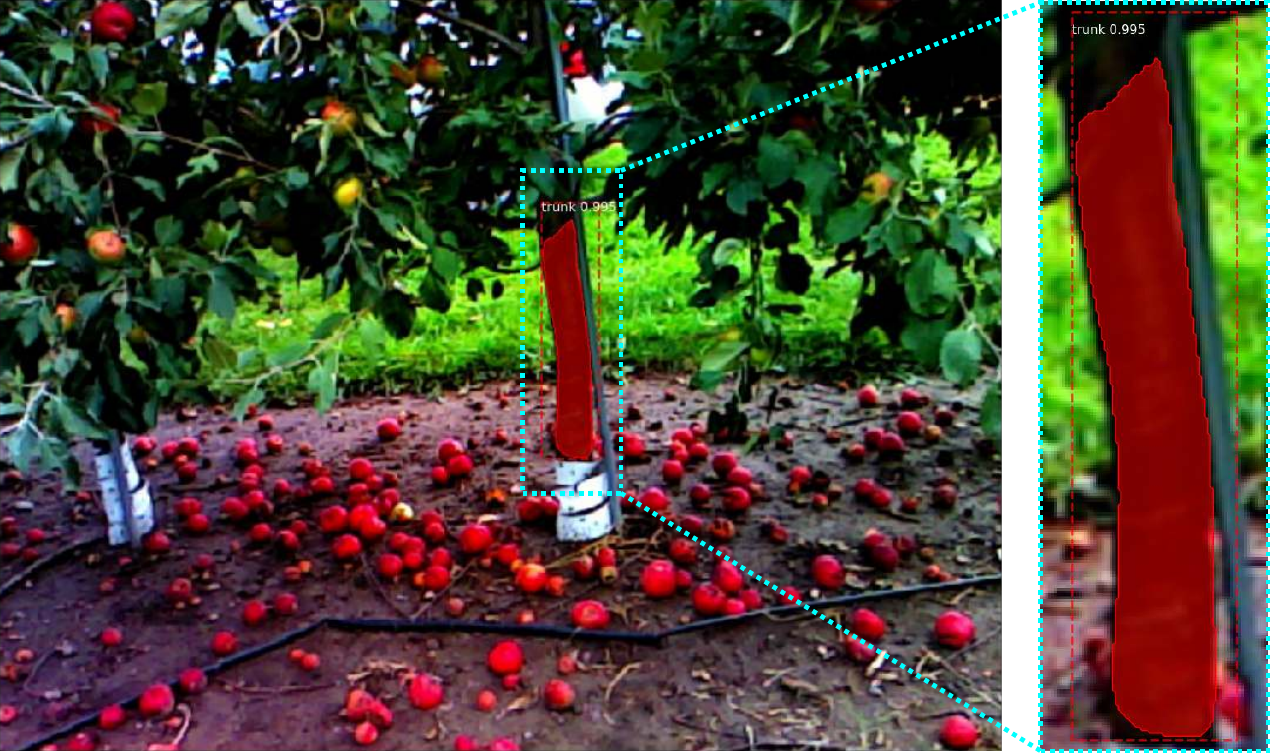}
		\caption{Back side}
	\end{subfigure}
	\caption{Zoom-in view of trunk detection for the same tree from two sides. Trunk boundaries along the trunk direction are clearly segmented.}
	\label{fig:zoomInTrunk}
\end{figure}

%\textbf{Evaluation and Results:}
%The Mask R-CNN model is first trained and evaluated on each test orchard (see Sec.~\ref{subsubsec:trunkDetectionResult}).
%In Table~\ref{table:detectionAcc}, we report the standard COCO metrics~\cite{lin2014microsoft} including AP (averaged over IoU thresholds), AP$_{50}$, and AP$_{75}$, where AP is evaluating using mask IoU (Intersection over Union). The high accuracy in AP$_{50}$ demonstrates the correctness of trunk prediction.
Fig.~\ref{fig:trunkDetection} presents qualitative segmentation results of five orchards.
%The model is finally trained using all the training images from five datasets based on the above-mentioned strategy.
The trunk detector outputs clear boundaries along the trunk direction (see Fig.~\ref{fig:zoomInTrunk}), which is crucial to the next step (see Sec.~\ref{subsubsec:trunk}). The training strategy and model evaluation details are explained in Sec.~\ref{subsubsec:trunkDetectionResult}.

\subsubsection{Trunk Cylinder Modeling} \label{subsubsec:trunk}
\begin{figure}[t]
	\centering
	\begin{subfigure}{0.27\textwidth}
		\centering
		\includegraphics[width=1.0\linewidth]{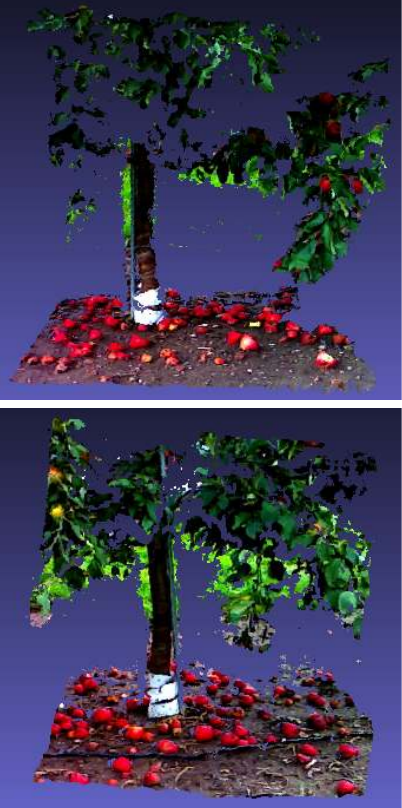}
		\caption{3D tree model}
	\end{subfigure}
	\qquad
	\begin{subfigure}{0.27\textwidth}
		\centering
		\includegraphics[width=1.0\linewidth]{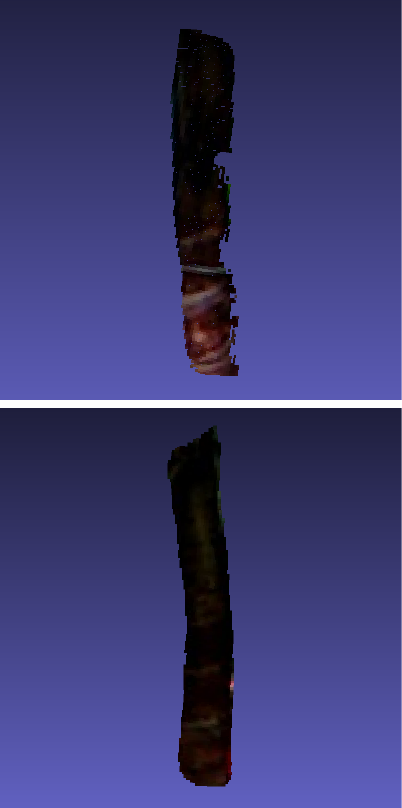}
		\caption{Trunk segmentation}
	\end{subfigure}
	\qquad
	\begin{subfigure}{0.27\textwidth}
		\centering
		\includegraphics[width=1.0\linewidth]{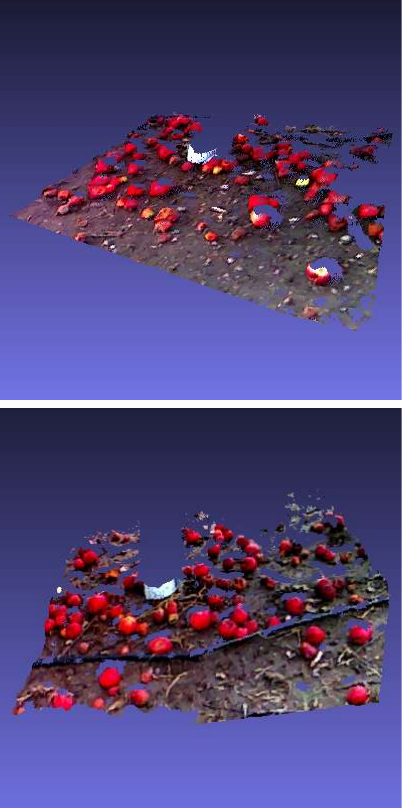}
		\caption{Local ground segmentation}
	\end{subfigure}
	\caption{Front-side (top) and back-side (bottom) volumetric fusion using nearby frames. The whole trunk is segmented by clear boundaries from trunk detection to remove outliers from the 3D model. The local ground is segmented using the prior knowledge of the trunk cylinder.}
	\label{fig:volumetricTrunkGround}
\end{figure}

\begin{figure}[t]
	\centering
	\includegraphics[width=0.8\columnwidth]{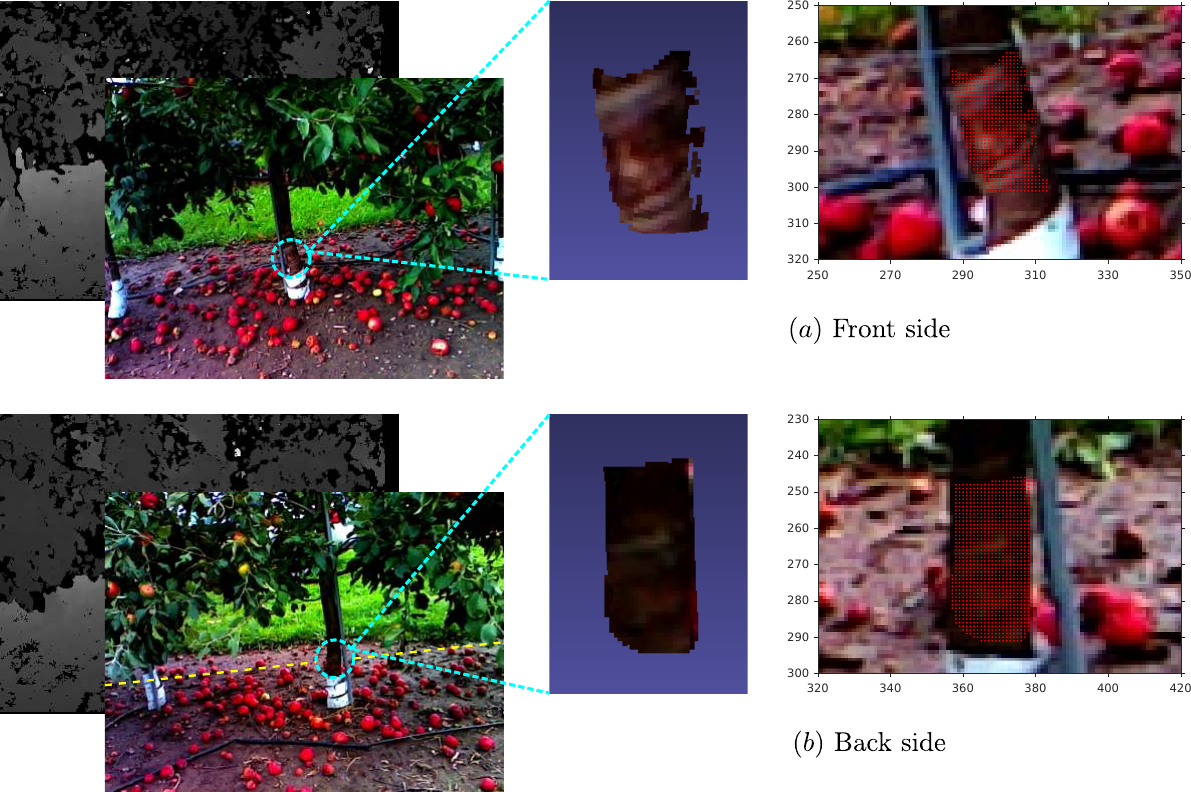}
	\caption{Cylindrical trunk segments that are most-detected by nearby frames from two sides. The height of trunk segments (yellow line) is determined by estimating ground planes from two sides. The depth pixels (red points) are selected by projecting the cylindrical segment onto the image.}
	\label{fig:trunkSegmentation}
\end{figure}

3D dense models of a tree from two sides (front and back sides) are obtained using the volumetric fusion of depth maps from all nearby frames (see Fig.~\ref{fig:volumetricTrunkGround}). The dense point cloud of the trunk is first segmented by taking the union of trunk detection masks from all frames. 
The split-and-merge approach~\cite{medeiros2017modeling} is implemented to divide the trunk point cloud into approximately cylindrical segments.
Given the ground plane as the reference (see Sec.~\ref{subsubsec:plane} for local ground estimation), the trunk segments within a height interval from two sides share the same cylinder model (see Sec.~\ref{subsubsec:trunkAlignment} for two-sides trunks association).
We select the segment that is most-detected by nearby frames (see Fig.~\ref{fig:trunkSegmentation}).
For each frame $c$ with trunk detection, the cylindrical segment is projected onto the image to select valid depth pixels within the trunk mask. We aim to fit the 3D depth points to a cylinder $d$ parameterized by its axis $^c\mathbf{n}_d$, center (origin) $^c\mathbf{O}_d$ and radius $^cr_d$. The height $^ch_d$ of the cylinder is determined by the bounding box of the 3D points along $^c\mathbf{n}_d$.

\begin{figure}[t]
	\centering
	\begin{subfigure}{0.38\textwidth}
		\centering
		\includegraphics[width=1\linewidth]{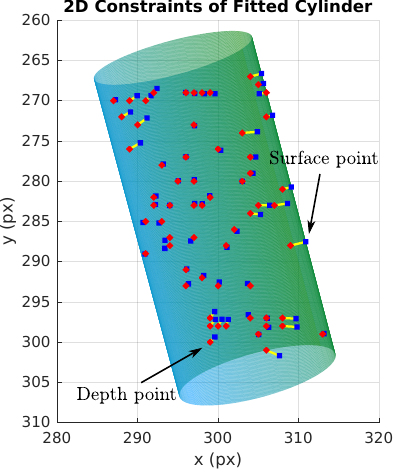}
		\caption{Trunk cylinder modeling}
	\end{subfigure}
	\qquad
	\begin{subfigure}{0.38\textwidth}
		\centering
		\includegraphics[width=1\linewidth]{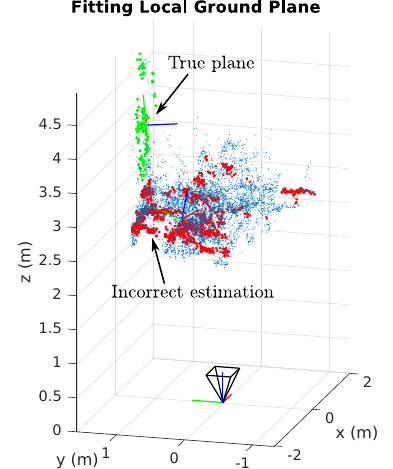}
		\caption{Local ground plane estimation}
	\end{subfigure}
	\caption{Trunk cylinder modeling and local ground plane estimation in a single frame. (a): 3D depth (red) and surface points (blue) are projected on the image (the points are subsampled and best viewed in color). The yellow line characterizes the reprojection error. (b): Without trunk information, standard plane estimation outputs a wrong plane (red), while the true ground (green) is estimated using the proposed algorithm.}
	\label{fig:cylinderPlane}
\end{figure}

An accurate cylinder model should not only fit the most 3D depth points but also obtain a reasonable size and pose from the image. To robustly model the cylinder, we integrate 2D constraints into a RANSAC scheme~\cite{fischler1981random} with the nine-point algorithm~\cite{beder2006direct}.
Specifically, surface points obtained by projecting 3D depth points onto the cylinder should have small 2D reprojection error with corresponding depth pixels on the image (see Fig.~\ref{fig:cylinderPlane}).
The trunk cylinder in frame $c$ is further optimized by minimizing the cost function
\begin{equation} \label{objectCylinder}
\begin{gathered}
\argmin\limits_{^c\mathbf{n}_d, ^c\mathbf{O}_d, ^cr_d} \sum\limits_{p} \rho\left(e_d^2(^c\mathbf{X}_p, d)\right) + \lambda \sum\limits_{p} \rho\left(\| ^c\mathbf{x}_p - ^d\mathbf{x}_p \|^2\right) \\
^d\mathbf{x}_p = p_r(^c\mathbf{X}_p, c, d)
\end{gathered} ,
\end{equation}
where $e_d$ is the distance function of a 3D point $^c\mathbf{X}_p$ to the cylinder, and $^c\mathbf{x}_p$ is the 2D depth pixel corresponding to $^c\mathbf{X}_p$.
The function $p_r$ projects $^c\mathbf{X}_p$ onto the cylinder surface, and further projects the surface point on the image frame $c$ to obtain $^d\mathbf{x}_p$.
The Huber loss $\rho$~\cite{huber1992robust} is applied to the quadratic errors for robustness.
The trunk in frame $c$ is thus described by the cylinder axis $^c\mathbf{n}_d$ and the origin $^c\mathbf{O}_d$.

\subsubsection{Local Ground Plane} \label{subsubsec:plane}
Without loss of generality, the local ground of a tree is assumed as a plane defined by its normal $^c\mathbf{n}_p$ and center (origin) $^c\mathbf{O}_p$ in frame $c$. Unlike trunk detection, only frame number is recorded for plane estimation. However, it is not always the case that the majority of 3D points are from the ground, which highly depends on the scene and the camera view. The standard RANSAC-based method fails to detect the ground plane (see Fig.~\ref{fig:cylinderPlane}b). We modify the degenerate condition of the RANSAC by using the prior information of the trunk axis $^c\mathbf{n}_d$ transformed from the closest frame with trunk detection: $^c\mathbf{n}_p$ should roughly align with $^c\mathbf{n}_d$, and the estimated plane should be on the boundary of all 3D points along $^c\mathbf{n}_p$ within the distance threshold $t_s$. The local ground in frame $c$ is thus defined by the plane normal $^c\mathbf{n}_p$ and the origin $^c\mathbf{O}_p$. Local ground estimation from two sides can further help cylindrical trunk segmentation (see Fig.~\ref{fig:trunkSegmentation}).

\subsection{Merging Two-Sides 3D Reconstruction} \label{subsec:merging}
For a tree row, the front-side and back-side reconstructions are expressed in their own frames ${\mathcal{F}}$ and ${\mathcal{B}}$, respectively. The goal is to first align two-sides reconstructions by estimating the initial rigid transformation $[^\mathcal{F}_\mathcal{B}\mathbf{R} | ^\mathcal{F}_\mathcal{B}\mathbf{t}]$, and further optimize the 3D reconstruction based on semantic information.

\subsubsection{Initial Transformation}
From a geometric view, to align the 3D models of a tree row from both sides, at least two detected trunks and one estimated local ground are required. 3D models are first constrained on the local ground plane. The translation and rotation along the ground plane are further constrained by two trunk-cylinders. Multiple trunks and local grounds can provide us a robust solution. In Sec.~\ref{subsec:trunkGround}, an $i$-th detected trunk from two-sides detection views is described by its cylinder axes $^{\mathcal{F}}\mathbf{n}_d^i$ and $^{\mathcal{B}}\mathbf{n}_d^i$ with a unit length, and its origins $^{\mathcal{F}}\mathbf{O}_d^i$ and $^{\mathcal{B}}\mathbf{O}_d^i$. Similarly, a $j$-th estimated local ground is described by its plane normals $^{\mathcal{F}}\mathbf{n}_p^j$ and $^{\mathcal{B}}\mathbf{n}_p^j$, and its origins $^{\mathcal{F}}\mathbf{O}_p^j$ and $^{\mathcal{B}}\mathbf{O}_p^j$.

First, cylinder axes and plane normals in ${\mathcal{B}}$ after the relative transformation must be equal to their corresponding ones in ${\mathcal{F}}$. Then, the first two constraints have the form
\begin{equation} \label{constraint1}
\begin{cases}
^\mathcal{F}_\mathcal{B}\mathbf{R} \cdot ^{\mathcal{B}}\mathbf{n}_d^i = ^{\mathcal{F}}\mathbf{n}_d^i \\
^\mathcal{F}_\mathcal{B}\mathbf{R} \cdot ^{\mathcal{B}}\mathbf{n}_p^j = ^{\mathcal{F}}\mathbf{n}_p^j
\end{cases} .
\end{equation}
Second, the origins of cylinders in ${\mathcal{B}}$ transformed to ${\mathcal{F}}$ should lie on the same axis-line. Then, the cross product between the cylinder axis and the difference of two-sides origins should be a zero vector
\begin{equation} \label{constraint2}
^{\mathcal{F}}\mathbf{n}_d^i \times \left( ^\mathcal{F}_\mathcal{B}\mathbf{R} \cdot ^{\mathcal{B}}\mathbf{O}_d^i + ^\mathcal{F}_\mathcal{B}\mathbf{t} - ^{\mathcal{F}}\mathbf{O}_d^i \right) = \mathbf{0} .
\end{equation}
At last, the origins of local planes in ${\mathcal{B}}$ after the transformation to ${\mathcal{F}}$ must lie on  the same plane. Thus, the dot product between the plane normal and the difference of two-sides origins should be zero
\begin{equation} \label{constraint3}
^{\mathcal{F}}\mathbf{n}_p^j \cdot \left( ^\mathcal{F}_\mathcal{B}\mathbf{R} \cdot ^{\mathcal{B}}\mathbf{O}_p^j + ^\mathcal{F}_\mathcal{B}\mathbf{t} - ^{\mathcal{F}}\mathbf{O}_p^j \right) = 0 .
\end{equation}

Following the order of the constraints above, Eqs.~(\ref{constraint1})-(\ref{constraint3}) can be rearranged into a system $\mathbf{A} \mathbf{x} = \mathbf{b}$ by treating each element of $[^\mathcal{F}_\mathcal{B}\mathbf{R} | ^\mathcal{F}_\mathcal{B}\mathbf{t}]$ as unknowns, where $^{\mathcal{B}}\mathbf{n}_d^i = [n^d_1, n^d_2, n^d_3]^{\top}$, $^{\mathcal{F}}\mathbf{n}_d^i = [{n^{\prime}}^d_1, {n^{\prime}}^d_2, {n^{\prime}}^d_3]^{\top}$, $^{\mathcal{B}}\mathbf{n}_p^j = [n^p_1, n^p_2, n^p_3]^{\top}$, and $^{\mathcal{F}}\mathbf{n}_p^j = [{n^{\prime}}^p_1, {n^{\prime}}^p_2, {n^{\prime}}^p_3]^{\top}$ for the axes, and the elements of the origins have a similar form. Here, the matrix $\mathbf{A}$ and vector $\mathbf{b}$ are
\begin{equation}
\begin{gathered}
\scalebox{0.92}
{$
\begin{bmatrix}
n^d_1 & 0 & 0 & n^d_2 & 0 & 0 & n^d_3 & 0 & 0 & 0 & 0 & 0 \\
0 & n^d_1 & 0 & 0 & n^d_2 & 0 & 0 & n^d_3 & 0 & 0 & 0 & 0 \\
0 & 0 & n^d_1 & 0 & 0 & n^d_2 & 0 & 0 & n^d_3 & 0 & 0 & 0 \\
n^p_1 & 0 & 0 & n^p_2 & 0 & 0 & n^p_3 & 0 & 0 & 0 & 0 & 0 \\
0 & n^p_1 & 0 & 0 & n^p_2 & 0 & 0 & n^p_3 & 0 & 0 & 0 & 0 \\
0 & 0 & n^p_1 & 0 & 0 & n^p_2 & 0 & 0 & n^p_3 & 0 & 0 & 0 \\
0 & -{n^{\prime}}^d_3 o^d_1 & {n^{\prime}}^d_2 o^d_1 & 0 & -{n^{\prime}}^d_3 o^d_2 & {n^{\prime}}^d_2 o^d_2 & 0 & -{n^{\prime}}^d_3 o^d_3 & {n^{\prime}}^d_2 o^d_3 & 0 & -{n^{\prime}}^d_3 & {n^{\prime}}^d_2 \\
{n^{\prime}}^d_3 o^d_1 & 0 & -{n^{\prime}}^d_1 o^d_1 & {n^{\prime}}^d_3 o^d_2 & 0 & -{n^{\prime}}^d_1 o^d_2 & {n^{\prime}}^d_3 o^d_3 & 0 & -{n^{\prime}}^d_1 o^d_3 & {n^{\prime}}^d_3 & 0 & -{n^{\prime}}^d_1 \\
-{n^{\prime}}^d_2 o^d_1 & {n^{\prime}}^d_1 o^d_1 & 0 & -{n^{\prime}}^d_2 o^d_2 & {n^{\prime}}^d_1 o^d_2 & 0 & -{n^{\prime}}^d_2 o^d_3 & {n^{\prime}}^d_1 o^d_3 & 0 & -{n^{\prime}}^d_2 & {n^{\prime}}^d_1 & 0 \\
{n^{\prime}}^p_1 o^p_1 & {n^{\prime}}^p_2 o^p_1 & {n^{\prime}}^p_3 o^p_1 & {n^{\prime}}^p_1 o^p_2 & {n^{\prime}}^p_2 o^p_2 & {n^{\prime}}^p_3 o^p_2 & {n^{\prime}}^p_1 o^p_3 & {n^{\prime}}^p_2 o^p_3 & {n^{\prime}}^p_3 o^p_3 & {n^{\prime}}^p_1 & {n^{\prime}}^p_2 & {n^{\prime}}^p_3
\end{bmatrix}
$} \\ 
\scalebox{0.92}
{$
\begin{bmatrix}
{n^{\prime}}^d_1 & {n^{\prime}}^d_2 & {n^{\prime}}^d_3 & {n^{\prime}}^p_1 & {n^{\prime}}^p_2 & {n^{\prime}}^p_3 & {n^{\prime}}^d_2{o^{\prime}}^d_3 - {n^{\prime}}^d_3{o^{\prime}}^d_2 & {n^{\prime}}^d_3{o^{\prime}}^d_1 - {n^{\prime}}^d_1{o^{\prime}}^d_3 & {n^{\prime}}^d_1{o^{\prime}}^d_2 - {n^{\prime}}^d_2{o^{\prime}}^d_1 & {n^{\prime}}^p_1{o^{\prime}}^p_1 + {n^{\prime}}^p_2{o^{\prime}}^p_2 + {n^{\prime}}^p_3{o^{\prime}}^p_3
\end{bmatrix}^{\top}
$}
\end{gathered} ,
\end{equation}
respectively, and $\mathbf{x} = [\mathbf{r}_1^{\top}, \mathbf{r}_2^{\top}, \mathbf{r}_3^{\top}, ^\mathcal{F}_\mathcal{B}\mathbf{t}^{\top}]^{\top}$ with $\mathbf{r}_1$, $\mathbf{r}_2$ and $\mathbf{r}_3$ being the three columns of $^\mathcal{F}_\mathcal{B}\mathbf{R}$.

We solve the system with multiple cylinders and planes for the least squares solution. The solution of $^\mathcal{F}_\mathcal{B}\mathbf{R}$ may not meet the properties of an orthonormal matrix, but can be computed to approximate a rotation matrix by minimizing the Frobenius norm of their difference~\cite{golub2012matrix}. An accurate initial value can be obtained from an analytical solution by using the resultant of polynomials~\cite{dong18novel}. With multiple pairs of cylinders and planes from both sides, we formulate an optimization problem (with the robust Huber loss $\rho$~\cite{huber1992robust} applied to the quadratic residuals)
\begin{equation}
\argmin\limits_{^\mathcal{F}_\mathcal{B}\mathbf{R}, ^\mathcal{F}_\mathcal{B}\mathbf{t}} \sum\limits_{i} \left( \rho\left(\|\mathbf{e}_1(i) \|^2\right) + \rho\left(\|\mathbf{e}_3(i) \|^2\right) \right) + \sum\limits_{j} \left( \rho\left(\|\mathbf{e}_2(j) \|^2\right) + \rho\left(e_4^2(j)\right) \right) ,
\end{equation}
where $\mathbf{e}_1$, $\mathbf{e}_2$, $\mathbf{e}_3$ and $e_4$ are residuals of Eqs.~(\ref{constraint1})-(\ref{constraint3}). The solution is further refined using the LM method with the rotation represented by the Rodrigues' formula~\cite{rodrigues1840lois}.

\subsubsection{Semantic Bundle Adjustment} \label{subsubsec:SBA}
\begin{figure}[t]
	\centering
	\includegraphics[width=0.8\columnwidth]{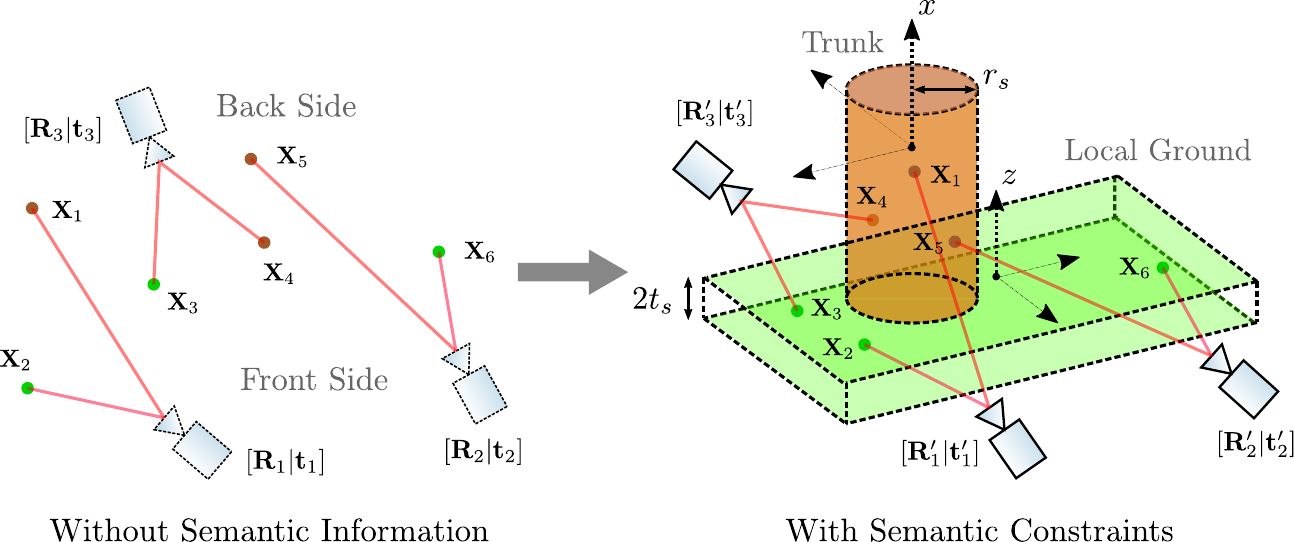}
	\caption{Scheme of semantic bundle adjustment. With semantic constraints, 3D points belonging to the same object are adjusted to fit onto the shape together with the camera poses corrected simultaneously.}
	\label{fig:semanticBA}
	%\vspace*{-4mm}
\end{figure}

To address the issue of accumulated errors of camera poses in Fig.~\ref{fig:loopTrajectory}c, the two-sides 3D reconstructions after initial alignment need to be further optimized. Intuitively, semantic information (i.e., trunks and local grounds) integrated in bundle adjustment will tune camera poses and 3D feature points until reasonable semantic conditions are reached. Specifically, two halves of a trunk from both sides should be well-aligned, and two-sides local grounds of a tree should refer to the same one (see Fig.~\ref{fig:semanticBA}).

Technically, a semantic object with index $s$ is characterized by its unique pose $[\mathbf{R}_s | \mathbf{t}_s ]$ in the world frame and its 3D shape $\mathbf{b}_s$. For a cylinder object, the shape is represented by its $x$-axis (as the cylinder axis), origin and a radius $r_s$. For a plane object, the shape is described by its $z$-axis (as the plane normal), origin and a threshold $t_s$ for bounding an interval along the plane normal. The cylinder radius $r_s$ and the plane-interval threshold $t_s$ are automatically determined by the fitting algorithms in Sec.~\ref{subsubsec:diameter} and Sec.~\ref{subsubsec:plane}, respectively. As a 3D feature point, the orientation $\mathbf{R}_s$ and the position $\mathbf{t}_s$ of an object are unknown and need to be estimated by semantic bundle adjustment.

Given the correspondences of objects between two sides, the objective function of semantic bundle adjustment is as follows
%\vspace*{-3mm}
\begin{equation} \label{objectSBA}
\begin{gathered}
\argmin_{\mathbf{R}_c, \mathbf{t}_c, \mathbf{R}_s, \mathbf{t}_s, \mathbf{X}_p} J^{\prime} =  J + \sum_{s} \sum_{c} \sum_{p \in \mathcal{V}(s,c)} \rho \left( \lambda_s E_b(s,c,p) \right) \\
E_b(s,c,p) = \phi_l \left( [\mathbf{R}_s | \mathbf{t}_s ] [\mathbf{R}_c | \mathbf{t}_c ]^{-1c} \bar{\mathbf{X}}_p, \mathbf{b}_s \right)^2
\end{gathered} ,
\end{equation}
where  $\phi_0$ ($l = 0$) is the loss function for a plane object $\phi_0(\mathbf{X}, \mathbf{b}_s)= \| \max \left( x_3-t_s, 0, -x_3-t_s \right) \|$, and $\phi_1$ ($l = 1$) is the loss function for a cylinder object $\phi_1(\mathbf{X}, \mathbf{b}_s) = \| \sqrt{x_2^2 + x_3^2} - r_s \|$, with an input 3D point $\mathbf{X} = [x_1, x_2, x_3]^{\top}$. The geometric meaning is that after transformation to the object frame, we penalize a 3D point belonging to a cylinder if it is far away from the cylinder surface. Similarly, a 3D point belonging to a plane is penalized if it is out of the boundary of the plane. The weight $\lambda_s$ balances between the cost $J$ of feature points and the cost of semantic object points. In theory, we treat equally both a 3D feature point and an object. As the rotation is defined by its angle-axis, semantic BA is performed by using the LM method with automatic differentiation in the Ceres Solver~\cite{agarwal2012ceres}.

\begin{figure}[t]
	\centering
	\includegraphics[width=0.77\columnwidth]{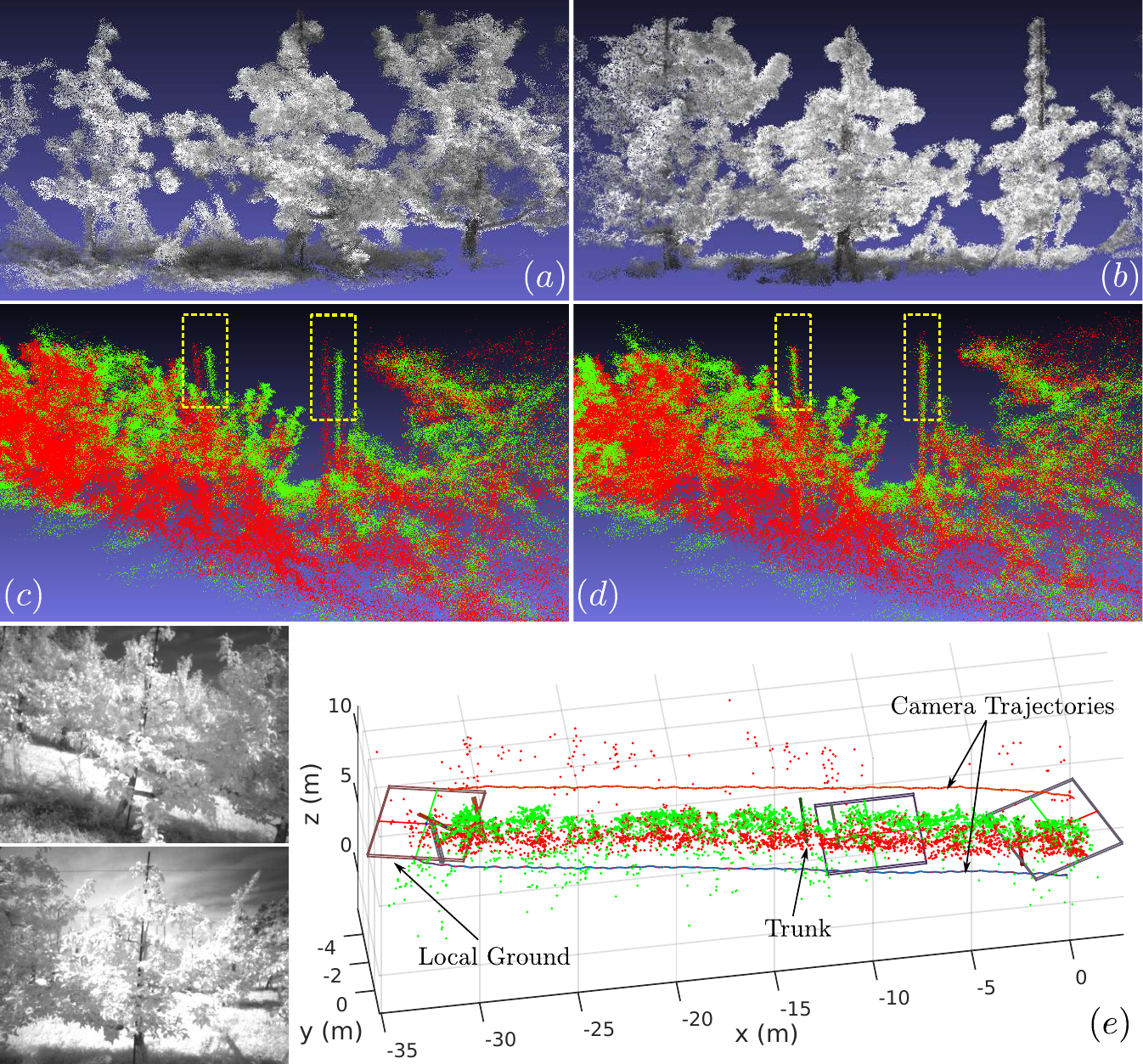}
	\caption{Merging 3D reconstruction of fruit trees for canopy volume estimation. (a) and (b): The 3D model viewed from both sides. (c): Some trunks are still misaligned after initial transformation. (d): Misalignments are eliminated by semantic BA. (e): 3D features from both sides are shown with camera poses  and semantic information (captured by stereo infrared cameras).}
	\label{fig:showSBA}
	%\vspace*{-4mm}
\end{figure}

\subsection{Measuring Tree Morphology and Yield Mapping} \label{sec:yieldmorph}
A coherent geometric representation of both sides of a tree row is a precursor to accurately estimating various semantic traits, such as trunk diameter, canopy volume, tree height and fruit count. Following the steps described in Sec.~\ref{subsec:globalign}$\sim$\ref{subsec:merging}, we obtain a merged reconstruction from both sides. In this section, we utilize this reconstruction to measure different desired semantic traits.
%We start with yield mapping.
%\subsection{Measuring Tree Morphology}
%In our framework, the trunk diameter estimation is first performed as an input for merging two-sides reconstruction. Canopy-volume and tree-height measurements are conducted based on the merged 3D model of fruit trees, which are illustrated using another dataset captured by stereo infrared cameras from a good view of tree canopies.

\subsubsection{Trunk Diameter} \label{subsubsec:diameter}
3D dense models of a tree from both sides $\mathcal{F}$ and $\mathcal{B}$ are obtained using volumetric fusion of depth maps from all nearby frames (see Fig.~\ref{fig:volumetricTrunkGround}). We first estimate the ground plane as discussed in Sec.~\ref{subsubsec:plane}.
The 3D points of the trunk for each detection frame $c$ are extracted based on 3D meshes of the most-detected cylindrical trunk segment from two sides (see Sec.~\ref{subsubsec:trunk}).
The trunk diameter is thus robustly estimated from both sides by minimizing the cost (with the robust Huber loss $\rho$ applied)
\begin{equation} \label{objectCylinderFB}
\begin{gathered}
\argmin_{^{\mathcal{F}}\mathbf{n}_d, ^{\mathcal{B}}\mathbf{n}_d, ^{\mathcal{F}}\mathbf{O}_d, ^{\mathcal{B}}\mathbf{O}_d, r_d} \sum_{p \in \{\mathcal{F},\mathcal{B}\}} \rho\left(e_d^2(\mathbf{X}_p, d)\right) + \lambda \sum_c \sum_{{p \in \mathcal{V}(c)}} \rho\left(E_r^2(p, c, d)\right) \\
E_r(p, c, d) =  \| ^c\mathbf{x}_p - p_r(\mathbf{X}_p, c, d) \|
\end{gathered} ,
\end{equation}
where $\mathbf{X}_p$ is a 3D trunk point either from $\mathcal{F}$ or $\mathcal{B}$, and $E_r$ is the 2D cost function that measures the distance between the depth pixel $^c\mathbf{x}_p$ of $\mathbf{X}_p$ visible from $c$-th frame and its corresponding surface point on the image.
The second term of the cost helps constrain the size, position and orientation of the cylinder from multiple image frames of two sides.
The trunk diameter is eventually $2r_d$, which serves as an input in Sec.~\ref{subsubsec:SBA}.

%\begin{figure}[t]
%	\centering
%	\includegraphics[width=0.9\columnwidth]{figs/method_fitting.pdf}
%	\caption{Front-side and back-side volumetric fusion using nearby frames. (a) and (b): Extracted 3D models of the trunk from both sides. (c) and (d): Extracted 3D models of the local ground from both sides.}
%	\label{fig:fittingOverview}
%	%\vspace*{-4mm}
%\end{figure}

%\begin{figure}[t]
%	\centering
%	\includegraphics[width=0.9\columnwidth]{figs/tree_morphology.pdf}
%	\caption{The scheme of estimating canopy volume and tree height. (a): Merged 3D model of a tree row (white front-side points and black back-side points) is partitioned by cutting planes. (b): Top-view tree segmentation based on the union of the cuboid and two-half cylinders. (c) and (d): Segmented tree viewed from both sides. (e): Generated alpha shape with a bounding box on the local ground.}
%	\label{fig:treeMorphology}
%	%\vspace*{-2mm}
%\end{figure}

\subsubsection{Canopy Volume} \label{subsubsec:volume}
\begin{figure}[t] % 0.490245085, 0.231627062, 0.278127853
	\centering
	\begin{subfigure}{0.451025478\textwidth}
		\centering
		\includegraphics[width=1\linewidth]{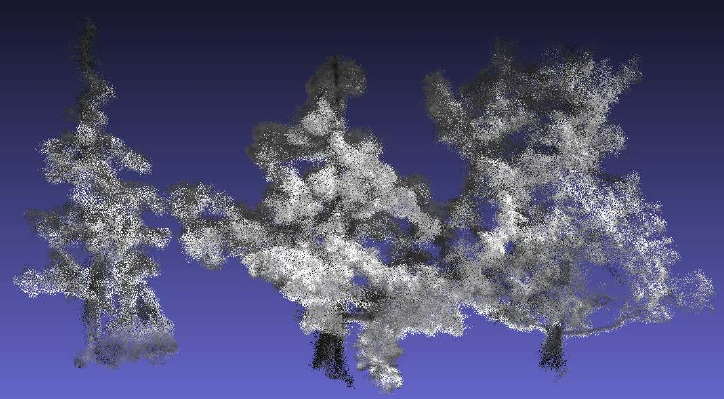}
		\caption{Merged 3D model of a tree row}
	\end{subfigure}
	~
	\begin{subfigure}{0.213096897\textwidth}
		\centering
		\includegraphics[width=1\linewidth]{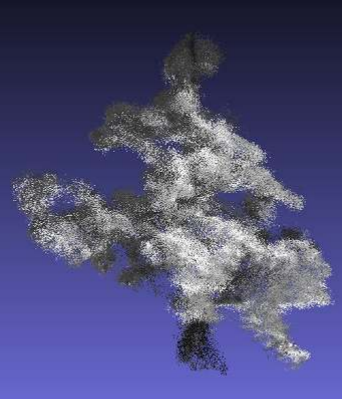}
		\caption{Segmented tree}
	\end{subfigure}
	~
	\begin{subfigure}{0.255877625\textwidth}
		\centering
		\includegraphics[width=1\linewidth]{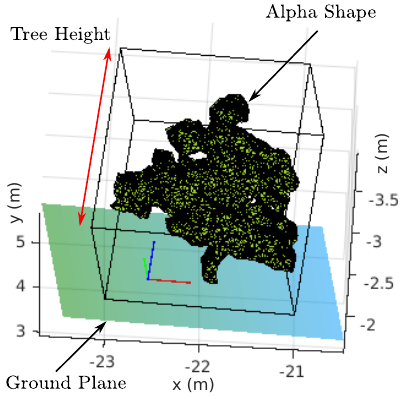}
		\caption{Tree volume and height}
	\end{subfigure}
	\caption{Scheme of estimating canopy volume and tree height. (a): Front-side points are colored white while back-side points are black. (b): The second tree is segmented by the proposed tree segmentation algorithm (see Sec.~\ref{subsubsec:volume}). (c): Generated alpha shape with a bounding box on the local ground.}
	\label{fig:treeMorphology}
\end{figure}

With a good view of canopies of fruit trees, two-sides 3D reconstructions are first merged in Fig.~\ref{fig:showSBA}. Local grounds are removed (see Fig.~\ref{fig:treeMorphology}a) given refined semantic information $[\mathbf{R}_s | \mathbf{t}_s ]$. Trunks information indicates the track of the tree row.

\begin{figure}[t]
	\centering
	\begin{subfigure}{0.48\textwidth}
		\centering
		\includegraphics[width=1\linewidth]{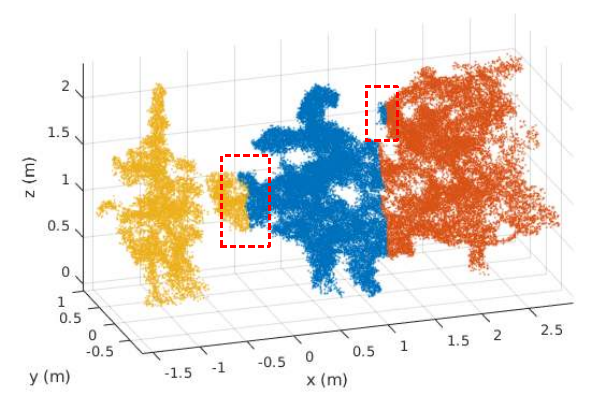}
		\caption{Tree segmentation based on points distribution}
	\end{subfigure}
	~
	\begin{subfigure}{0.48\textwidth}
		\centering
		\includegraphics[width=1\linewidth]{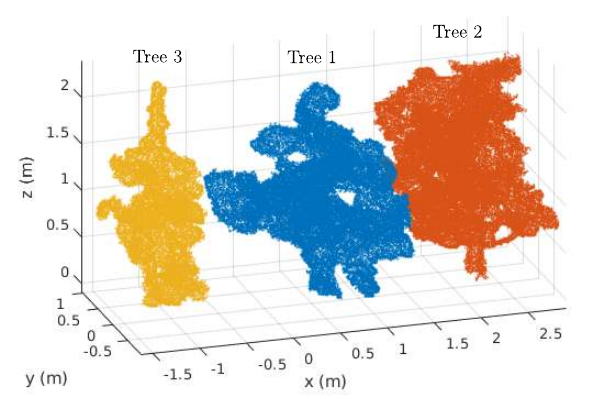}
		\caption{Proposed tree segmentation}
	\end{subfigure}
	\caption{Comparison results of tree segmentation. (a): 3D points enclosed by red dashed boxes are from a single tree but misclassified into different trees. (b): Surface points obtained from the alpha shape of the merged 3D tree model are well segmented using the shrink-and-expand algorithm.}
	\label{fig:treeSegmentation}
\end{figure}

\textbf{Shrink-and-expand Segmentation:} Tree segmentation is a critical component of canopy volume estimation, as it enables measurements to be associated with each individual tree. In modern orchards, there is often contact between adjacent trees. Points-distribution-based approaches (such as hidden semi-Markov model~\cite{bargoti2015pipeline} and k-means clustering~\cite{arthur2007k}) always lead to misclassified tree segmentation (see Fig.~\ref{fig:treeSegmentation}a). To address this misclassification, we propose a shrink-and-expand approach to well separate trees from each other (see Fig.~\ref{fig:treeSegmentation}b). The tree segmentation is performed as follows.

\begin{figure}[t]
	\centering
	\begin{subfigure}{0.32\textwidth}
		\centering
		\includegraphics[width=1\linewidth]{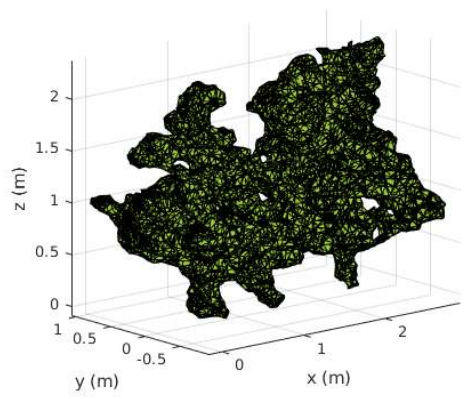}
		\caption{Input alpha shape}
	\end{subfigure}
	~
	\begin{subfigure}{0.32\textwidth}
		\centering
		\includegraphics[width=1\linewidth]{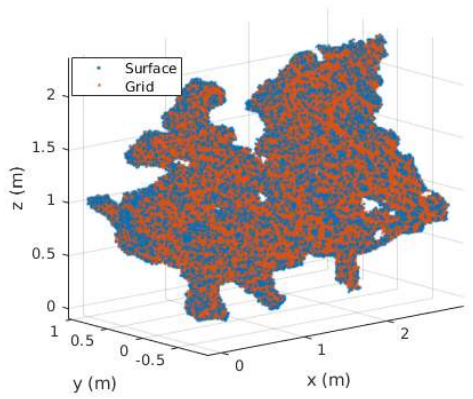}
		\caption{Surface and grid points}
	\end{subfigure}
	~
	\begin{subfigure}{0.32\textwidth}
		\centering
		\includegraphics[width=1\linewidth]{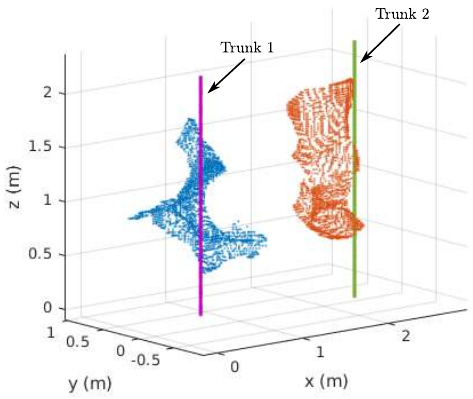}
		\caption{Shrink stage}
	\end{subfigure}
	\\
	\begin{subfigure}{0.32\textwidth}
		\centering
		\includegraphics[width=1\linewidth]{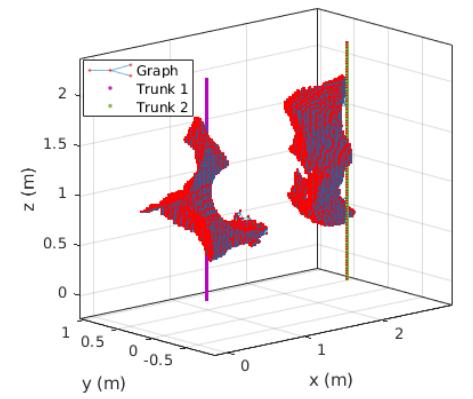}
		\caption{Generated graph}
	\end{subfigure}
	~
	\begin{subfigure}{0.32\textwidth}
		\centering
		\includegraphics[width=1\linewidth]{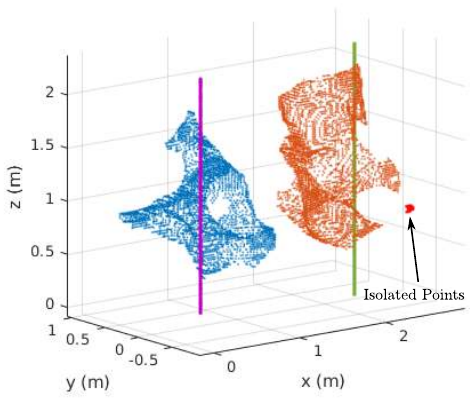}
		\caption{Expand stage}
	\end{subfigure}
	~
	\begin{subfigure}{0.32\textwidth}
		\centering
		\includegraphics[width=1\linewidth]{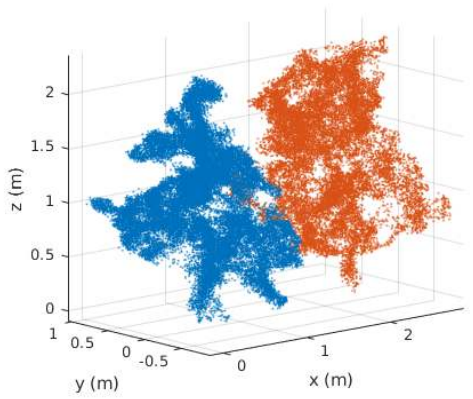}
		\caption{Output segmentation}
	\end{subfigure}
	\caption{Tree segmentation process of the shrink-and-expand algorithm.}
	\label{fig:segmentationSteps}
\end{figure}

\begin{enumerate} [label=(\alph*)]
	\item A pair of adjacent 3D tree models are extracted using the points-distribution-based method. We build an alpha shape enclosing all 3D points of two trees (see Fig.~\ref{fig:segmentationSteps}a). The alpha radius is the smallest producing the alpha shape that has only one region~\cite{edelsbrunner1994three}.
	\item We fill the alpha shape with grid points in the 3D space. For each point of the tree model, its closest neighbor point is retrieved and their distance is calculated. The grid resolution is determined by the median value of the closest distances for all 3D points. We keep all surface points of the alpha shape and all grid points (see Fig.~\ref{fig:segmentationSteps}b).
	\item At the shrink stage, for each iteration $i$, we remove a layer of points $\mathbf{L}_i$ whose distances from the surface points are less than the grid resolution, and rebuild an alpha shape to enclose the remaining points. Isolated points $\mathbf{v}_i$ are also removed if the volume of their alpha-shape region is less than a threshold (e.g., 10\%) of the principal region. We keep removing points layer by layer until there are only two principal regions left, which are close to their corresponding trunk lines perpendicular to local grounds (see Fig.~\ref{fig:segmentationSteps}c).
	\item We generate a graph where all grid points of principal regions are connected to their nearby grid points, and trunk points are connected to their closest grid points. A grid point, for example, is classified as tree 1 if it has the shortest path to trunk 1, otherwise as tree 2 (see Fig.~\ref{fig:segmentationSteps}d).
	\item At the expand stage, we retrieve points layer by layer following the decreased-iteration order. A graph is rebuild for retrieved points, where each point is connected to its closest tree point. For a point of layer $\mathbf{L}_i$, it is classified as tree 1 if it has the shortest path to tree 1. For a point of isolated region $\mathbf{v}_i$, it is classified as tree 2 if it has the shortest path to tree 2 (see Fig.~\ref{fig:segmentationSteps}e).
	\item The segmentation output is obtained after all removed points are retrieved.
	Now the alpha shape built in (a) is separated for both trees.
	We thus obtain the segmented 3D depth points of each tree that are enclosed by the corresponding separated alpha shape (see Fig.~\ref{fig:segmentationSteps}f). The graph-based shrink-and-expand algorithm conserves contacted tree structure, such as a branch stretching out towards other trees.
\end{enumerate}
%Based on 3D points distribution~\cite{bargoti2015pipeline}, initial tree segmentation is performed by cutting planes perpendicular to the row track. 
%The cuboid bounding box of a tree is created.
%From a top view, we assume that a tree is centered at its trunk location projected onto the local ground.
%To take care of the canopy overlap, the half side of a tree is enclosed by a cylinder with the radius $R_s = \sqrt{2}d_s$, where $d_s$ is the distance from the trunk to the cutting plane (see Fig.~\ref{fig:treeMorphology}). 
%Each tree is thus segmented by taking the union of the bounding box and two-half cylinders.

Two pairs of trees (e.g., trees 1 \& 2 and trees 1 \& 3) are segmented using the shrink-and-expand algorithm.
For each tree pair, we cut the common tree (i.e. tree 1) using a plane through its trunk line, and obtain the whole segmented tree by taking the union of two half-cut models (see Fig.~\ref{fig:treeSegmentation}b).
We build an alpha shape~\cite{edelsbrunner1983shape} enclosing all 3D points of each segmented tree (see Fig.~\ref{fig:treeMorphology}c). The choice of alpha radius depends on different horticultural applications. The canopy volume is automatically calculated by the alpha-shape algorithm~\cite{edelsbrunner1994three}.

\subsubsection{Tree Height}
Semantic bundle adjustment outputs optimized information of trunks and local grounds. Based on the trunk location, the pole in the middle of a tree is first segmented out for modern orchards. A bounding box for each tree is then created to enclose its alpha shape from the local ground plane to the top (see Fig.~\ref{fig:treeMorphology}c). The tree height is thus obtained as the height of the bounding box.

%\subsection{Measuring Tree Morphology and Yield Mapping} \label{sec:yieldmorph}
%A coherent geometric representation of the both sides of a tree row is a precursor to accurately estimating different various semantic traits such as trunk diameters, canopy volume, tree height and fruit counts. Following the steps described in Sec.~\ref{subsec:globalign}$\sim$\ref{subsec:merging}, we obtain a merged reconstruction from both sides. In this section, we will utilize this reconstruction to measure different desired semantic traits.
%%We start with yield mapping.
\subsubsection{Yield Mapping}\label{subsection:yieldmapmethod}
%Yield mapping in specialty crops is a well-studied problem in literature~\cite{wang,das2015devices,hung2015feature,gongal2016apple,roy2016counting,peng2016semantic,isler2016robotic}. Most of the existing systems including ours, strive to compute fruit counts from the single side of a row and establish a correlation with ground truth.
\begin{figure}[t]
	\centering
	\includegraphics[width=0.95\columnwidth]{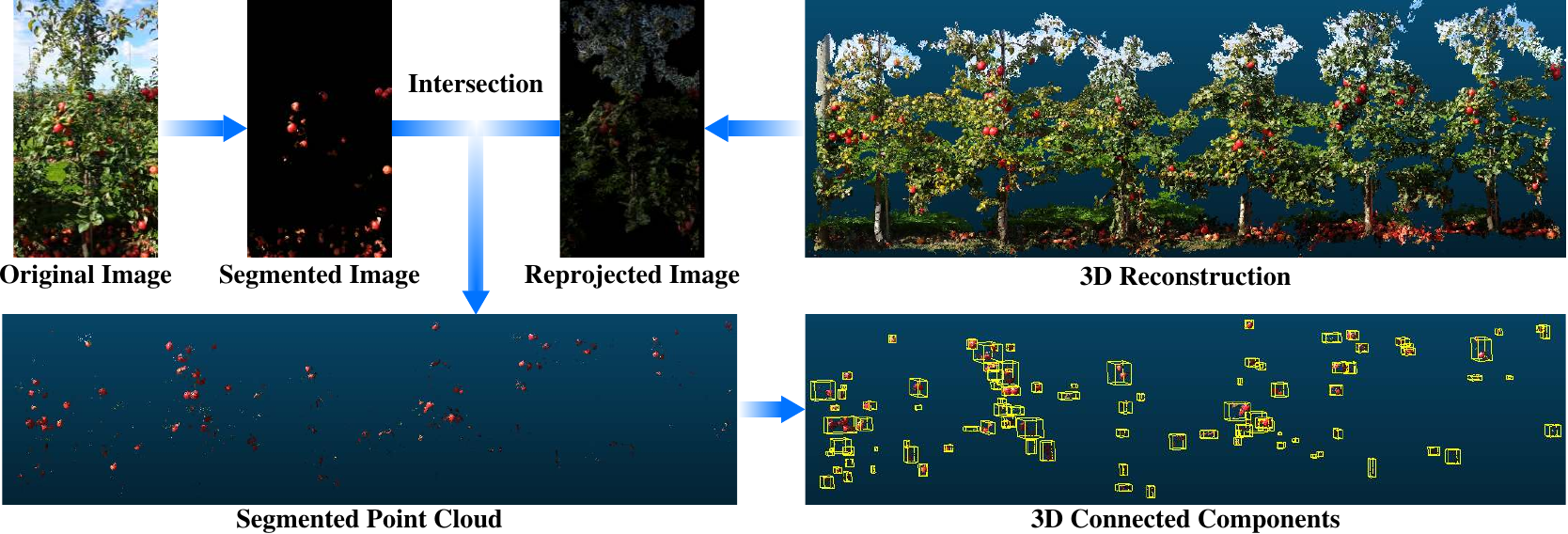}           
	\caption{Segmenting apples in 3D. We segment the original images using our segmentation method~\cite{roy2018arxiv}. Concurrently, we project the 3D points to the camera frames to create reprojected images. By intersecting these two images, we obtain the 3D points belonging to apples. Finally, we perform a connected component analysis on these 3D points.}
	\label{fig:segmentApples3D}
\end{figure}

\begin{figure}[t]
	\centering
	\includegraphics[width=0.9\columnwidth]{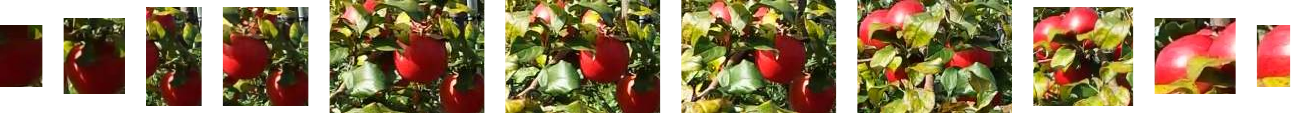}           
	\caption{Tracking an apple cluster over multiple frames.}
	\label{fig:trackApples}
\end{figure}

\begin{figure}[t]
	\centering
	\includegraphics[width=0.99\columnwidth]{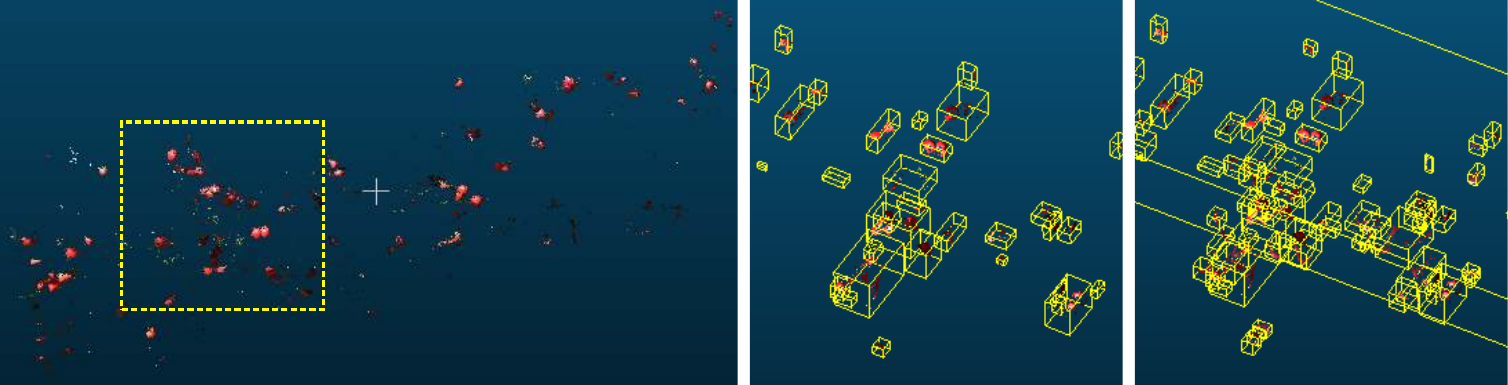}    
	\caption{Merging apple counts from both sides. The left figure shows the detected apples in 3D from the front side. Figure in the middle shows the computed connected components on the same side. Figure in the right shows how the connected components from the back are intersecting with the front ones. Apple counts from the intersecting clusters are merged using the inclusion-exclusion principle~\cite{andreescu2004inclusion}.}
	\label{fig:applecountingbothsides}
\end{figure}

As we show in Sec.~\ref{subsec:yieldMappingexp}, the number of visible fruits from a single side can vary significantly, resulting in erroneous estimates. For precise mapping, in addition to tracking the fruits from a single side, we need to register them from both sides of the tree row. Our merged reconstruction enables us to accomplish this task for both RGB and RGB-D data. First, we detect the fruits in the input images using our previously developed segmentation method~\cite{roy2018arxiv}. We utilize the detected apples in images to segment the apples in the 3D reconstruction. This segmentation step is different for RGB and RGB-D data:
\begin{itemize}
\item For RGB-D reconstructions, we have a one-to-one correspondence between the detected apple pixels and the 3D points. Therefore, the 3D segmentation is trivial.
\item For RGB data, we obtain semi-dense reconstructions based on the SfM output. Here, the point cloud does not have a one-to-one correspondence with the pixels. To establish the correspondence, we first project the reconstructed point cloud to the camera frames. Concurrently, we create a binary mask from the detected apples. Afterward, we compute the intersection between the reprojected image and the binary mask to identify the 3D points belonging to the detected apples. Subsequently, we perform a connected component analysis. Fig.~\ref{fig:segmentApples3D} shows an example of this process.
\end{itemize}

We project the individual clusters back to the images by utilizing the estimated camera motion. Intersecting this reprojected cluster image with the binary masks produced by the segmentation method, we obtain segmented images for each 3D clusters from multiple views. We count the fruits from these segmented images using our counting method developed in~\cite{roy2016counting}. 

A 3D cluster may appear in several frames (see Fig.~\ref{fig:trackApples}). We choose three frames with the highest amount of detected apple pixels and report the median count of these three frames as the fruit count for the cluster. It is notable that we remove the apples on the ground for all the single-side counts by using the computed ground planes. We perform these steps for both sides of the row.

To merge counts from both sides, we compute the intersection of the connected components from both sides (see Fig.~\ref{fig:applecountingbothsides}). Then, we compute the total counts by using the inclusion-exclusion principle~\cite{andreescu2004inclusion}. Essentially, we sum the counts from all the connected components, compute the intersection area among them (among 1, 2, ..., the total number of intersecting clusters) and add/subtract the weighted parts accordingly. This process is validated in Sec.~\ref{subsec:yieldMappingexp}.
%We continue with the measurement of different morphological traits in the following sections. Next, we present a robust method for trunk diameter estimation.

%----------------------------------------------------------------------
% SECTION IV: Experiments
%----------------------------------------------------------------------
\section{Experiments} \label{sec:experiments}
In this section, we conduct real experiments to evaluate our proposed system for merging 3D mapping of fruit trees from two sides and measuring their semantic traits.

\subsection{Datasets}%and Evaluation Metrics
The proposed system is tested using three RGB-D and three RGB datasets of apple-tree rows in different orchards separately captured from two sides (see Fig.~\ref{fig:experimentMerge} and Fig.~\ref{fig:dataset2_recons}).
Our merging algorithm is performed on each dataset to output complete 3D models of tree rows, in which we measure tree morphology for the RGB-D datasets and estimate fruit yield for the RGB datasets.
%Dataset-I is about an apple-tree row with a lot of wild weed captured in a horizontal view.
%Dataset-II is captured in a tilted view with a focus on tree trunks.
%Dataset-III is collected by a camera attached to a stick in a tilted-top view of tree canopies.
%\todo{In addition, we use three RGB datasets:}
The description of each dataset is as follows:
\begin{description}
	\item[RGB-D Datasets] \emph{Dataset-I} (968 images at 30 fps, 1920$\times$1080) contains an apple-tree row of 21 trees with a lot of wild weed captured in a horizontal view. \emph{Dataset-II} (2394 images at 60 fps, 640$\times$480) contains 27 trees captured in a tilted view with a focus on tree trunks. \emph{Dataset-III} (2020 images at 60 fps, 640$\times$480) of 30 trees is collected by a camera attached to a stick in a tilted-top view of the tree canopies.
	\item[RGB Datasets] \emph{Dataset-IV} (873 images at 30 fps, 1920$\times$1080) contains six trees that are mostly planar. Most of the apples on these trees (with $270$ apples in total) are fully red and visible from two sides. \emph{Dataset-V} (1065 images at 30 fps, 1920$\times$1080) contains ten trees that have non-planar geometry. Apples ($274$ in total) in these trees are mostly red. \emph{Dataset-VI} (831 images at 30 fps, 1920$\times$1080) contains six trees that have non-planar geometry. Fruits ($414$ apples in total) in these trees are a mixture of red and green apples.
\end{description}
%\subsection{Implementation Details}
%Dataset-I contains 21 trees.

%\begin{table}[t]
%	\centering
%	\begin{tabu} to 0.99\columnwidth {| X[2.4,m,c] | | X[m,c] | X[m,c] | X[m,c] | X[m,c] | X[m,c] | X[m,c] |} 
% 	\hline
% 	\multirow{2}{*}{Model} & \multicolumn{6}{c|}{Section ID of Mean Canopy Volume (m$^3$)}\\
% 	\cline{2-7}
% 	& V-1 & V-2 & V-3 & V-4 & V-5 & V-6\\
% 	\hline\hline
% 	Cylinder & 2.957 & 3.105 & 2.503 & 2.185 & 3.155 & 3.307\\
% 	\hline
% 	Alpha Shape & 1.585 & 1.873 & 1.351 & 1.227 & 1.777 & 1.912\\
% 	\hline
% 	Convex Hull & 1.805 & 2.177 & 1.460 & 1.322 & 2.064 & 2.202\\
% 	\hline
%	\end{tabu}
%	\caption{Mean canopy volume of 6 tree sections using different models.}
%	\label{table:volume}
% 	%\vspace*{-6mm}
%\end{table}

\subsection{Morphology Measurements and Yield Estimation Results}

\subsubsection{Trunk Detection} \label{subsubsec:trunkDetectionResult}
\textbf{Training:} For the weight initialization, we transfer the learning results from the pre-trained weights of the COCO dataset~\cite{lin2014microsoft} to exploit low-level features from lower layers. Since the COCO dataset has 81 classes and ours has only 2 classes (trunk and background), we first train the heads of the network (RPN, FCN and mask branch) for 30 epochs with the learning rate at $10^{-3}$. All the layers are then fine tuned for 30 epochs with the learning rate at $10^{-4}$.
The training is performed on a 12GB GPU(AWS GPU p2.xlarge) with the batch size set to 2. The model rescales input images as 1024px$\times$1024px with zero padding, and performs image augmentation (flipping images left/right 50\% of the time).
\begin{table}[!htbp]
	\caption{Evaluation results of trunk detection on five test orchards.} \label{table:detectionAcc}
	\begin{center}
		\begin{tabular}{|c||*{3}{>{\centering\arraybackslash}p{.07\linewidth}|}}
			\hline
			\multirow{2}{*}{Test Orchard ID} & \multicolumn{3}{c|}{Average Precision (AP)} \\
			\cline{2-4}
			& AP & AP$_{50}$ & AP$_{75}$\\
			\hline\hline
			Orchard-1 & 63.9 & 94.0& 72.8 \\
			\hline
			Orchard-2 & 51.2 & 98.3 & 57.7\\
			\hline
			Orchard-3 & 59.4 & 97.9 & 67.8\\
			\hline
			Orchard-4 & 50.8 & 93.3 & 55.4\\
			\hline
			Orchard-5 & 53.9 & 94.4 & 59.4\\
			\hline
		\end{tabular}
	\end{center}
\end{table}

\begin{figure}[t]
	\centering
	\includegraphics[width=0.95\columnwidth]{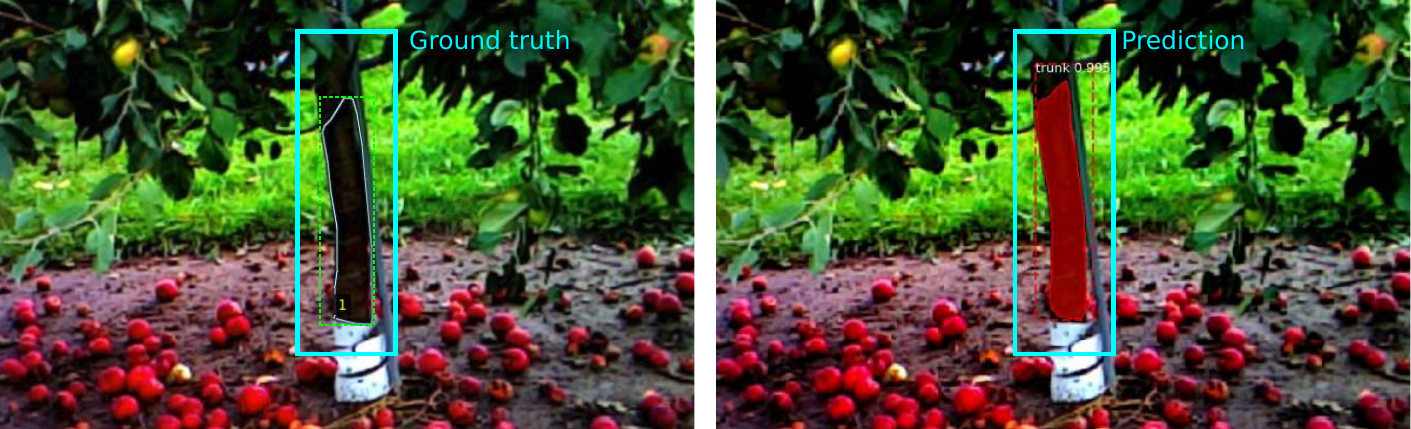}
	\caption{A trunk prediction instance (right) and its ground truth (left). The upper part of this trunk is occluded by a small branch, so the detector provides a weak prediction around its upper end but still outputs correct trunk boundaries along the trunk direction.}
	\label{fig:detectionBaseline}
\end{figure}

\begin{figure}[t]
	\centering
	\includegraphics[width=0.5\columnwidth]{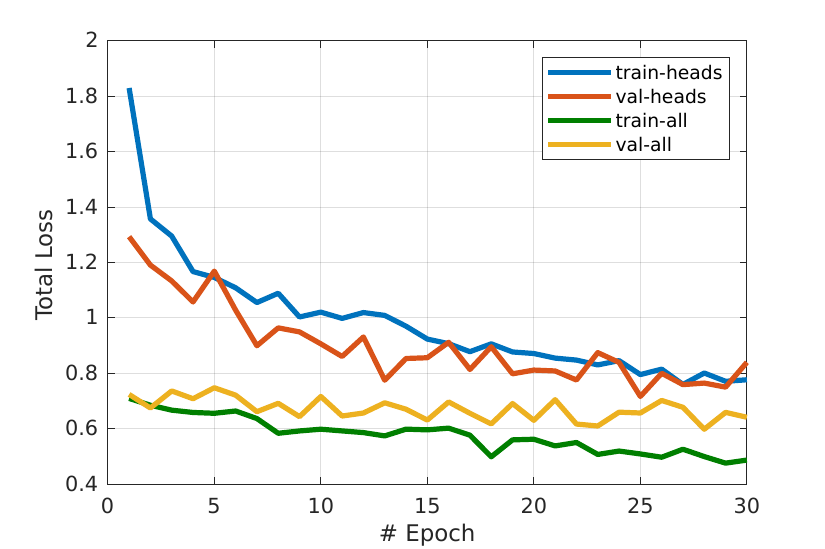}
	\caption{Total loss of train and validation data for each training procedure (training head layers and training all layers) versus the number of epochs.}
	\label{fig:trainValLoss}
\end{figure}

As stated in Sec.~\ref{subsubsec:trunkDetection}, the Mask R-CNN model is first trained on each set of training orchards and evaluated on its corresponding test orchard. In Table~\ref{table:detectionAcc}, we report the standard COCO metrics~\cite{lin2014microsoft} including AP (averaged over IoU thresholds), AP$_{50}$, and AP$_{75}$, where AP is evaluated using mask IoU (Intersection over Union). The high accuracy in AP$_{50}$ demonstrates the correctness of trunk prediction (see Fig.~\ref{fig:trunkDetection}).
%Fig.~\ref{fig:trunkDetection} presents qualitative segmentation results of five orchards.
Even though AP and AP$_{75}$ have relatively low accuracy, the model outputs clear boundaries along the trunk direction, which provides qualitative segmentation results for modeling trunk cylinder and merging steps (see Sec.~\ref{subsubsec:trunk}$\sim$\ref{sec:yieldmorph}).
For example, Fig.~{\ref{fig:detectionBaseline}} shows a correct prediction with the IoU of instance mask around $90\%$. The predicted trunk area is longer than the ground truth but still maintains correct boundaries. The difficult case is when the trunk is occluded by leaves or branches, so the detector has weak predictions around two trunk ends.
%The trunk detector is finally trained using all the training images from five datasets based on the above-mentioned strategy.
The model is finally trained using all the training images from five datasets based on the above-mentioned training strategy.
%We finally train the mode using all the training data of five orchards.
The train loss and evaluation loss for two training procedures (training heads first and then training all layers) are shown in Fig.~\ref{fig:trainValLoss}, which indicates that the model is not overfitted.

\subsubsection{Merging Two-Sides 3D Reconstructions}
To validate the proposed merging algorithm, we first visually check if the misalignment of landmarks (e.g., poles and tree trunks) is eliminated. The objective is to maintain a globally reasonable model of tree rows (from both sides) for measuring geometric traits from this 3D information. The accuracy of the merging and estimation algorithms are further tested by comparison with manual measurements of trunk diameter and tree height (see Sec.~\ref{subsubsec:measureTreeMorphology}).

\begin{figure*}[t]
	\centering
	\includegraphics[width=0.99\textwidth]{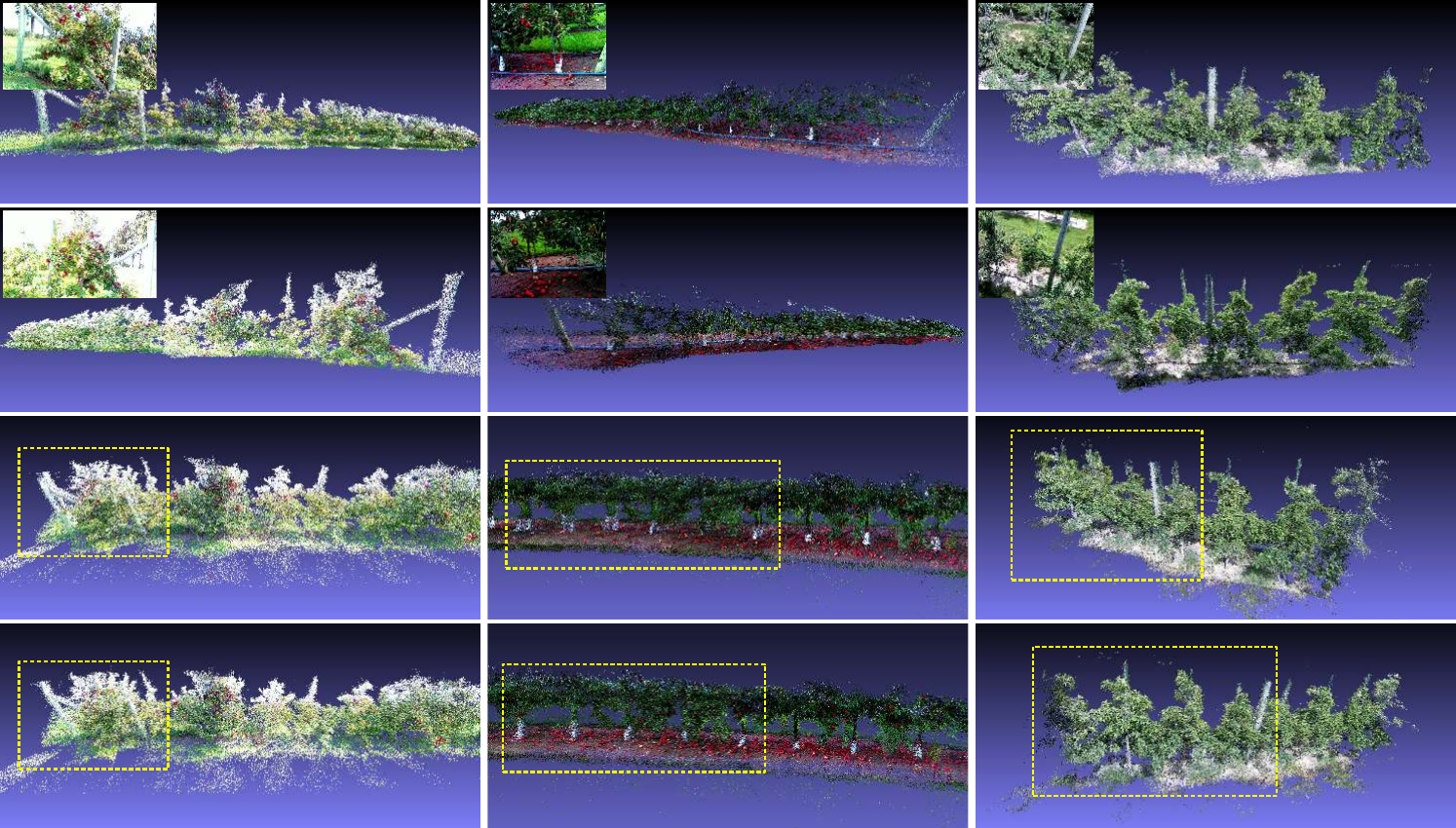}
	\caption{Merged 3D reconstructions from two sides of tree rows for Dataset-I, Dataset-II and Dataset-III. Rows 1 and 2: Front-side and back-side 3D reconstructions with scene images. Row 3: Misalignments (yellow boxes) of some landmarks after initial transformation. Row 4: Good 3D models are obtained by eliminating misalignments from semantic BA.}
	\label{fig:experimentMerge}
	%\vspace*{-2mm}
\end{figure*}

\begin{figure}[t]
	\centering
	\includegraphics[width =0.99\columnwidth]{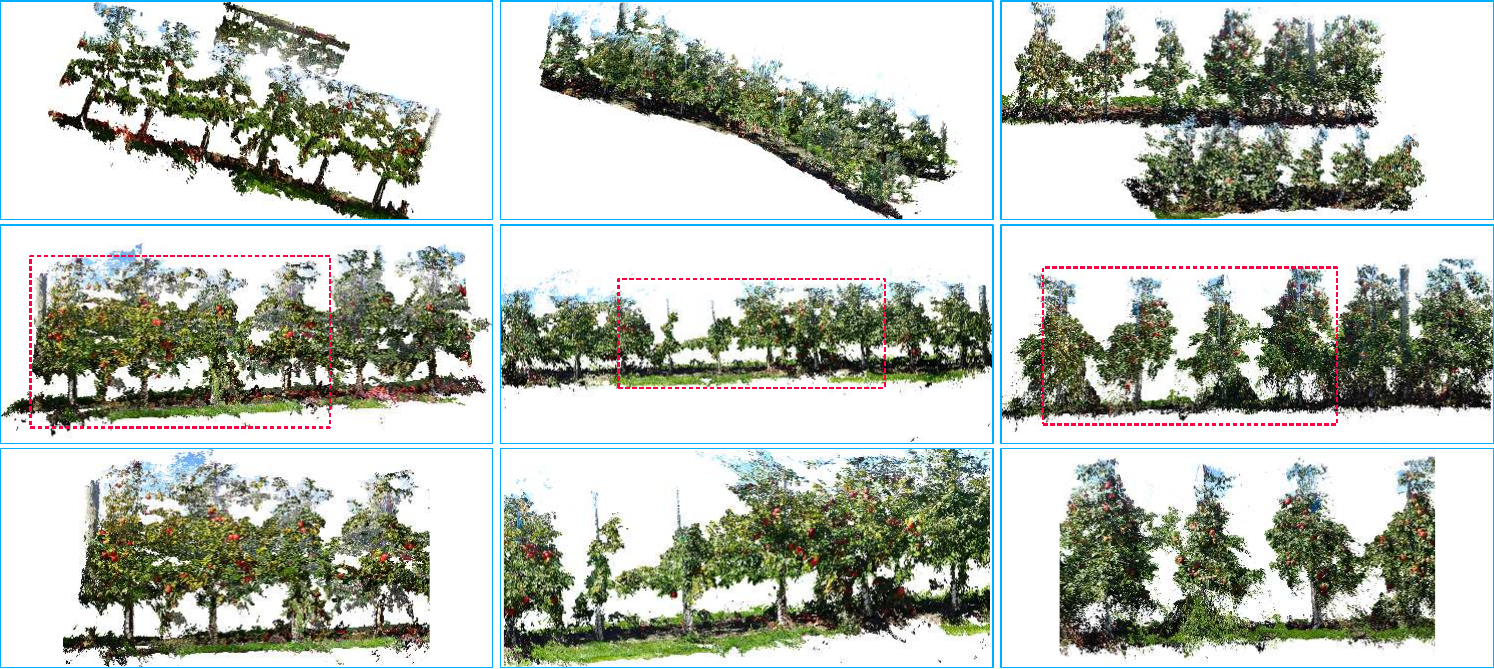}
	\caption{Merged 3D reconstructions from two sides of tree rows for Dataset-IV, Dataset-V and Dataset-VI. Rows 1: Input front-side and back-side 3D reconstructions. Rows 2: Output merged 3D models from both sides. Row 3: 3D models are well merged without misalignments from a zoom-in view.}
	\label{fig:dataset2_recons}
\end{figure}

As shown in Fig.~\ref{fig:experimentMerge} and Fig.~\ref{fig:dataset2_recons}, the proposed method is able to build well-aligned global 3D models of tree rows even without trunk detection for each tree. Specifically, duplicated poles and trunks are all merged. In practice, the merging algorithm only requires two-sides object correspondences around two ends and the middle of each tree row. When there is no need for estimating trunks diameter (e.g., counting fruits only), we can roughly fit a long section of a detected trunk as a cylinder, or even detect other landmarks, such as supporting poles and stakes. We make a general assumption that the local ground for each tree rather than the whole orchard ground is modeled as a plane, which makes our method applicable to any orchard environments without concern about the terrain.

\subsubsection{Measuring Tree Morphology} \label{subsubsec:measureTreeMorphology}
%\subsubsection{Comparison and Analysis}
%%%%followed by trunk diameter estimation in Dataset-II, and the estimation of canopy volume and tree height in Dataset-III.
Due to the interference of wild weed in Dataset-I, only three trunks and three local grounds are used as semantic information for the merging algorithm. For Dataset-II, 27 trunks are all detected with totally 3$\sim$4 frames per each from two sides in order to estimate trunks diameter. We use a caliper to measure the actual trunks diameter as the Ground Truth (GT).
In Dataset-II, we select 14 trees among 27 to demonstrate in detail the accuracy of our algorithm for trunks diameter estimation. Without 2D constraints, trunk diameters are always estimated larger than GT due to unreliable depth values around scene boundaries. Table~\ref{table:diameter} shows that with 2D constraints the average error of our diameter estimation is around 5 mm. Due to the integration of information from two sides, merged 3D tree models have more consistent diameter outputs (with lower mean error) compared with single-side estimations. For small trunks, the estimated results are still larger than GT, since the camera is relatively far from the trunks. Large pixel errors of trunk detection (low resolution for trunk boundaries) thus cause the diameter overfitting. This implies that the camera should closely capture trees with small trunks.
\begin{table*}[!htbp]
	\caption{Estimation errors of trunk diameter in Dataset-II.} \label{table:diameter}
	\begin{center}
		\begin{adjustbox}{width=\textwidth}
			\begin{tabular}{|c||c|c|c|c|c|c|c|c|c|c|c|c|c|c|c|}
				\hline
				Tree ID & T-2 & T-4 & T-6 & T-8 & T-9 & T-11 & T-13 & T-15 & T-18 & T-19 & T-22 & T-24 & T-26 & T-27 & Mean\\
				\hline\hline
				Ground Truth & 5.39 & 4.12 & 4.77 & 8.22 & 6.68 & 6.82 & 5.08 & 5.23 & 4.37 & 5.00 & 5.70 & 5.63 & 5.24 & 4.61 & $-$\\
				\hline
				Two-sides Est. & 5.24 & 5.10 & 5.48 & 8.04 & 6.56 & 6.50 & 5.51 & 5.87 & 5.29 & 5.70 & 5.99 & 5.49 & 5.77 & 5.37 & $-$\\
				\hline
				Error (cm) & 0.15 & 0.98 & 0.74 & 0.18 & 0.12 & 0.32 & 0.43 & 0.64 & 0.92 & 0.70 & 0.29 & 0.14 & 0.53 & 0.76 & {\bf 0.49}\\
				\hline\hline
				Front-side Est. & 5.59 & 5.25 & 5.44 & 8.49 & 6.51 & 7.27 & 4.67 & 6.05 & 5.24 & 5.86 & 6.07 & 5.40 & 5.75 & 5.26 & $-$\\ %
				\hline
				F. Error (cm) & 0.20 & 1.13 & 0.67 & 0.27 & 0.17 & 0.45 & 0.41 & 0.82 & 0.87 & 0.86 & 0.37 & 0.23 & 0.51 & 0.65 & 0.54\\
				\hline\hline
				Back-side Est. & 5.13 & 5.04 & 5.67 & 7.93 & 6.95 & 6.40 & 5.65 & 5.84 & 5.48 & 5.68 & 5.31 & 5.93 & 5.88 & 5.45 & $-$\\ %
				\hline
				B. Error (cm) & 0.26 & 0.92 & 0.90 & 0.29 & 0.27 & 0.42 & 0.57 & 0.61 & 1.11 & 0.68 & 0.39 & 0.30 & 0.64 & 0.84 & 0.59\\
				\hline
			\end{tabular}
		\end{adjustbox}
	\end{center}
\end{table*}

In Dataset-III, a subsample of six trees from 30 are chosen for merging assessment. Since the focus of this dataset is estimating canopy volume and measuring tree height, only tree trunks and their local grounds (the middle and two ends) are marked for merging.
%We use a caliper to measure the actual trunks diameter as the Ground Truth (GT).
The GT of trees height and their canopies diameter is obtained by using a measuring stick and a tape, respectively.
For Dataset-III, we perform tree height estimation of 14 trees chosen among 30. Table~\ref{table:height} shows that the average error of our tree height estimation is around 4 cm. We observe that for some trees (such as trees H-4 and H-23), there are inconsistent height estimations between two single sides. The reason is that the highest branch is not fully observed from one side but well captured from the other side. The estimation results for trunk diameter (Dataset-II) and tree height (Dataset-III) thus demonstrate the high accuracy of the proposed vision system.
\begin{table*}[!htbp]
	\caption{Estimation errors of tree height in Dataset-III.} \label{table:height}
	\begin{center}
		\begin{adjustbox}{width=\textwidth}
			\begin{tabular}{|c||c|c|c|c|c|c|c|c|c|c|c|c|c|c|c|}
				\hline
				Tree ID & H-1 & H-2 & H-3 & H-4 & H-5 & H-6 & H-7 & H-16 & H-18 & H-19 & H-20 & H-21 & H-22 & H-23 & Mean\\
				\hline\hline
				Ground Truth & 2.159 & 2.032 & 2.362 & 2.515 & 2.083 & 1.981 & 2.108 & 2.438 & 2.413 & 2.337 & 2.032 & 2.057 & 2.489 & 2.413 & $-$\\
				\hline
				Two-sides Est. & 2.145 & 2.050 & 2.453 & 2.463 & 2.131 & 1.997 & 2.087 & 2.357 & 2.456 & 2.311 & 1.990 & 2.084 & 2.496 & 2.361 & $-$\\
				\hline
				Error (m) & 0.014 & 0.018 & 0.091 & 0.052 & 0.048 & 0.016 & 0.021 & 0.081 & 0.043 & 0.026 & 0.042 & 0.027 & 0.007 & 0.052 & {\bf 0.038}\\
				\hline\hline
				Front-side Est. 
				& 2.149 & 2.061 & 2.467 & 2.368 & 2.139 & 1.894 & 2.096 & 2.301 & 2.464 & 2.318 & 1.974 & 2.092 & 2.510 & 2.352 & $-$\\
				\hline
				Error (m) 
				& 0.010 & 0.029 & 0.105 & 0.147 & 0.056 & 0.087 & 0.012 & 0.137 & 0.051 & 0.019 & 0.058 & 0.035 & 0.021 & 0.061 & 0.059\\
				\hline\hline
				Back-side Est. 
				& 2.092 & 1.961 & 2.445 & 2.474 & 2.128 & 1.990 & 2.031 & 2.343 & 2.447 & 2.295 & 1.999 & 2.072 & 2.475 & 2.306 & $-$\\
				\hline
				Error (m) 
				& 0.067 & 0.071 & 0.083 & 0.041 & 0.045 & 0.009 & 0.077 & 0.095 & 0.034 & 0.042 & 0.033 & 0.015 & 0.014 & 0.107 & 0.052\\
				\hline
			\end{tabular}
		\end{adjustbox}
	\end{center}
\end{table*}

\begin{figure*}[t]
	\centering
	\includegraphics[width=0.99\textwidth]{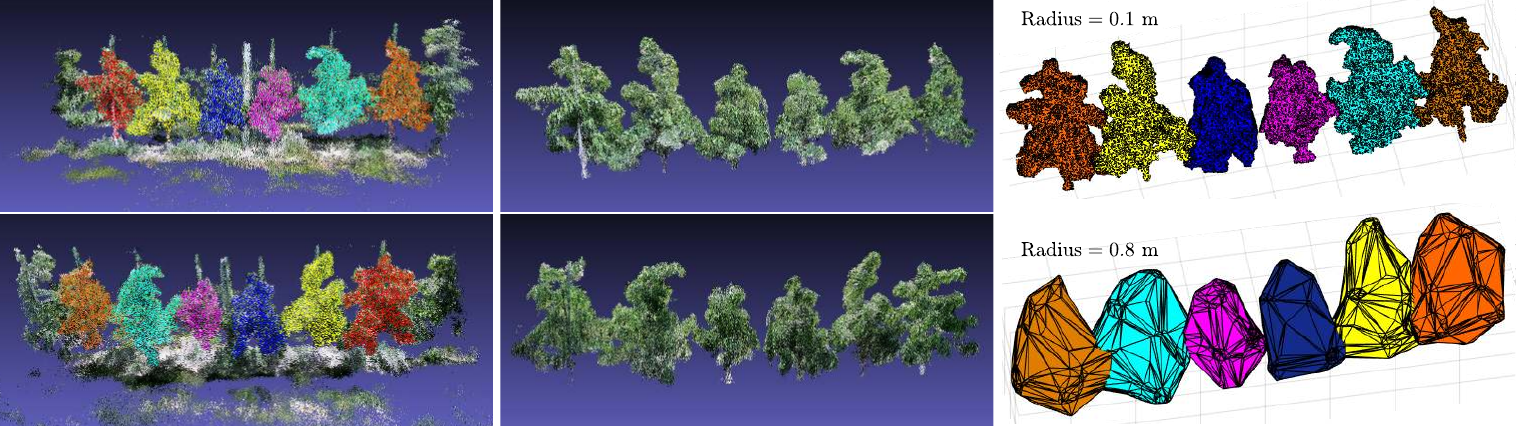}
	\caption{Six sample trees in Dataset-III are segmented and enclosed by alpha shapes. Column 1: Each tree is differentiated from front-side and back-side reconstructions. Column 2: Six trees are segmented out from both sides. Colume 3: Alpha shapes of six trees are generated using two different alpha radiuses from two-side views.}
	\label{fig:experimentShape}
	%\vspace*{2mm}
\end{figure*}

\begin{figure}[t]
	\centering
	\includegraphics[width=0.55\columnwidth]{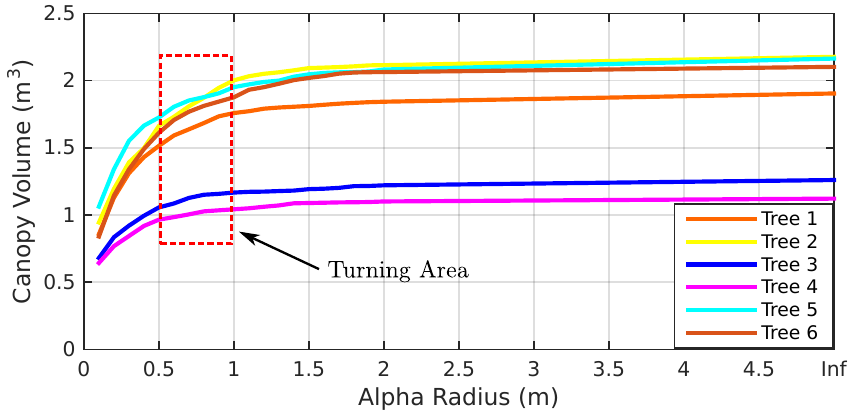}
	\caption{Canopy volumes estimated by alpha shape versus the alpha radius.}
	\label{fig:experimentRadius}
	%\vspace*{-2mm}
\end{figure}

To demonstrate canopy volume estimation, we first segment out six sample trees in Dataset-III and generate enclosing alpha shapes (see Fig.~\ref{fig:experimentShape}) to represent their canopies. However, the alpha radius should be appropriately chosen. The alpha shape with a small radius value will produce holes inside the canopy, which is not desirable from the view of horticultural study. Fig.~\ref{fig:experimentRadius} shows that the canopy volume increases and converges to a constant value as the alpha radius increases to infinity, which produces a convex hull. The best value of alpha radius should represent a canopy model without holes and produce the smallest volume. Here, we set the radius as $0.8$ m within the turning area (See Fig.~\ref{fig:experimentRadius} and Fig.~\ref{fig:experimentShape}). The alpha radius can be varied and selected based on different purposes of horticultural applications.

\begin{table}[!htbp]
	\caption{Mean canopy volume of six tree sections using different models.} \label{table:volume}
	\begin{center}
		\begin{tabular}{|c||c|c|c|c|c|c|}
			\hline
			\multirow{2}{*}{Model} & \multicolumn{6}{c|}{Section ID of Mean Canopy Volume (m$^3$)} \\
			\cline{2-7}
			& V-1 & V-2 & V-3 & V-4 & V-5 & V-6\\
			\hline\hline
			Alpha Shape & 1.585 & 1.873 & 1.351 & 1.227 & 1.777 & 1.912\\
			\hline
			Convex Hull (Merged Two Sides) & 1.805 & 2.177 & 1.460 & 1.322 & 2.064 & 2.202\\
			\hline
			Convex Hull (Summed Single Sides) & 3.126 & 3.328 & 2.419 & 2.265 & 3.465 & 3.353\\
			\hline
			Cylinder Assumption & 2.957 & 3.105 & 2.503 & 2.185 & 3.155 & 3.307\\
			\hline
		\end{tabular}
	\end{center}
\end{table}
One of the common methods used in horticultural science for modeling canopies is to treat a tree as a cylinder.
Simply summing up the tree volumes from two single-side reconstructions (i.e., the front side and the back side) is also performed.
To show the difference among different models of canopies, we divide 18 trees from Dataset-III into six sections based on their relatively similar sizes, and report the mean canopy volume of each section in Table~\ref{table:volume}. It should be notable that simple cylinder model overestimates the canopy volume.
The volumes added together from two single sides are even larger than the results using the cylinder model, since there is a substantial overlap between two-sides reconstructions in modern orchards (white front-side points and black back-side points are both visible from a single side, as shown in Fig.~\ref{fig:treeMorphology}).
Thus, it is reasonable to consider that the proposed method for canopy volume estimation is more suitable to generalize the geometry of tree structures, which is promising to build the ground truth of tree canopies for horticulturists using our vision system.

\subsection{Yield Mapping} \label{subsec:yieldMappingexp}
In this section, we evaluate our yield mapping technique described in Sec.~\ref{sec:yieldmorph}. Essentially, we quantify how the yield map from the merged reconstruction is more accurate than single-side estimates. Three RGB datasets (Dataset-IV, Dataset-V and Dataset-VI) are used for this purpose.

\begin{figure}[t]
	\centering
	\begin{subfigure}{0.48\textwidth}
		\centering
		\includegraphics[width=\linewidth]{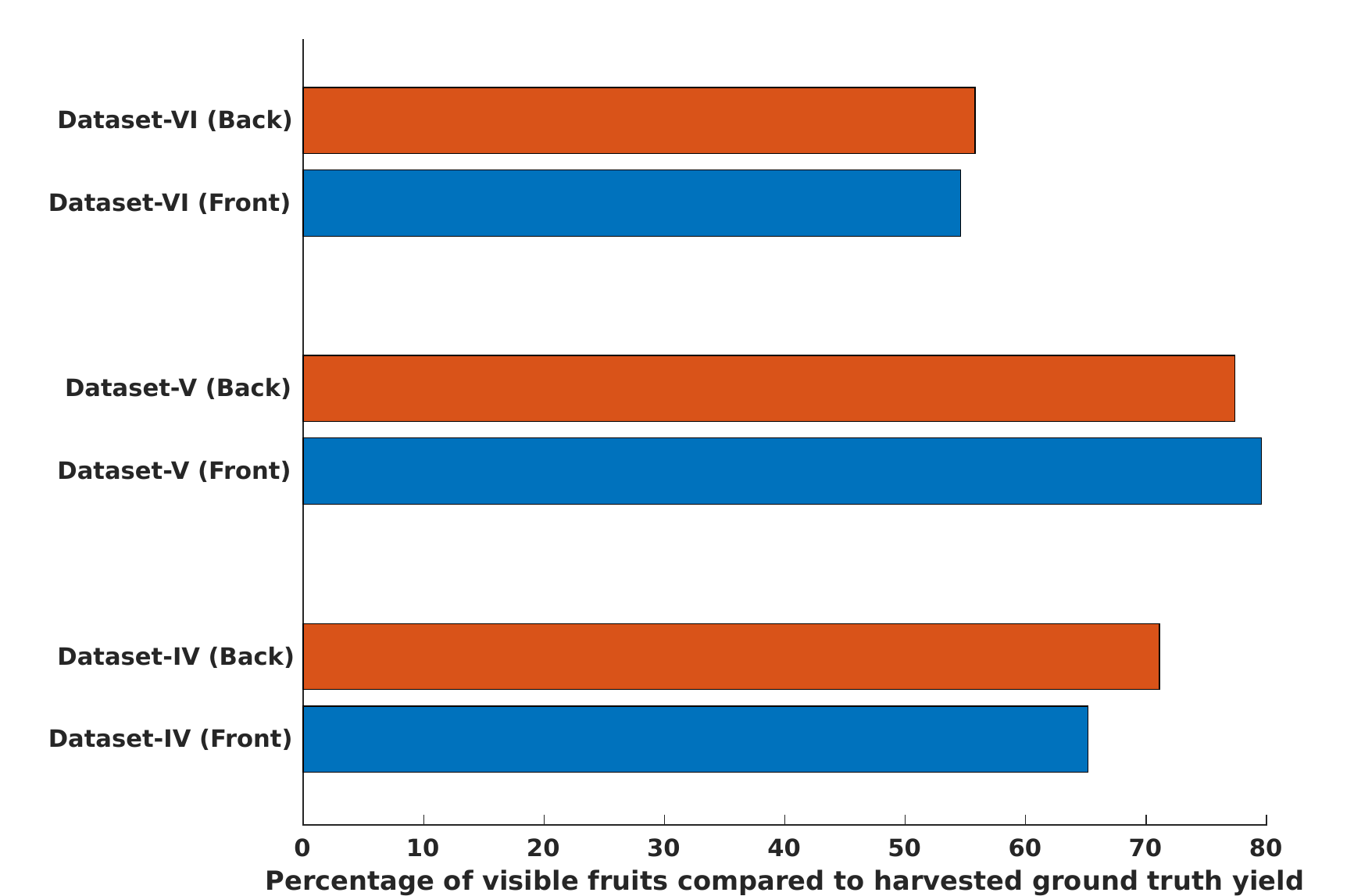}
		\caption{Percentage of visible apples from a single side compared to the yield of GT.}
		\label{fig:labelgt}   
	\end{subfigure}
	~
	\begin{subfigure}{0.48\textwidth}
		\includegraphics[width=\linewidth]{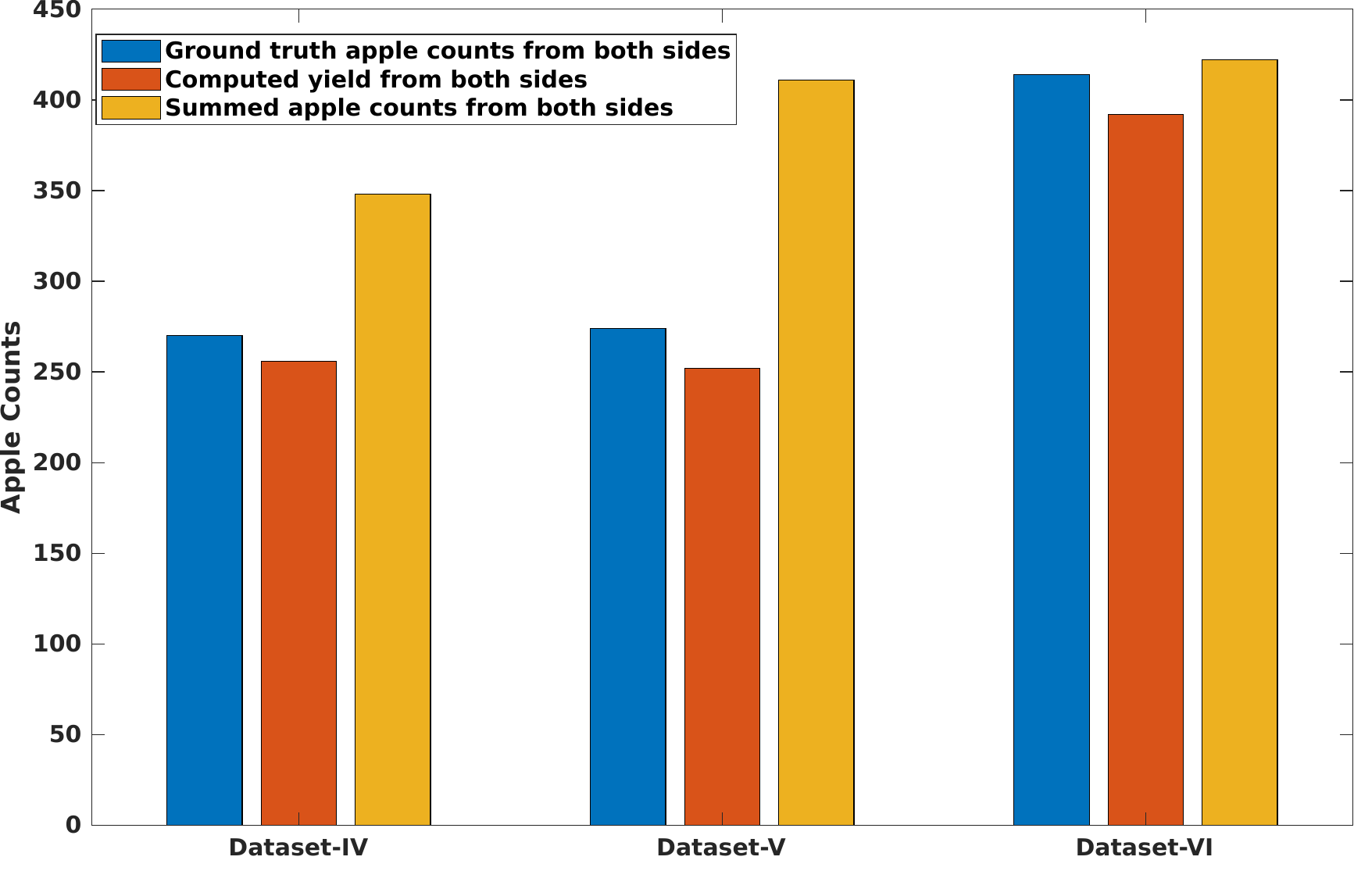}
		\caption{Summed apple counts from both sides compared to obtained GT.}
		\label{fig:yieldsum}   
	\end{subfigure}
	\caption{Our fruit counts and actual yield. (a): The percentage of visible apples from a single side compared to ground truth (GT) ($40.85\%\sim79.83\%$). (b): Fruit counts of both sides ($91.98\%\sim94.81\%$) and summed fruit counts from individual sides compared to the fruit counts of GT ($101.93\%\sim150\%$). Except for Dataset-VI, our computed counts from both sides are closer to GT.}
	\label{fig:bothsideyield}
\end{figure}

Before evaluating the performance of our yield mapping method, we find out how many fruits we can see from the single side of a row. Toward this goal, as described in Sec.~$5.2$ in~\cite{roy2018arxiv}, we annotate the apples in the images by hand and track them using estimated camera motion. We treat this hand-annotated fruit count as the total number of visible apples from a single side. From Fig.~\ref{fig:labelgt}, it is evident that the number of visible apples from a single side varies greatly across different datasets ($54.59\%\sim79.83\%$). This is expected in orchards where the trees are not well trimmed, and the size and shape of the trees vary significantly. Our data collection site (University of Minnesota Horticultural Research Center) resides in this category.
\begin{table}[!htbp]
	\caption{Summary of yield results in terms of fruit counts (FCs).} \label{tab:yield}
	\begin{center}
		\begin{tabular}{|c||c|c|c|}
			\hline
			Datasets & Harvested FCs & Merged FCs from both sides & Sum of FCs from single sides\\
			\hline\hline
			Dataset-IV & $270$ & $256$ ($94.81\%$) & $348$ ($128.89\%$)\\
			\hline
			Dataset-V\text{ } & $274$ & $252$ ($91.98\%$) & $411$ ($150\%$)\\
			\hline
			Dataset-VI & $414$ & $392$ ($94.68\%$) & $422$ ($101.93\%$)\\
			\hline
		\end{tabular}
	\end{center}
\end{table}

%\begin{table*}[t]
%\centering
%\begin{tabular}{|c|c|c|c|}
%\hline 
%\rule[-1ex]{0pt}{2.5ex} Datasets & Harvested fruit counts & Merged fruit counts from both sides & Sum of fruit counts from single sides \\ 
%\hline 
%\rule[-1ex]{0pt}{2.5ex} Dataset-IV & $270$ & $256$ ($94.81\%$) & $348$ ($128.89\%$) \\ 
%\hline 
%\rule[-1ex]{0pt}{2.5ex} Dataset-V & $274$ & $252$ ($91.98\%$) & $411$ ($150\%$)\\ 
%\hline 
%\rule[-1ex]{0pt}{2.5ex} Dataset-VI & $414$ & $392$ ($94.68\%$) & $422$ ($101.93\%$)\\ 
%\hline 
%\end{tabular} 
%\caption{Summary of yield results.}
%\label{tab:yield}
%\end{table*}

Now, as we find that it is hard to correlate the number of visible fruits from a single side to the total yield, we investigate the next simple solution. We simply add the apple counts from both sides and find how close we are to the actual yield. As shown in Fig.~\ref{fig:yieldsum} and Table~\ref{tab:yield}, the summed yields vary considerably across datasets ($101.93\%\sim150\%$). Finally, we find the yield using our method in Sec.~\ref{subsection:yieldmapmethod}, which ($91.98\%\sim94.81\%$) is much more consistent compared to the summed yield (see Fig.~\ref{fig:yieldsum} and Table~\ref{tab:yield}). 
%\subsection{Discussion}

\subsection{System Evaluation} \label{subsec:systemEvaluation}
In this section, we first conduct a system evaluation to qualitatively illustrate the performance of each component on a single dataset (Dataset-I), and in the next section we further discuss the limitations of our current vision system and practical considerations for applicability in robotics systems.
\begin{table*}[!htbp]
	\caption{Processing time of the proposed vision system on Dataset-I.} \label{table:processingTime}
	\begin{center}
		\begin{adjustbox}{width=\textwidth}
			\begin{tabular}{|c||c|c|c|c|}
				\hline
				\makecell{System \\ Component} & \makecell{Single-Side \\ Reconstruction} & \makecell{Trunk Detection \\ (per frame)} & \makecell{Trunk Modeling \\ (per trunk)} & \makecell{Ground Fitting \\ (per tree)}\\
				\hline
				\makecell{Processing Time} & $\sim15$ min & $\sim300$ ms & $\sim3$ s & $\sim3$ s\\
				\hline\hline
				\makecell{System \\ Component} & \makecell{Initial Alignment \\ (before semantic BA)} & \makecell{Semantic BA \\ (in total)} & \makecell{Tree Segmentation \\ (per tree)} & \makecell{Yield Estimation \\ (in total)}\\
				\hline
				\makecell{Processing Time} & $\sim 3$ min & $\sim8$ min & $\sim2$ min & $\sim 6$ min\\
				\hline
			\end{tabular}
		\end{adjustbox}
	\end{center}
\end{table*}
\begin{figure*}[t]
	\centering
	\includegraphics[width=0.95\textwidth]{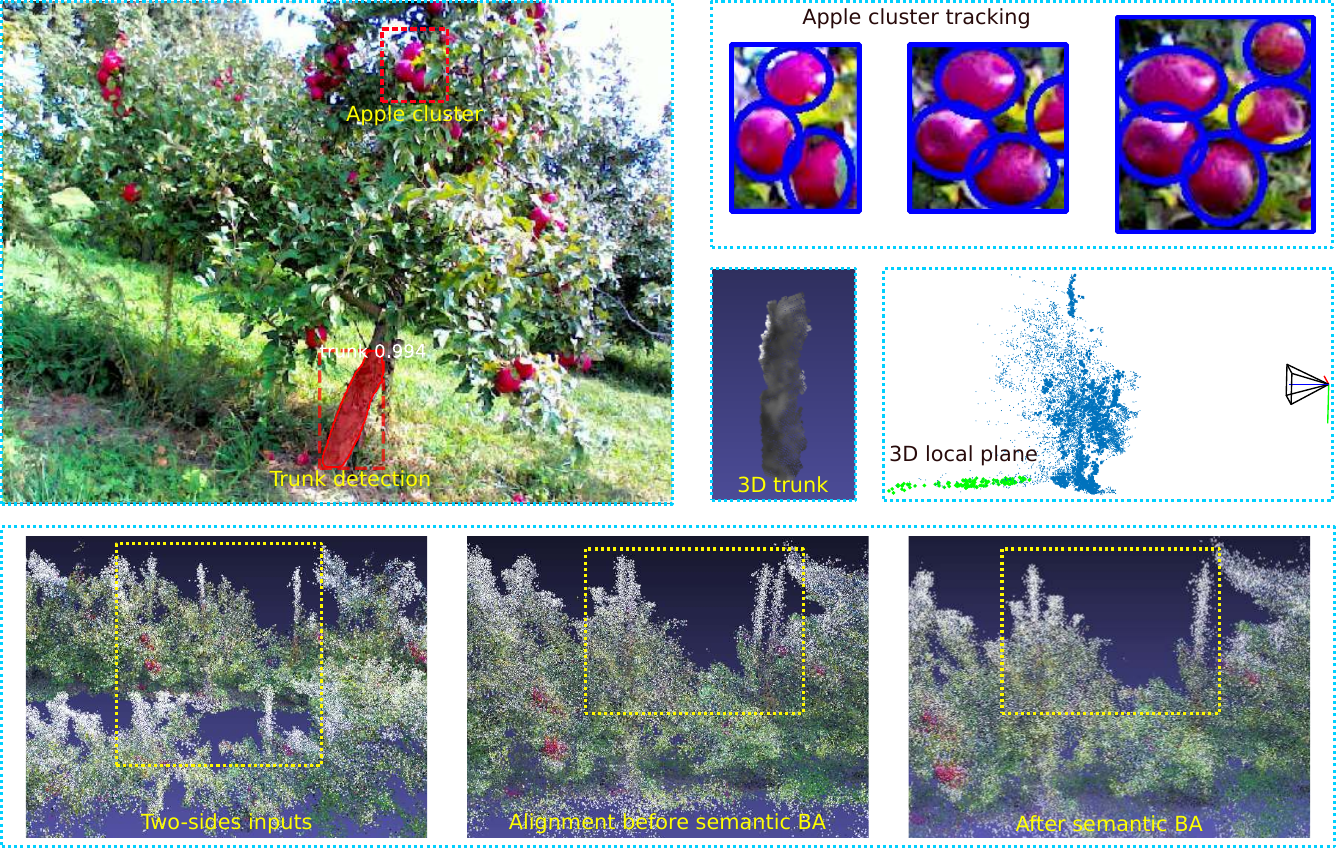}
	\caption{Trunk detection and yield estimation on a subset of Dataset-I. To complete the whole merging process, 3D trunk and local plane modeling, input reconstructions from two sides, and alignments before and after semantic BA of this subset are also presented.}
	\label{fig:systemEvaluation}
	%\vspace*{-2mm}
\end{figure*}

For Dataset-I, we already evaluate and show the 3D reconstructions from two sides (1st column in Fig.~{\ref{fig:experimentMerge})}, 3D trunk and local ground modeling (Fig.~{\ref{fig:showSBA}}e), alignments before and after semantic bundle adjustment (Fig.~{\ref{fig:experimentMerge}}), tree height estimation (Fig.~{\ref{fig:treeMorphology}}), and tree canopy segmentation (Fig.~{\ref{fig:treeSegmentation}}), respectively. The remaining components of our system are presented in Fig.~{\ref{fig:systemEvaluation}} on a subset of Dataset-I: trunk detection, and yield estimation. Dataset-I contains around 1000 RGB-D images collected by a camera moving at a fast speed of about {\SI{1}{\meter/\second}}. As stated in Sec.~{\ref{subsubsec:measureTreeMorphology}}, we detect three trunks and estimate their corresponding local grounds from two sides. The processing time of the whole vision system is shown in Table~{\ref{table:processingTime}}. Since after merging two-sides tree reconstructions and performing tree segmentation, trunk diameter and tree height estimations can be directly inferred from optimized trunk cylinder and local ground plane models, respectively. We thus omit the processing time for these two components. Although the most time-consuming part is the single-side reconstruction using the SfM algorithm, the processing time can be largely reduced by exploiting the 3D inputs from the real-time SLAM system.

\subsection{Discussion} \label{subsec:discussion}
During the field experiment, the camera moves back and forth facing toward the tree rows to capture trunks and canopies. For estimating tree height and volume, the camera is lifted up and slightly tilted down to observe the entire tree from a good viewpoint. Facing up toward the sky should be avoided since the camera would suffer from image saturation due to the sunlight.
For good reconstruction quality, the camera should be moved without significant orientation changes between consecutive frames to guarantee sufficient image overlap.
Depth images are recorded and utilized only for measuring trunk diameter, tree height and canopy volume, since the absolute scale of the real word should be inferred from 3D reconstructions with the depth information.

To make a general training data for the trunk detector, we incorporate multiple datasets collected from different orchards~\cite{hani2018comparative}. Several aspects are taken into account during the data collection, such as different camera viewing angles, the existence of wrappers on trunks, and the existence of leaves on branches. In the extreme case, we tested the trunk detection on a specific orchard when the detector is trained only on other orchards. The performance is satisfactory as long as there is no large difference in tree appearance of the test data from the training data. For example, the failure case is when the test dataset only contains dormant trees (the yellow box in Fig.~{\ref{fig:trunkDetection}}) collected during the winter. Without any leaves on trees, the trunk detector (trained only on the data with leaves) cannot output trunk segmentations with high confidence scores. However, as stated in Sec.~{\ref{subsubsec:trunkDetection}}, adding a small portion of the new dataset solves this problem.

The key idea behind the alignment of orthographic projection boundaries is that 3D reconstructions from two sides vary considerably in terms of geometry (i.e., they are different in scale, and might have different missing parts in the models). This is precisely the reason for applying the CPD method to align them at first.
This iterative method enables us to treat the small changes in geometry as noise and outliers, and to achieve approximately correct alignments.
However, the failure case is that the input reconstructions from two sides are not accurate enough or have no mutual information. For example, we aim to merge 3D reconstructions of a row of ten trees. The 3D model from the front side only reconstructs four trees, while the reconstruction from the back side builds the six trees that are not covered (or less than two trees are covered) from the front side. In this case, the alignment of orthographic projection boundaries will fail.

The proposed tree segmentation approach (shrink-and-expand segmentation) is designed based on the assumption of spatial settings common in modern orchards, where fruit trees are densely planted but with separate space in between. In such a scenario, our tree segmentation method compared to existing methods has better performance for handling contact between adjacent trees. Without the assumption above, the degenerate case is that two tree trunks are grown extremely close to each other (which rarely happens in modern orchards). In this case, our segmentation method has a similar result as the points-distribution-based approaches: the 3D model of such two trees is separated right in the middle as cut by a plane perpendicular to the ground. Without detailed information about the branch structures inside the overlapped area, these two trees are hardly separated.

\section{Conclusion and Future Work} \label{sec:conclusion}
In the modern high-density orchard setting, fruit trees have complex structures. Each individual tree can overlap substantially with its neighboring trees. Furthermore, due to the wind, image features detected by the camera are not stable and cannot be processed and tracked through long consecutive frames by directly applying classic mapping algorithms.
%Hence, we proposed robust detection and fitting methods with reasonable heuristics based on the physical traits of the trees from two sides to create accurate and consistent 3D models of each individual tree. These models allow us to further successfully measure the semantic parameters of interest and can be exploited to estimate additional parameters such as 6D poses and sizes of each individual fruit, if necessary.

In this work, we presented a vision system that merges reconstructions from two sides of tree rows, and uses this 3D information to measure semantic traits for phenotyping (i.e., fruit count, canopy volume, trunk diameter, and tree height). 3D models of fruit trees from two sides are generated separately and merged into a coherent model by exploiting global features (i.e., occlusion boundaries from orthographic views) and semantic information (i.e., detected trunks and local grounds).
It was shown that the process of merging the 3D models highly improves the accuracy of yield estimation of two sides.
Canopy volume can be readily computed based on the segmentation of each tree. We also estimate trunk diameter and tree height using our robust detection and fitting algorithms.
Our system is evaluated using multiple different types of tree datasets collected in orchards.
This is the first vision system that can measure the semantic parameters of trees in fruit orchards by using only an imaging device.
Although the moving platform of our proposed vision system is based on a handheld stick, the whole system can be mounted on Unmanned Aerial Vehicles (UAVs)~\cite{stefas2016vision} and Unmanned Ground Vehicles (UGVs)~\cite{peng2016semantic} which have been demonstrated to navigate through modern orchard environments.
%We verified the method with multiple simulated and real datasets.

In our ongoing work, we are testing our merging technique on other species of fruit trees (such as orange and peach trees) which have different tree characteristics. Our method also generalizes to these types of trees, since the essential components (i.e., trunk detection, local ground estimation, initial global alignment, and tree segmentation) generalize to these settings. Further, we would like to extend our existing counting method~\cite{roy2016counting} so that it utilizes 3D information for fruit counting. Additionally, we aim to localize individual apples in the 3D space and build a complete map, which a picker robot can use for path planning.

\subsubsection*{Acknowledgments}
We thank Professors James Luby, Cindy Tong, and Emily Hoover from the Department of Horticultural Science, University of Minnesota, for their expertise and help with the experiments.
We also thank our colleagues Joshua Anderson, Cheng Peng, and Nicolai H\"{a}ni from the University of Minnesota, for providing valuable feedback throughout this research.
Finally, we thank the anonymous reviewers for their careful reading of our manuscript and their insightful comments and suggestions.
This work is supported in part by USDA NIFA MIN-98-G02, and a subgrant from NSF \#1722310.
%The authors thank Professors Emily Hoover, Cindy Tong, and James Luby from the Department of Horticultural Science, University of Minnesota, for their expertise and help with the experiments.
%This work is supported in part by USDA NIFA MIN-98-G02, and a subgrant from NSF \#1722310.

\bibliographystyle{apalike}
\bibliography{jfrReferences}

\begin{thebibliography}{}

\bibitem[Agarwal et~al., 2012]{agarwal2012ceres}
Agarwal, S., Mierle, K., et~al. (2012).
\newblock Ceres solver.

\bibitem[Agisoft and St~Petersburg, 2017]{agisoft2017agisoft}
Agisoft, L. and St~Petersburg, R. (2017).
\newblock Agisoft photoscan.
\newblock {\em Professional Edition}.

\bibitem[Andreescu and Feng, 2004]{andreescu2004inclusion}
Andreescu, T. and Feng, Z. (2004).
\newblock Inclusion-exclusion principle.
\newblock In {\em A Path to Combinatorics for Undergraduates}, pages 117--141.
  Springer.

\bibitem[Arthur and Vassilvitskii, 2007]{arthur2007k}
Arthur, D. and Vassilvitskii, S. (2007).
\newblock k-means++: The advantages of careful seeding.
\newblock In {\em Proceedings of the eighteenth annual ACM-SIAM symposium on
  Discrete algorithms}, pages 1027--1035. Society for Industrial and Applied
  Mathematics.

\bibitem[Bac et~al., 2014]{bac2014stem}
Bac, C., Hemming, J., and Van~Henten, E. (2014).
\newblock Stem localization of sweet-pepper plants using the support wire as a
  visual cue.
\newblock {\em Computers and electronics in agriculture}, 105:111--120.

\bibitem[Bargoti and Underwood, 2017]{bargoti2017image}
Bargoti, S. and Underwood, J.~P. (2017).
\newblock Image segmentation for fruit detection and yield estimation in apple
  orchards.
\newblock {\em Journal of Field Robotics}, 34(6):1039--1060.

\bibitem[Bargoti et~al., 2015]{bargoti2015pipeline}
Bargoti, S., Underwood, J.~P., Nieto, J.~I., and Sukkarieh, S. (2015).
\newblock A pipeline for trunk detection in trellis structured apple orchards.
\newblock {\em Journal of Field Robotics}, 32(8):1075--1094.

\bibitem[Beder and F{\"o}rstner, 2006]{beder2006direct}
Beder, C. and F{\"o}rstner, W. (2006).
\newblock Direct solutions for computing cylinders from minimal sets of 3d
  points.
\newblock {\em Computer Vision--ECCV 2006}, pages 135--146.

\bibitem[Bonanni et~al., 2017]{bonanni20173}
Bonanni, T.~M., Della~Corte, B., and Grisetti, G. (2017).
\newblock 3-d map merging on pose graphs.
\newblock {\em IEEE Robotics and Automation Letters}, 2(2):1031--1038.

\bibitem[Bowman et~al., 2017]{bowman2017probabilistic}
Bowman, S.~L., Atanasov, N., Daniilidis, K., and Pappas, G.~J. (2017).
\newblock Probabilistic data association for semantic slam.
\newblock In {\em Robotics and Automation (ICRA), 2017 IEEE International
  Conference on}, pages 1722--1729. IEEE.

\bibitem[Cohen et~al., 2016]{cohen2016indoor}
Cohen, A., Sch{\"o}nberger, J.~L., Speciale, P., Sattler, T., Frahm, J.-M., and
  Pollefeys, M. (2016).
\newblock Indoor-outdoor 3d reconstruction alignment.
\newblock In {\em European Conference on Computer Vision}, pages 285--300.
  Springer.

\bibitem[Curless and Levoy, 1996]{curless1996volumetric}
Curless, B. and Levoy, M. (1996).
\newblock A volumetric method for building complex models from range images.
\newblock In {\em Proceedings of the 23rd annual conference on Computer
  graphics and interactive techniques}, pages 303--312. ACM.

\bibitem[Das et~al., 2015]{das2015devices}
Das, J., Cross, G., Qu, C., Makineni, A., Tokekar, P., Mulgaonkar, Y., and
  Kumar, V. (2015).
\newblock Devices, systems, and methods for automated monitoring enabling
  precision agriculture.
\newblock In {\em Proceedings of IEEE Conference on Automation Science and
  Engineering}.

\bibitem[del Moral-Mart{\'\i}nez et~al., 2015]{del2015georeferenced}
del Moral-Mart{\'\i}nez, I., Arn{\'o}, J., Sanz, R., Masip-Vilalta, J.,
  Rosell-Polo, J.~R., et~al. (2015).
\newblock Georeferenced scanning system to estimate the leaf wall area in tree
  crops.
\newblock {\em Sensors}, 15(4):8382--8405.

\bibitem[Dong and Isler, 2017]{dong2017linear}
Dong, W. and Isler, V. (2017).
\newblock Linear velocity from commotion motion.
\newblock In {\em Intelligent Robots and Systems (IROS), 2017 IEEE/RSJ
  International Conference on}, pages 3467--3472. IEEE.

\bibitem[Dong and Isler, 2018a]{dong18novel}
Dong, W. and Isler, V. (2018a).
\newblock A novel method for the extrinsic calibration of a 2d laser
  rangefinder and a camera.
\newblock {\em IEEE Sensors Journal}, 18(10):4200--4211.

\bibitem[Dong and Isler, 2018b]{dong2018tree}
Dong, W. and Isler, V. (2018b).
\newblock Tree morphology for phenotyping from semantics-based mapping in
  orchard environments.
\newblock {\em arXiv preprint arXiv:1804.05905}.

\bibitem[Edelsbrunner et~al., 1983]{edelsbrunner1983shape}
Edelsbrunner, H., Kirkpatrick, D., and Seidel, R. (1983).
\newblock On the shape of a set of points in the plane.
\newblock {\em IEEE Transactions on information theory}, 29(4):551--559.

\bibitem[Edelsbrunner and M{\"u}cke, 1994]{edelsbrunner1994three}
Edelsbrunner, H. and M{\"u}cke, E.~P. (1994).
\newblock Three-dimensional alpha shapes.
\newblock {\em ACM Transactions on Graphics (TOG)}, 13(1):43--72.

\bibitem[Fischler and Bolles, 1981]{fischler1981random}
Fischler, M.~A. and Bolles, R.~C. (1981).
\newblock Random sample consensus: a paradigm for model fitting with
  applications to image analysis and automated cartography.
\newblock {\em Communications of the ACM}, 24(6):381--395.

\bibitem[Forsyth and Ponce, 2011]{forsyth2011computer}
Forsyth, D. and Ponce, J. (2011).
\newblock {\em Computer vision: a modern approach}.
\newblock Upper Saddle River, NJ; London: Prentice Hall.

\bibitem[Golub and Van~Loan, 2012]{golub2012matrix}
Golub, G.~H. and Van~Loan, C.~F. (2012).
\newblock {\em Matrix computations}, volume~3.
\newblock JHU Press.

\bibitem[H{\"a}ni et~al., 2018a]{hani2018counting}
H{\"a}ni, N., Roy, P., and Isler, V. (2018a).
\newblock Apple counting using convolutional neural networks.
\newblock In {\em Intelligent Robots and Systems (IROS), 2018 IEEE/RSJ
  International Conference on}. IEEE.

\bibitem[H{\"a}ni et~al., 2018b]{hani2018comparative}
H{\"a}ni, N., Roy, P., and Isler, V. (2018b).
\newblock A comparative study of fruit detection and counting methods for yield
  mapping in apple orchards.
\newblock {\em arXiv preprint arXiv:1810.09499}.

\bibitem[He et~al., 2017]{he2017mask}
He, K., Gkioxari, G., Dollar, P., and Girshick, R. (2017).
\newblock Mask r-cnn.
\newblock In {\em 2017 IEEE International Conference on Computer Vision
  (ICCV)}, pages 2980--2988. IEEE.

\bibitem[He et~al., 2016]{he2016deep}
He, K., Zhang, X., Ren, S., and Sun, J. (2016).
\newblock Deep residual learning for image recognition.
\newblock In {\em Proceedings of the IEEE conference on computer vision and
  pattern recognition}, pages 770--778.

\bibitem[Huber, 1992]{huber1992robust}
Huber, P.~J. (1992).
\newblock Robust estimation of a location parameter.
\newblock In {\em Breakthroughs in statistics}, pages 492--518. Springer.

\bibitem[Hung et~al., 2015]{hung2015feature}
Hung, C., Underwood, J., Nieto, J., and Sukkarieh, S. (2015).
\newblock A feature learning based approach for automated fruit yield
  estimation.
\newblock In {\em Field and Service Robotics}, pages 485--498. Springer.

\bibitem[Levenberg, 1944]{levenberg1944method}
Levenberg, K. (1944).
\newblock A method for the solution of certain non-linear problems in least
  squares.
\newblock {\em Quarterly of applied mathematics}, 2(2):164--168.

\bibitem[Lin et~al., 2014]{lin2014microsoft}
Lin, T.-Y., Maire, M., Belongie, S., Hays, J., Perona, P., Ramanan, D.,
  Doll{\'a}r, P., and Zitnick, C.~L. (2014).
\newblock Microsoft coco: Common objects in context.
\newblock In {\em European conference on computer vision}, pages 740--755.
  Springer.

\bibitem[Lowe, 2004]{lowe2004distinctive}
Lowe, D.~G. (2004).
\newblock Distinctive image features from scale-invariant keypoints.
\newblock {\em International journal of computer vision}, 60(2):91--110.

\bibitem[Marquardt, 1963]{marquardt1963algorithm}
Marquardt, D.~W. (1963).
\newblock An algorithm for least-squares estimation of nonlinear parameters.
\newblock {\em Journal of the society for Industrial and Applied Mathematics},
  11(2):431--441.

\bibitem[Medeiros et~al., 2017]{medeiros2017modeling}
Medeiros, H., Kim, D., Sun, J., Seshadri, H., Akbar, S.~A., Elfiky, N.~M., and
  Park, J. (2017).
\newblock Modeling dormant fruit trees for agricultural automation.
\newblock {\em Journal of Field Robotics}, 34(7):1203--1224.

\bibitem[M{\'e}ndez et~al., 2014]{mendez2014deciduous}
M{\'e}ndez, V., Rosell-Polo, J.~R., Sanz, R., Escol{\`a}, A., and Catal{\'a}n,
  H. (2014).
\newblock Deciduous tree reconstruction algorithm based on cylinder fitting
  from mobile terrestrial laser scanned point clouds.
\newblock {\em Biosystems Engineering}, 124:78--88.

\bibitem[Mur-Artal and Tard{\'o}s, 2017]{mur2017orb}
Mur-Artal, R. and Tard{\'o}s, J.~D. (2017).
\newblock Orb-slam2: An open-source slam system for monocular, stereo, and
  rgb-d cameras.
\newblock {\em IEEE Transactions on Robotics}, 33(5):1255--1262.

\bibitem[Myronenko and Song, 2010]{myronenko2010point}
Myronenko, A. and Song, X. (2010).
\newblock Point set registration: Coherent point drift.
\newblock {\em IEEE transactions on pattern analysis and machine intelligence},
  32(12):2262--2275.

\bibitem[Newcombe et~al., 2011]{newcombe2011kinectfusion}
Newcombe, R.~A., Izadi, S., Hilliges, O., Molyneaux, D., Kim, D., Davison,
  A.~J., Kohi, P., Shotton, J., Hodges, S., and Fitzgibbon, A. (2011).
\newblock Kinectfusion: Real-time dense surface mapping and tracking.
\newblock In {\em Mixed and augmented reality (ISMAR), 2011 10th IEEE
  international symposium on}, pages 127--136. IEEE.

\bibitem[Peng et~al., 2016]{peng2016semantic}
Peng, C., Roy, P., Luby, J., and Isler, V. (2016).
\newblock Semantic mapping of orchards.
\newblock {\em IFAC-PapersOnLine}, 49(16):85--89.

\bibitem[Rodrigues, 1840]{rodrigues1840lois}
Rodrigues, O. (1840).
\newblock {\em Des lois g{\'e}om{\'e}triques qui r{\'e}gissent les
  d{\'e}placements d'un syst{\`e}me solide dans l'espace: et de la variation
  des cordonn{\'e}es provenant de ces d{\'e}placements consid{\'e}r{\'e}s
  ind{\'e}pendamment des causes qui peuvent les produire}.

\bibitem[Rosell and Sanz, 2012]{rosell2012review}
Rosell, J. and Sanz, R. (2012).
\newblock A review of methods and applications of the geometric
  characterization of tree crops in agricultural activities.
\newblock {\em Computers and Electronics in Agriculture}, 81:124--141.

\bibitem[Rosell et~al., 2009]{rosell2009obtaining}
Rosell, J.~R., Llorens, J., Sanz, R., Arno, J., Ribes-Dasi, M., Masip, J.,
  Escol{\`a}, A., Camp, F., Solanelles, F., Gr{\`a}cia, F., et~al. (2009).
\newblock Obtaining the three-dimensional structure of tree orchards from
  remote 2d terrestrial lidar scanning.
\newblock {\em Agricultural and Forest Meteorology}, 149(9):1505--1515.

\bibitem[Roy et~al., 2018a]{roy2018registering}
Roy, P., Dong, W., and Isler, V. (2018a).
\newblock Registering reconstructions of the two sides of fruit tree rows.
\newblock In {\em Intelligent Robots and Systems (IROS), 2018 IEEE/RSJ
  International Conference on}. IEEE.

\bibitem[Roy and Isler, 2016a]{roy2016surveying}
Roy, P. and Isler, V. (2016a).
\newblock Surveying apple orchards with a monocular vision system.
\newblock In {\em Automation Science and Engineering (CASE), 2016 IEEE
  International Conference on}, pages 916--921. IEEE.

\bibitem[Roy and Isler, 2016b]{roy2016counting}
Roy, P. and Isler, V. (2016b).
\newblock Vision-based apple counting and yield estimation.
\newblock In {\em Experimental Robotics}. Springer.

\bibitem[Roy and Isler, 2017]{roy2017active}
Roy, P. and Isler, V. (2017).
\newblock Active view planning for counting apples in orchards.
\newblock In {\em Intelligent Robots and Systems (IROS), 2017 IEEE/RSJ
  International Conference on}, pages 6027--6032. IEEE.

\bibitem[Roy et~al., 2018b]{roy2018arxiv}
Roy, P., Kislay, A., Plonski, P.~A., Luby, J., and Isler, V. (2018b).
\newblock Vision-based preharvest yield mapping for apple orchards.
\newblock {\em arXiv preprint arXiv:1808.04336}.

\bibitem[Salas-Moreno et~al., 2013]{salas2013slam++}
Salas-Moreno, R.~F., Newcombe, R.~A., Strasdat, H., Kelly, P.~H., and Davison,
  A.~J. (2013).
\newblock Slam++: Simultaneous localisation and mapping at the level of
  objects.
\newblock In {\em Computer Vision and Pattern Recognition (CVPR), 2013 IEEE
  Conference on}, pages 1352--1359. IEEE.

\bibitem[Sivic and Zisserman, 2009]{sivic2009efficient}
Sivic, J. and Zisserman, A. (2009).
\newblock Efficient visual search of videos cast as text retrieval.
\newblock {\em IEEE transactions on pattern analysis and machine intelligence},
  31(4):591--606.

\bibitem[Sotoodeh, 2006]{sotoodeh2006outlier}
Sotoodeh, S. (2006).
\newblock Outlier detection in laser scanner point clouds.
\newblock {\em International Archives of Photogrammetry, Remote Sensing and
  Spatial Information Sciences}, 36(5):297--302.

\bibitem[Stefas et~al., 2016]{stefas2016vision}
Stefas, N., Bayram, H., and Isler, V. (2016).
\newblock Vision-based uav navigation in orchards.
\newblock {\em IFAC-PapersOnLine}, 49(16):10--15.

\bibitem[Strasdat et~al., 2010]{strasdat2010scale}
Strasdat, H., Montiel, J., and Davison, A.~J. (2010).
\newblock Scale drift-aware large scale monocular slam.
\newblock {\em Robotics: Science and Systems VI}, 2.

\bibitem[Sturm et~al., 2012]{sturm2012benchmark}
Sturm, J., Engelhard, N., Endres, F., Burgard, W., and Cremers, D. (2012).
\newblock A benchmark for the evaluation of rgb-d slam systems.
\newblock In {\em Intelligent Robots and Systems (IROS), 2012 IEEE/RSJ
  International Conference on}, pages 573--580. IEEE.

\bibitem[Tabb and Medeiros, 2017]{tabbrobotic}
Tabb, A. and Medeiros, H. (2017).
\newblock A robotic vision system to measure tree traits.
\newblock In {\em Intelligent Robots and Systems (IROS), 2017 IEEE/RSJ
  International Conference on}, pages 6005--6012. IEEE.

\bibitem[Underwood et~al., 2016]{underwood2016mapping}
Underwood, J.~P., Hung, C., Whelan, B., and Sukkarieh, S. (2016).
\newblock Mapping almond orchard canopy volume, flowers, fruit and yield using
  lidar and vision sensors.
\newblock {\em Computers and Electronics in Agriculture}, 130:83--96.

\bibitem[Underwood et~al., 2015]{underwood2015lidar}
Underwood, J.~P., Jagbrant, G., Nieto, J.~I., and Sukkarieh, S. (2015).
\newblock Lidar-based tree recognition and platform localization in orchards.
\newblock {\em Journal of Field Robotics}, 32(8):1056--1074.

\bibitem[van~der Heijden et~al., 2012]{van2012spicy}
van~der Heijden, G., Song, Y., Horgan, G., Polder, G., Dieleman, A., Bink, M.,
  Palloix, A., van Eeuwijk, F., and Glasbey, C. (2012).
\newblock Spicy: towards automated phenotyping of large pepper plants in the
  greenhouse.
\newblock {\em Functional Plant Biology}, 39(11):870--877.

\bibitem[Wang et~al., 2013]{wang}
Wang, Q., Nuske, S., Bergerman, M., and Singh, S. (2013).
\newblock Automated crop yield estimation for apple orchards.
\newblock In Desai, J.~P., Dudek, G., Khatib, O., and Kumar, V., editors, {\em
  Experimental Robotics}, volume~88 of {\em Springer Tracts in Advanced
  Robotics}, pages 745--758. Springer International Publishing.

\bibitem[Wang and Li, 2014]{wang2014size}
Wang, W. and Li, C. (2014).
\newblock Size estimation of sweet onions using consumer-grade rgb-depth
  sensor.
\newblock {\em Journal of Food Engineering}, 142:153--162.

\bibitem[Wu, 2013]{wu2013towards}
Wu, C. (2013).
\newblock Towards linear-time incremental structure from motion.
\newblock In {\em 3DTV-Conference, 2013 International Conference on}, pages
  127--134. IEEE.

\bibitem[Yu et~al., 2015]{yu2015semantic}
Yu, F., Xiao, J., and Funkhouser, T. (2015).
\newblock Semantic alignment of lidar data at city scale.
\newblock In {\em Proceedings of the IEEE Conference on Computer Vision and
  Pattern Recognition}, pages 1722--1731.

\bibitem[Zhou et~al., 2016]{zhou2016fast}
Zhou, Q.-Y., Park, J., and Koltun, V. (2016).
\newblock Fast global registration.
\newblock In {\em European Conference on Computer Vision}, pages 766--782.
  Springer.

\end{thebibliography}

\end{document}